\PassOptionsToPackage{dvipsnames}{xcolor}
\documentclass[sigconf]{acmart}

\usepackage{graphicx,subcaption}
\usepackage{commath}
\usepackage{array}
\usepackage{dsfont} 
\DeclareMathOperator*{\argmax}{arg\,max} 

\copyrightyear{2026}
\acmYear{2026}
\setcopyright{cc}
\setcctype{by}
\acmConference[WWW '26]{Proceedings of the ACM Web Conference 2026}{April 13--17, 2026}{Dubai, United Arab Emirates}
\acmBooktitle{Proceedings of the ACM Web Conference 2026 (WWW '26), April 13--17, 2026, Dubai, United Arab Emirates}
\acmPrice{}
\acmDOI{10.1145/3774904.3792518}
\acmISBN{979-8-4007-2307-0/2026/04}

\author{Pallabee Das}
\orcid{0009-0002-5146-7463}
\affiliation{%
  \institution{Paderborn University}
  \city{Paderborn}
  \country{Germany}
}
\email{pdas@mail.uni-paderborn.de}

\author{Stefan Heindorf}
\orcid{0000-0002-4525-6865}
\affiliation{%
  \institution{Paderborn University}
  \city{Paderborn}
  \country{Germany}
}
\email{heindorf@uni-paderborn.de}

\renewcommand{\shortauthors}{Pallabee Das and Stefan Heindorf}

\begin{document}

\title{Discrete Diffusion-Based Model-Level Explanation of Heterogeneous GNNs with Node Features}

\begin{abstract}
Many real-world datasets, such as citation networks, social networks, and molecular structures, are naturally represented as heterogeneous graphs, where nodes belong to different types and have additional features. For example, in a citation network, nodes representing ``Paper'' or ``Author'' may include attributes like keywords or affiliations. A critical machine learning task on these graphs is node classification, which is useful for applications such as fake news detection, corporate risk assessment, and molecular property prediction. Although Heterogeneous Graph Neural Networks (HGNNs) perform well in these contexts, their predictions remain opaque. Existing post-hoc explanation methods lack support for actual node features beyond one-hot encoding of node type and often fail to generate realistic, faithful explanations. To address these gaps, we propose DiGNNExplainer, a model-level explanation approach that synthesizes heterogeneous graphs with realistic node features via discrete denoising diffusion. In particular, we generate realistic discrete features (e.g., bag-of-words features) using diffusion models within a discrete space, whereas previous approaches are limited to continuous spaces. We evaluate our approach on multiple datasets and show that DiGNNExplainer produces explanations that are realistic and faithful to the model’s decision-making, outperforming state-of-the-art methods.
\end{abstract}

\keywords{Graph neural networks; Explainable artificial intelligence; Diffusion models}

\ccsdesc[300]{Computing methodologies~Supervised learning by classification}
\ccsdesc[300]{Computing methodologies~Neural networks}

\maketitle

\section{Introduction}

\begin{figure*}[t!]
  \centering
  \includegraphics[width=0.8\linewidth]{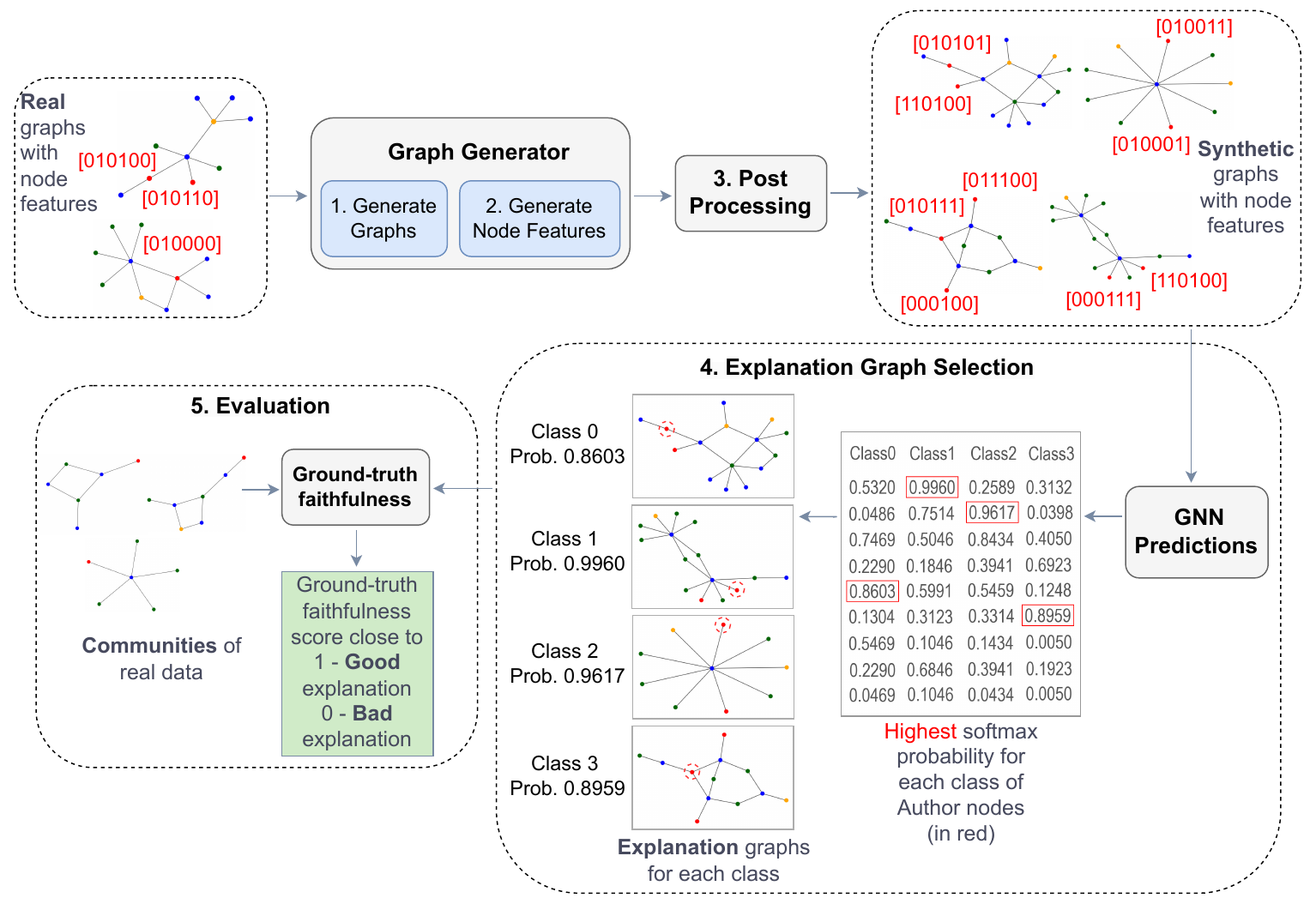}
  \caption{Overview of DiGNNExplainer. The example, based on the \textsc{DBLP} dataset, illustrates the process of generating explanations for a node classification task. Nodes represent different entity types: authors (red, \textcolor{red}{$\bullet$}), papers (blue, \textcolor{blue}{$\bullet$}), conferences (orange, \textcolor{orange}{$\bullet$}), and terms (green, \textcolor{OliveGreen}{$\bullet$}). Node features, such as discrete bag-of-words representations of author keywords (e.g., \textcolor{red}{[010100]}), are encoded in red. The task is to predict the class of each author, i.e., their research area: database, data mining, artificial intelligence, or information retrieval. (1) \& (2) The graph generator takes as input small real-world graphs with discrete node features and produces synthetic graphs with discrete node features. (3) The validity of these generated graphs is checked against the metagraph. (4) A trained GNN, which is to be explained, is applied to each of the nine author nodes of valid, synthesized graphs, producing a prediction for each of the four classes. For each class, we identify the node and corresponding graph with the highest prediction probability, yielding one explanation graph per class. (5)~To evaluate our approach, we compute various metrics, including the ground-truth faithfulness
  between our explanation graphs and the communities within the real dataset.}
  \label{fig:overview}
\end{figure*}

Complex real-world data, such as citation networks, social networks, and molecular structures, often exhibit an underlying graph structure. Nodes in these graphs typically have different types and are associated with additional features, i.e., they are heterogeneous graphs. For instance, in a citation network, there might be nodes of type ``Paper'' and ``Author'' each of which has additional node features like keywords or affiliations. Whereas related work sometimes uses the term ``node feature'' only to refer to a one-hot encoding of the node type~\cite{Yuan2020XGNN, Wang2023GNNInterpreter}, we refer to actual attributes of nodes (e.g., age of persons, keywords of papers). A key machine learning task on such graphs is node classification, e.g., for fake news detection~\cite{Zhang2024HeteroSGT}, corporate risk assessment~\cite{Zhang2024RiskAssessment}, or predicting the mutagenicity of molecules~\cite{Luo2020PGExplainer}. Towards this end, a number of powerful approaches for heterogeneous graphs have been proposed, e.g., Heterogeneous Graph Neural Networks (HGNNs)~\cite{Wang2019HAN}. 

However, the predictions made by HGNNs are opaque, providing no insight into the reasoning behind their decisions. In high-stakes applications---such as healthcare, finance, or criminal justice---this lack of transparency is a significant limitation, as it prevents users from understanding the rationale behind the model's decisions. Towards this end, several post-hoc explanation methods have been proposed, such as GNNExplainer~\cite{Ying2019GNNExplainer}, SubgraphX~\cite{Yuan2021SubgraphX}, PGExplainer~\cite{Luo2020PGExplainer}, XGNN~\cite{Yuan2020XGNN}, GNNInterpreter~\cite{Wang2023GNNInterpreter}, and D4Explainer~\cite{Chen2023d4explainer}. Despite their contributions, these methods face significant limitations: (1)~the first three do not provide model-level explanations, (2)~the second three lack support for heterogeneous graphs and node features, (3)~they are not designed to generate plausible, realistic graphs as explanations. Although node feature support has been identified as important future work~\cite{Chen2023d4explainer}, most explainers still lack it. This is a critical gap, as GNNs tend to perform poorly without node features~\cite{Duong2019GNNnodefeatures}.

In this paper, we introduce DiGNNExplainer, a model-level explanation approach for heterogeneous GNNs that synthesizes realistic heterogeneous graphs with \emph{node features} (Figure~\ref{fig:overview}). Our approach consists of five key steps:
(1)~We generate realistic graph structures using discrete denoising diffusion with DiGress~\cite{Vignac2023DiGress}.
(2)~We generate realistic node features using discrete denoising diffusion. Specifically, we extend TabDDPM~\cite{Kotelnikov2023TabDDPM}, a diffusion model originally designed for tabular data, to synthesize realistic, discrete node features---yielding a novel feature generator dubbed DiTabDDPM. Both our graph generator and our feature generator are trained on the same dataset as the GNN, keeping the generated graphs in-distribution, i.e., they reflect the patterns and statistical properties of the training data.
(3)~We check whether the synthesized graphs adhere to the metagraph and discard graphs that do not.
(4)~For each class, we select the synthetic graph with node features that maximizes the GNN's prediction score for that class, ensuring that the selected graph captures the essential information relevant to the model's decision-making. 
(5)~We evaluate our synthesized explanation graphs with node features both in terms of realism and faithfulness. Specifically, we assess whether they preserve graph distributions, e.g., node degrees, clustering coefficients, and spectrum in terms of maximum mean discrepancy (MMD), and the generated explanations are faithful, i.e., achieve a high GNN prediction score (predictive faithfulness), and retain graph communities of the original graph (ground-truth faithfulness). We compare DiGNNExplainer to previous model-level explanation approaches and conduct an ablation study on different node feature generation techniques.

We make the following contributions:
\begin{itemize}
\item DiGNNExplainer generates \emph{model-level} explanations including \emph{actual node features}, unlike previous methods that either lack model-level explanations or ignore node features.
\item DiGNNExplainer introduces a novel explanation pipeline as shown in Figure~\ref{fig:overview}.
\item We introduce a novel explanation graph generation technique, combining forest fire sampling with multiple DiGress models for different graph sizes.
\item Unlike existing tabular data generators~\cite{Kotelnikov2023TabDDPM, Jo2022Score-based}, which map all node features into a continuous space, our novel DiTabDDPM directly models discrete features using discrete diffusion, improving performance.
\item Our evaluation shows that DiGNNExplainer synthesizes realistic explanation graphs closely matching the graph statistics (Table~\ref{tab:table-mmd-main}) and node feature statistics of the original training dataset (Table~\ref{tab:table-evaluation-feature-selection}), and achieves higher faithfulness---both predictive faithfulness and ground-truth faithfulness (Table~\ref{tab:table-gnn-faithfulness}).
\end{itemize} 

\section{Related Work}

We briefly survey explanation methods for graph neural networks (GNNs) and generative models, which we employ to generate explanations.

\emph{Model-level GNN explainers} explain why a GNN predicts a certain class by generating a graph that maximizes the predicted score for this class. Notable methods include XGNN~\cite{Yuan2020XGNN}, GNNInterpreter~\cite{Wang2023GNNInterpreter}, and D4Explainer~\cite{Chen2023d4explainer}. We compare them in terms of their graph generation, node feature generation, and evaluation. \emph{Graph generation:} XGNN employs reinforcement learning to iteratively add edges or nodes with one-hot encoded types. GNNInterpreter uses a probabilistic generative model generating edges and nodes independently with a fixed probability. D4Explainer, like our approach, utilizes diffusion-based graph generation but lacks support for different node types. \emph{Node feature generation:} XGNN only supports one-hot encoded node types. GNNInterpreter proposes the use of actual node features but evaluates only node types. D4Explainer briefly discusses node features but focuses on structural graph generation. \emph{Evaluation: } All three methods evaluate graphs based on predicted class scores. Only D4Explainer considers simple graph statistics; XGNN and GNNInterpreter do not. None of the approaches quantitatively assesses the graphs in terms of generated communities and distribution of generated node features. While some methods provide limited support for node features, none of them effectively address node feature generation as shown in our evaluation.

Alternative explanation strategies use description logics or natural language instead of graph structures as explanation~\cite{Kohler2024Utilizing, Sapkota2024EDGE, Heindorf2025NLEF, Himmelhuber2021Combining, Pluska2024Logical}, e.g., via OntoLearn~\cite{Demir2025OntoLearn} and EvoLearner~\cite{Heindorf2022EvoLearner}. While some of these approaches support node features, they do not generate graphs as final explanations.

\emph{Instance-level GNN explainers} such as GNNExplainer \cite{Ying2019GNNExplainer} and SubgraphX~\cite{Yuan2021SubgraphX} explain each GNN prediction by identifying a different subgraph, making it challenging to understand the overall GNN behavior. As pointed out by \citet{Chen2023d4explainer}, these methods derive their subgraphs from the training datasets, leading to unreliable explanations for out-of-distribution data.

\emph{Graph generators} learn the distribution of a set of graphs to generate similar graphs. GraphVAE~\cite{Simonovsky2018GraphVAE} employs a variational autoencoder to embed graphs into a continuous latent space and generates new graphs by sampling from this space. DiGress~\cite{Vignac2023DiGress} uses a discrete diffusion-based approach reconstructing graphs step-by-step in a discrete space, preserving structural information and node-type distributions more effectively. Both GraphVAE and DiGress have demonstrated strong performance on molecular datasets like \textsc{MUTAG} and \textsc{QM9}. HGEN~\cite{Chen2021HGEN} employs a Generative Adversarial Network (GAN) to preserve metapath relations of heterogeneous datasets, such as \textsc{DBLP}, \textsc{PubMed}, and \textsc{IMDB}. HGEN generates graphs with the same number of nodes as the input graph with a minimum number of 100 nodes, limiting its ability to produce small, human-intelligible graphs that may serve as an explanation. These approaches focus on structural graph generation with limited support for actual node features: HGEN does not generate node features at all, while GraphVAE and DiGress only generate discrete, categorical node features focusing on one-hot encoding of node types. Their performance for node feature generation has not been evaluated. Moreover, none of these approaches has been used and evaluated for generating explanation graphs as done in this work.

\emph{Tabular feature generation} TabDDPM~\cite{Kotelnikov2023TabDDPM} uses diffusion models to synthesize discrete and continuous tabular data by learning a continuous space. To the best of our knowledge, it has never been used and evaluated to generate node features of graphs. Moreover, it has never been used for generating explanations. We extend TabDDPM to synthesize realistic discrete features by modeling them directly within a discrete space.

\section{DiGNNExplainer}

Our DiGNNExplainer provides model-level explanations for heterogeneous GNNs. To achieve this, we generate explanation graphs by sampling from the training data distribution using a graph generator (Figure~\ref{fig:overview}). We use the trained GNN that we want to explain to make predictions on the synthetically generated graphs. The graph with the highest predicted probability for a class serves as the explanation for that class. Our graph generator is designed to produce plausible explanation graphs that do not need to be subgraphs of the training data.

\subsection{Heterogeneous Graph with Node Features}
\label{subsec:formal-definition}

Our explanation graph can be formally defined as a heterogeneous graph $G = (V, E, T, X)$ where $V$ denotes the set of nodes, $E \subseteq \{\{ u,v \}|u,v \in V\}$ the set of undirected edges, $T$ the set of node types, and $X=\{\textbf{X}_1, \ldots, \textbf{X}_{|T|}\}$ the set of feature matrices. Each node $v \in V$ is associated with a node type $\tau \in T$. The feature matrix for node type~$\tau$ is denoted by $\textbf{X}_\tau\in\mathbb{R}^{N_{\tau}\times d_{\tau}}$, where $N_\tau$ is the number of nodes of type $\tau$ and $d_\tau$ the number of features per node. Each node type can either have continuous $\textbf{X}_\tau\in\mathbb{R}^{N_\tau\times d_\tau}$ or discrete $\textbf{X}_{\tau} \in \mathcal{X}_\tau^{N_\tau\times d_\tau}$ features where $\mathcal{X}_{\tau}=\{x_{\tau}^1, \ldots, x_{\tau}^{k_\tau}\}$ denotes the set of allowed discrete values for type $\tau$.

\subsection{Generating Graphs and Node Features with Diffusion Models}

We generate explanation graphs and node features through diffusion approaches (see Figure~
\ref{fig:overview}). We do so
(1)~using diffusion models for the graph structure and 
(2)~diffusion models for the node features,
(3)~discard graphs that are inconsistent with the metagraph, and
(4)~we select one explanation graph per class.
In the following, we give an overview of our approach. Detailed descriptions of the diffusion processes using DiGress and TabDDPM, and formal definitions of DiGNNExplainer, are provided in the Appendix.

\paragraph{1. Graph generation}
When synthesizing graphs, our goal is twofold: first, to synthesize graphs that are plausible and match the distribution of our training data; second, to produce small, human-intelligible graphs that can serve as explanations.

We build upon DiGress~\cite{Vignac2023DiGress} to synthesize heterogeneous graph structures. Since DiGress performs well in generating graphs of a fixed size~$n$ and requires training data of the same size, while the best size of an explanation graph is not known in advance, we create multiple training sets for DiGress with different graph sizes and train a different DiGress model for each graph size.%
\footnote{In our evaluation, we typically use graph sizes between 10 and 15, as they provide meaningful explanations while remaining interpretable.}
For graph classification, we train the graph generators directly on datasets containing multiple small graphs of different sizes. In contrast, for node classification, where the data consists of a single large graph, we extract subgraphs of varying sizes using graph sampling techniques: We employ the forest fire sampling algorithm~\cite{Rozemberczki2020Little}, known for preserving graph connectivity and performing well in prior studies. To ensure that the sampled subgraphs include the nodes to be classified, we initiate the sampling process at nodes of the relevant types. For example, we start a fire at author nodes in the \textsc{DBLP} dataset. Specifically, for each graph size $n$, we generate $k$ subgraphs by repeating the following process until $k$ valid subgraphs are obtained: First, we randomly select a node of the type that is to be classified, then we start the fire from this node and stop once exactly $n$ nodes have been burned. If fewer than $n$ nodes are burned, we repeat the process. The nodes burned serve as sampled graphs. Following prior work, we use the default hyperparameters of the little-ball-of-fur library~\cite{Rozemberczki2020Little}, including a burning probability of 0.4.%
\footnote{\raggedright\url{https://little-ball-of-fur.readthedocs.io/en/latest/_modules/littleballoffur/exploration_sampling/forestfiresampler.html}}

\paragraph{2. Node feature generation}
Given a node type $\tau$ with node feature matrix $\mathbf{X}_\tau$ (cf., Section \ref{subsec:formal-definition}), we train multiple node feature generators as follows: If $\tau$ is the node type to be classified, we train one feature generator per class to simulate different feature distributions; if $\tau$ is not the node type to be classified, we train one feature generator; if $\tau$ is a node type with no node features at all, no feature generator is trained. For example, for the \textsc{DBLP} dataset, we train six node feature generators: four node feature generators for the four different classes of author nodes (database, data mining, artificial intelligence, information retrieval), one for paper nodes, one for term nodes, and zero for conference nodes as they do not have node features.

If $\tau$ is a node type with a \emph{continuous} feature matrix, we use the original TabDDPM~\cite{Kotelnikov2023TabDDPM} approach with continuous noise; for each node type $\tau$ with \emph{discrete} feature matrix $\mathbf{X}_\tau$, we use our novel DiTabDDPM which combines the discrete noise from DiGress~\cite{Vignac2023DiGress} and the \texttt{MLPBlock} architecture from TabDDPM~\cite{Kotelnikov2023TabDDPM} as follows:

\emph{Forward diffusion process.} The forward diffusion process for node features follows DiGress~\cite{Vignac2023DiGress}, using discrete noise, marginal transitions, and a cosine noise scheduler. 

\emph{Reverse diffusion process.} In the reverse process, for each time step~$t$ and node type~$\tau$, given a discrete feature vector $\mathbf{x_t} \in \mathcal{X}_{\tau}^{d_\tau}$, we encode it via one-hot encoding as a vector $\mathbf{x_t^{(ohe)}} \in \{0,1\}^{d_\tau \times k_\tau}$ into the denoising network which outputs the feature distribution vector $\mathbf{p}_{t-1} \in [0,1]^{d_\tau \times k_\tau}$ for its previous step. Then we sample the discrete features $\mathbf{x}_{t-1} \in \mathcal{X}_\tau^{d_\tau}$ of the previous time step $t-1$ from the multinomial distribution defined by $\mathbf{p}_{t-1}$ and repeat the process. 

\emph{Node feature loss}. The node feature training loss $L_{train}$ for a single node is the cross-entropy
\[
L_{train} := \frac{1}{d_\tau}\sum_{i=1}^{d_\tau} \mathrm{CrossEntropy}(\mathbf{p}^{(i)}_{pred}, \mathbf{x}^{(i)}_{true})
\]
where $\mathbf{p}^{(i)}_{pred} \in [0,1]^{k_\tau}$ is the vector of predicted probabilities, and $\mathbf{x}^{(i)}_{true} \in \{x^1_\tau, \ldots , x^{k_\tau}_\tau\}$ is the true discrete value of feature $i$ in the dataset.

\paragraph{3. Postprocessing.} We check whether the generated graph is consistent with the metagraph of the dataset. Otherwise, we discard the graph.

\paragraph{4. Explanation graph selection.} We obtain one explanation graph per class with the highest prediction score. 

\section{Evaluation}

In this section, we evaluate how well our DiGNNExplainer explains the predictions of GNNs. After briefly introducing our evaluation setup, we compare DiGNNExplainer to three state-of-the-art GNN explanation methods. Additionally, we compare our discrete diffusion-based approach with a variant that uses variational autoencoders, and we investigate the impact of varying the number of discrete features. Our code is publicly available.%
\footnote{\url{https://github.com/ds-jrg/DiGNNExplainer}}
Additional experimental results, including dataset statistics, hyperparameters, and implementation details, can be found in the Appendix.

\subsection{Evaluation Setup}
\label{subsec:evaluation-setup}

\paragraph{Datasets.} We conduct experiments on seven datasets for node and graph classification~\cite{Fu2020MAGNN, Morris2020TUDataset,Ying2019GNNExplainer, Wang2021ReFine}---three real-world datasets (\textsc{DBLP}, \textsc{IMDB}, \textsc{MUTAG}) and four synthetic datasets (\textsc{BA-Shapes}, \textsc{Tree-Cycle}, \textsc{Tree-Grids}, \textsc{BA-3Motif}) in Table~\ref{tab:table-datasets}. \textsc{DBLP} and \textsc{IMDB} have actual node features, the other datasets not.

\paragraph{GNNs.} We use a high-performing GNN architecture for each dataset: GraphSAGE~\cite{Hamilton2017GraphSAGE} for DBLP, HAN~\cite{Wang2019HAN} for \textsc{IMDB}, and three-layer GCNs for MUTAG and the synthetic datasets. To prevent overfitting, we apply early stopping with a patience of 100 epochs, halting training if the validation score does not improve within this window.

\paragraph{Baseline explainers.}
We compare DiGNNExplainer to the following state-of-the-art model-level explanation methods:
(1)~XGNN~\cite{Yuan2020XGNN}, which leverages reinforcement learning to generate explanations. We extend the original implementation to support node classification tasks.
(2)~GNNInterpreter~\cite{Wang2023GNNInterpreter}, which improves upon XGNN by being computationally efficient. We extend it for node classification by creating class-specific embeddings using average node pooling.
(3)~D4Explainer~\cite{Chen2023d4explainer}, which employs a diffusion-based approach to generate model-level explanations for both node and graph classification tasks.

\paragraph{Baseline graph generators with node features.} 
We compare our discrete diffusion-based graph generator against two baselines: Our first baseline is a variational autoencoder-based model that jointly learns the graph structure and node features. Our second baseline replaces our discrete diffusion-based node feature generator DiTabDDPM with the continuous diffusion-based generator TabDDPM. Regarding the former, for MUTAG and the synthetic datasets, we use the original implementation of GraphVAE%
\footnote{\url{https://github.com/deepfindr/gvae}}
\cite{Simonovsky2018GraphVAE} using only node types as node features. For \textsc{DBLP} and \textsc{IMDB}, we extend GraphVAE to generate explanation graphs with actual node features by concatenating one-hot encoded node types to actual node features. We omit GraphVAE's support for node order invariance as it does not scale to graphs of our size. As GraphVAE requires all nodes to have the same number of features, but \textsc{DBLP} has a varying number of features per node type, we select a feature size of 50 (using feature selection) for all node types of \textsc{DBLP}, to be able to compare all our \textsc{DBLP} experiments with the baseline. For both the VAE and TabDDPM baselines, which always operate in a continuous feature space, we discretize features that are supposed to be discrete (e.g., bag-of-word features) via a threshold.

\paragraph{Evaluation metrics.} We evaluate our approach in terms of three aspects: whether the generated graph structures have a similar distribution as the training set, whether the node features have a similar distribution as the training set, and whether our explanations are faithful.

Regarding the graph structure, following \citet{Chen2023d4explainer}, we compare the distribution of node degrees, clustering coefficients, and spectrum in terms of maximum mean discrepancy (MMD). Additionally, we consider the MMD of the node type distribution as we are dealing with heterogeneous graphs.

To assess the similarity between synthetically generated node feature vectors $\mathbf{X}_{\text{gen}}$ and those from the training set $\mathbf{X}_{\text{train}}$, we compute the average pairwise cosine similarity
$\frac{1}{mn} \sum_{i=1}^{m} \sum_{j=1}^{n} \cos\left(\mathbf{x}^{(i)}_{\text{gen}}, \mathbf{x}^{(j)}_{\text{train}}\right)$, 
where \( \cos(\mathbf{x}, \mathbf{y}) = \frac{\mathbf{x} \cdot \mathbf{y}}{\|\mathbf{x}\| \|\mathbf{y}\|} \), and \( \mathbf{x}^{(i)}_{\text{gen}} \in \mathbf{X}_{\text{gen}},\ \mathbf{x}^{(j)}_{\text{train}} \in \mathbf{X}_{\text{train}} \).

To assess \emph{faithfulness}, following~\cite{agarwal2022openxai}, we assess both \emph{predictive} faithfulness $\mathit{PF}$
and \emph{ground-truth} faithfulness $\mathit{GF}$:
Regarding the former, we compute the average class score across all classes as
$\mathit{PF}:= \frac{1}{C} \sum_{i=1}^{C} P({G_{i}})$
where \( C \) is the number of classes, \( \mathcal{G}_i \) is our explanation graph for class \( i \), and \( P(G_{i}) \) is the GNN's predicted probability for class \( i \) on explanation graph \( G_i \). A higher score indicates greater alignment between the explanation graphs and the GNN's original decision. Regarding the latter, we compute ground-truth faithfulness as follows: For each class, we assume that we have a set of ground-truth explanation motifs (which are readily available for synthetic datasets and can be obtained via community detection for real-world datasets as described in the appendix). Then we define 
$\mathit{GF} := \frac{1} {C}\sum_{i=1}^{C} \frac{1}{M_i} \sum_{j=1}^{M_i} \mathds{1}_{(M_{i,j} \subseteq G_i)}$
where $M_i$ is the number of available ground-truth motifs for class $i$, $G_i$ is the explanation graph for class $i$, and $\mathds{1}_{(M_{i,j} \subseteq G_i)}$ is the indicator function that equals 1 if the motif \(M_{i,j}\), is a subgraph of the explanation graph, and 0 otherwise. A higher ground-truth faithfulness score indicates that our generated explanations are closer to the ground-truth explanations.

\paragraph{Train-test splits.} For the node classification datasets \textsc{DBLP}, \textsc{IMDB}, \textsc{BA-Shapes}, \textsc{Tree-Cycle}, and \textsc{Tree-Grids}, we synthesize 1,536 graphs per dataset, some of which may be discarded as they are inconsistent with the metagraph. We use 200 graphs to train the GNN, splitting them into 128 for training, 32 for validation, and 40 for testing the graph generation model. For \textsc{MUTAG} (58 graphs), we allocate 38 for training, 9 for validation, and 11 for testing. For \textsc{BA-3Motif} (323 graphs), the split is 206 for training, 52 for validation, and 65 for testing. To generate candidate explanation graphs, we use our graph generation model to produce 256 graphs for each of the 6 graph sizes (either 10 to 15 nodes or 5 to 10 nodes), resulting in 1,536 candidate graphs in total.

\begin{table*}[htbp]
  \centering
  \caption{Comparisons for average MMD distances between explanation graphs of each class and the reference graphs, for node type distributions (NTD), node degree distributions (Deg.), clustering coefficients (Clus.), spectrum distributions (Spec.).}
  \label{tab:table-mmd-main}
  \scriptsize
  \setlength{\tabcolsep}{0.5pt}
  \begin{tabular}{@{}>{\arraybackslash}m{0.7in}>
    {\centering\arraybackslash}m{0.21in}>{\centering\arraybackslash}m{0.21in}>{\centering\arraybackslash}m{0.21in}>{\centering\arraybackslash}m{0.21in}>
    {\centering\arraybackslash}m{0.21in}>{\centering\arraybackslash}m{0.21in}>{\centering\arraybackslash}m{0.21in}>{\centering\arraybackslash}m{0.21in}>
    {\centering\arraybackslash}m{0.21in}>{\centering\arraybackslash}m{0.21in}>{\centering\arraybackslash}m{0.21in}>{\centering\arraybackslash}m{0.21in}>
    {\centering\arraybackslash}m{0.21in}>{\centering\arraybackslash}m{0.21in}>{\centering\arraybackslash}m{0.21in}>{\centering\arraybackslash}m{0.21in}>
    {\centering\arraybackslash}m{0.21in}>{\centering\arraybackslash}m{0.21in}>{\centering\arraybackslash}m{0.21in}>{\centering\arraybackslash}m{0.21in}>
    {\centering\arraybackslash}m{0.21in}>{\centering\arraybackslash}m{0.21in}>{\centering\arraybackslash}m{0.21in}>{\centering\arraybackslash}m{0.21in}> 
    {\centering\arraybackslash}m{0.21in}>{\centering\arraybackslash}m{0.21in}>{\centering\arraybackslash}m{0.21in}>{\centering\arraybackslash}m{0.21in}@{}}
    \toprule
    \textbf{Approaches} & \multicolumn{4}{c}{\textbf{\textsc{DBLP}}} & \multicolumn{4}{c}{\textbf{\textsc{IMDB}}} & \multicolumn{4}{c}{\textbf{\textsc{MUTAG}}} & \multicolumn{4}{c}{\textbf{\textsc{BA-Shapes}}} & \multicolumn{4}{c}{\textbf{\textsc{Tree-Cycle}}} & \multicolumn{4}{c}{\textbf{\textsc{Tree-Grids}}} & \multicolumn{4}{c}{\textbf{\textsc{BA-3Motif}}} \\
    \cmidrule(lr){2-5} \cmidrule(lr){6-9} \cmidrule(lr){10-13} \cmidrule(lr){14-17} \cmidrule(lr){18-21} \cmidrule(lr){22-25} \cmidrule(lr){26-29}
    &NTD$\downarrow$ & Deg.$\downarrow$ & Clus.$\downarrow$ & Spec.$\downarrow$&NTD$\downarrow$ & Deg.$\downarrow$ & Clus.$\downarrow$ & Spec.$\downarrow$ & NTD$\downarrow$ & Deg.$\downarrow$ & Clus.$\downarrow$ & Spec.$\downarrow$ & NTD$\downarrow$ & Deg.$\downarrow$ & Clus.$\downarrow$ & Spec.$\downarrow$ & NTD$\downarrow$ & Deg.$\downarrow$ & Clus.$\downarrow$ & Spec.$\downarrow$ & NTD$\downarrow$ & Deg.$\downarrow$ & Clus.$\downarrow$ & Spec.$\downarrow$ & NTD$\downarrow$ & Deg.$\downarrow$ & Clus.$\downarrow$ & Spec.$\downarrow$ \\
    \toprule
    \textsc{XGNN} & 0.708 & 0.802 & 0.983 & \textbf{0.054} & 1.065 & 1.138 & 1.122 & 0.064 & 0.0 & 0.216 & 0.853 & 0.080 & 0.0 & \textbf{0.103} & \textbf{0.138} & \textbf{0.041} & 0.0 & 0.239 & \textbf{0.378} & 0.239 & 0.0 & 0.172 & 0.880 & 0.047 & 0.0 & 0.174 & \textbf{0.247} & \textbf{0.060} \\  
    \textsc{GNNInterpreter} & 0.708 & 0.946 & 1.311 & 0.077 & 1.065 & 0.730 & 1.161 & 0.058 & 0.0 & 0.240 & 1.002 & 0.071 & 0.0 & 1.223 & 1.070 & 0.043 & 0.0 & 1.261 & 1.480 & 0.047 & 0.0 & 0.976 & 1.405 & 0.046 & 0.0 & 1.411 & 1.251 & 0.068 \\
    \textsc{D4Explainer} & 0.708 & 0.493 & 0.001 & 0.111 & 1.065 & 0.356 & 0.806 & \textbf{0.051} & 0.0 & 0.603 & 0.033 & 0.587 & 0.0 & 0.980 & 1.037 & 0.042 & 0.0 & 0.644 & 0.829 & \textbf{0.046} & 0.0 & 0.625 & 1.181 & \textbf{0.043} & 0.0 & 1.371 & 0.968 & 0.066 \\
    \textsc{VAE} & 0.121 & 0.136 & \textbf{0.0} & 0.570 & 0.357 & 0.205 & \textbf{0.0} & 0.590 & 0.0 & 0.683 & 1.031 & \textbf{0.062} & 0.0 & 1.319 & 1.006 & 1.021 & 0.0 & 1.484 & 1.660 & 0.778 & 0.0 & 0.860 & 1.404 & 0.296 & 0.0 & 0.768 & 0.652 & 0.641 \\ 
    \midrule
    \textsc{\textbf{DiGNNExplainer}} & \textbf{0.023} & \textbf{0.096} & \textbf{0.0} & 0.556 & \textbf{0.011} & \textbf{0.062} & \textbf{0.0} & 0.288 & 0.0 & \textbf{0.213} & \textbf{0.0} & 1.042 & 0.0 & 0.404 & 0.891 & 1.021 & 0.0 & \textbf{0.174} & 0.777 & 1.026 & 0.0 & \textbf{0.129} & \textbf{0.060} & 0.258 & 0.0 & \textbf{0.113} & 0.272 & 0.728\\
    \bottomrule
  \end{tabular}
\end{table*}

\begin{table}
  \centering
  \scriptsize
  \setlength{\tabcolsep}{10pt}
  \caption{Feature selection experiments for \textsc{DBLP} and \textsc{IMDB}. We report the average cosine similarity for node features (Node feat. (cos)), and the mean and variance of predictive faithfulness (PF) and ground-truth faithfulness (GF) over 10 runs. We compare our proposed DiTabDDPM with the original TabDDPM and the VAE baseline.}
  \label{tab:table-evaluation-feature-selection}
  \begin{tabular}{@{}lccc@{}}
    \toprule
    \textbf{Approach} & \textbf{Node feat. (cos)$\uparrow$}& \textbf{PF$\uparrow$}& \textbf{GF$\uparrow$} \\
    \midrule
    \multicolumn{4}{c}{\textsc{DBLP}} \\ 
    \midrule
    DiTabDDPM (50 feat.)       & 0.229 & \textbf{0.999} \(\pm\) 0.0 & \textbf{0.882} \(\pm\) 0.029  \\
    DiTabDDPM (8 discr. feat.) & 0.353 & 0.835 \(\pm\) 0.090        & 0.794 \(\pm\) 0.052           \\
    DiTabDDPM (4 discr. feat.) & 0.386 & 0.818 \(\pm\) 0.056        & 0.809 \(\pm\) 0.051           \\
    DiTabDDPM (2 discr. feat.) & \textbf{0.397} & 0.975 \(\pm\) 0.036        & 0.817 \(\pm\) 0.048           \\
    \midrule
    TabDDPM (50 feat.)       & 0.196 & 0.993 \(\pm\) 0.005 & 0.829 \(\pm\) 0.058  \\ 
    TabDDPM (8 discr. feat.) & 0.221 & 0.995 \(\pm\) 0.002 & 0.822 \(\pm\) 0.052  \\
    TabDDPM (4 discr. feat.) & 0.232 & 0.965 \(\pm\) 0.046 & 0.825 \(\pm\) 0.086  \\
    TabDDPM (2 discr. feat.) & 0.235 & 0.979 \(\pm\) 0.037 & 0.827 \(\pm\) 0.032  \\
    \midrule
    VAE (baseline) & 0.0 & 0.655 \(\pm\) 0.042 & 0.717 \(\pm\) 0.046  \\
    \midrule
    \multicolumn{4}{c}{\textsc{IMDB}} \\ 
    \midrule
    DiTabDDPM (3066 feat., all) & 0.029 & 0.955 \(\pm\) 0.084 & 0.890 \(\pm\) 0.029  \\
    DiTabDDPM (10 discr. feat.) & \textbf{0.083} & 0.924 \(\pm\) 0.091 & 0.900 \(\pm\) 0.042  \\
    DiTabDDPM (5 discr. feat.)  & 0.069 & 0.954 \(\pm\) 0.073 & 0.883 \(\pm\) 0.030  \\
    DiTabDDPM (2 discr. feat.)  & 0.040 & 0.976 \(\pm\) 0.065 & 0.910 \(\pm\) 0.051  \\
    \midrule
    TabDDPM (3066 feat., all) & 0.048 & 0.912 \(\pm\) 0.104 & 0.890 \(\pm\) 0.036  \\
    TabDDPM (10 discr. feat.) & 0.016 & 0.954 \(\pm\) 0.073 & 0.903 \(\pm\) 0.034  \\
    TabDDPM (5 discr. feat.)  & 0.011 & 0.988 \(\pm\) 0.019 & 0.909 \(\pm\) 0.042  \\
    TabDDPM (2 discr. feat.)  & 0.005 & 0.974 \(\pm\) 0.065 & \textbf{0.913} \(\pm\) 0.039  \\
    \midrule
    VAE (baseline) & 0.0 & \textbf{0.999} \(\pm\) 0.0 & 0.360 \(\pm\) 0.165  \\
    \bottomrule
  \end{tabular}
\end{table}

\begin{table}[htbp]
  \centering
  \caption{Comparison of explainers in terms of predictive faithfulness (PF) and ground-truth faithfulness (GF).}
  \label{tab:table-gnn-faithfulness}
  \tiny
  \setlength{\tabcolsep}{0.9pt}
  \begin{tabular}{ @{}>{\arraybackslash}m{0.69in}>
    {\centering\arraybackslash}m{0.15in}>{\centering\arraybackslash}m{0.15in}>
    {\centering\arraybackslash}m{0.15in}>{\centering\arraybackslash}m{0.15in}>
    {\centering\arraybackslash}m{0.15in}>{\centering\arraybackslash}m{0.15in}>
    {\centering\arraybackslash}m{0.15in}>{\centering\arraybackslash}m{0.15in}>
    {\centering\arraybackslash}m{0.15in}>{\centering\arraybackslash}m{0.15in}>
    {\centering\arraybackslash}m{0.15in}>{\centering\arraybackslash}m{0.15in}>
    {\centering\arraybackslash}m{0.15in}>{\centering\arraybackslash}m{0.15in}@{}}
    \toprule
    \textbf{Approaches} & \multicolumn{2}{c}{\textbf{\textsc{DBLP}}} & \multicolumn{2}{c}{\textbf{\textsc{IMDB}}}&\multicolumn{2}{c}{\textbf{\textsc{MUTAG}}} & \multicolumn{2}{c}{\textbf{\textsc{BA-Shapes}}} & \multicolumn{2}{c}{\textbf{\textsc{Tree-Cycle}}} & \multicolumn{2}{c}{\textbf{\textsc{Tree-Grids}}} & \multicolumn{2}{c}{\textbf{\textsc{BA-3Motif}}}\\
    \cmidrule(lr){2-3} \cmidrule(lr){4-5} \cmidrule(lr){6-7} \cmidrule(lr){8-9} \cmidrule(lr){10-11} \cmidrule(lr){12-13} \cmidrule(lr){14-15}
    & PF$\uparrow$ & GF$\uparrow$ & PF$\uparrow$ & GF$\uparrow$ & PF$\uparrow$ & GF$\uparrow$ & PF$\uparrow$ & GF$\uparrow$ & PF$\uparrow$ & GF$\uparrow$ & PF$\uparrow$ & GF$\uparrow$ & PF$\uparrow$ & GF$\uparrow$ \\
    \toprule
    \textsc{XGNN} & 0.243 & 0.549 & 0.268 &0.627 & 0.952 & 0.367 & 0.275 & 0.292 & 0.566 & 0.288 & \textbf{0.521} & 0.608 & 0.333 & 0.296 \\
    \textsc{GNNInterpreter} & 0.561 & 0.729 & 0.394 & 0.661 & 0.780 & 0.324 & 0.263 & 0.654 & \textbf{0.601} & 0.565 & 0.520 & 0.667 & 0.485 & 0.453 \\
    \textsc{D4Explainer} & 0.306 & 0.418 & 0.379 & 0.519 & 0.433 & 0.294 & \textbf{0.343} & 0.368 & 0.519 & 0.396 & 0.510 & 0.760 & \textbf{0.666} & 0.213\\
    \midrule
    \textsc{DiGNNExplainer-GCN} & 0.641 & 0.800 & 0.410 & 0.850 & \textbf{1.0} & \textbf{0.425} & 0.315 & \textbf{0.766} & 0.525 & \textbf{0.716} & 0.507 & \textbf{0.893} & 0.432 & \textbf{0.639} \\  
     \textbf{\textsc{DiGNNExplainer}} & \textbf{0.999} & \textbf{0.882} & \textbf{0.974} & \textbf{0.913} & \textbf{1.0} & \textbf{0.425} & 0.315 & \textbf{0.766} & 0.525 & \textbf{0.716} & 0.507 & \textbf{0.893} & 0.432 & \textbf{0.639} \\
    \bottomrule
  \end{tabular}
\end{table}

\subsection{Realism of Graph Structure}

In this section, we identify the approach for generating the most realistic explanation graphs. We compare the graph distributions of the explanations generated by DiGNNExplainer, VAE, and the baseline explainers (Table~\ref{tab:table-mmd-main}) with real graphs.

\paragraph{Implementation} (1)~We generate graphs that serve as explanations. In our experiments, we restrict the explanation graph size to a range of 5 to 15 nodes to ensure human-intelligible explanation graphs. (2)~We compare the average MMD (Table~\ref{tab:table-mmd-main}) between the explanation graphs of each class and the real graphs for node type distribution, node degree, clustering coefficient, and spectrum distributions for 50 graphs. 

\paragraph{Results} DiGNNExplainer achieves the highest similarity for node degree distribution across all datasets except \textsc{BA-Shapes}, and the highest similarity for node type distribution (NTD) and clustering coefficient (Clus.) for the real-world datasets (Table~\ref{tab:table-mmd-main}). 

\paragraph{Interpretation} DiGNNExplainer generates the most realistic graphs, preserving the distributions of graphs for all datasets and node types for heterogeneous datasets. The homogeneous datasets have a single node type, hence the comparison always yields a score of 0.0.

\subsection{Realism of Node Features}

In this section, we identify the approach for generating the most realistic node features using cosine similarity (Node feat. (cos) in Table~\ref{tab:table-evaluation-feature-selection}). We compare DiTabDDPM and TabDDPM in terms of node feature realism for various node feature sizes. The notations used for our feature selection experiments in the Table~\ref{tab:table-evaluation-feature-selection} are defined as: (i) DiTabDDPM---discrete (bag-of-words) node features generated using DiTabDDPM and continuous node features generated using TabDDPM (ii) TabDDPM---all node features generated using TabDDPM (iii) DiTabDDPM (8 discr. feat.) indicates that the node to be classified has a feature size of 8.

\paragraph{Implementation} 
(1) For discrete node features, we use one-hot encoding of word frequencies, i.e., in the range of 0 to 5 for paper nodes. 
(2) For VAE, we learn the graphs along with node features, which include all features, i.e., 50 for \textsc{DBLP} and 3066 for \textsc{IMDB}. In VAE, we average the cosine similarity of the node features of the node to be classified for 50 graphs.

\paragraph{Results} Cosine similarity (Node feat.) results are higher
for DiTabDDPM compared to TabDDPM, with the highest score of 0.397 for \textsc{DBLP}.

\paragraph{Interpretation} DiTabDDPM, which learns a discrete space for discrete features, generates more realistic node features compared to the original TabDDPM, which learns a continuous space for discrete features. 

\subsection{Faithfulness of Explanations}

The goal of this section is to determine which approach produces the most faithful explanations in terms of (1)~predictive faithfulness and (2)~ground-truth faithfulness.

In Table~\ref{tab:table-gnn-faithfulness}, we compare the faithfulness of our approach against existing model-level explanation approaches that do not utilize actual node features. For this comparison, we use the original homogeneous datasets from the original baseline explainer implementations. Across all settings, we keep the generated graphs and ground-truth motifs fixed and vary only the node feature sizes.

In Table~\ref{tab:table-evaluation-feature-selection}, we train DiTabDDPM, TabDDPM, and VAE to generate heterogeneous explanation graphs with varying node feature sizes (as in \cite{Dong2021MobileGCN}). We assess their realism by comparing the generated features to the ground truth features (Node feat. (cos)). Moreover, we evaluate the faithfulness of the resulting explanations by applying a GNN.

\paragraph{Implementation}
We use the original model architecture for each baseline. Since the baseline explainers do not support real-valued node features or heterogeneous message passing, we convert the heterogeneous datasets \textsc{DBLP} and \textsc{IMDB} into homogeneous graphs with one-hot node types---the only setting in which all baselines can be executed. In DiGNNExplainer-GCN, we train a GCN on the converted homogeneous versions of \textsc{DBLP} and \textsc{IMDB}, as well as on \textsc{MUTAG}, and synthetic datasets,  whereas in DiGNNExplainer, we train GraphSAGE or HAN on the original heterogeneous versions of \textsc{DBLP} and \textsc{IMDB}, and use a GCN for the other datasets. To ensure a fair comparison, following DiGNNExplainer, each baseline utilizes a different model for each graph size. We use early stopping for all trained GNNs to prevent overfitting. 

\paragraph{Results} 
In Table~\ref{tab:table-gnn-faithfulness}, DiGNNExplainer, which utilizes real node features, achieves superior predictive and ground-truth faithfulness compared to all baseline explainers (on \textsc{DBLP}, \textsc{IMDB}, and \textsc{MUTAG}). DiGNNExplainer also achieves the best ground-truth faithfulness for the synthetic datasets. On \textsc{DBLP} and \textsc{IMDB}, DiGNNExplainer-GCN performs worse than DiGNNExplainer but still outperforms all baseline explainers in both metrics.

In Table~\ref{tab:table-evaluation-feature-selection}, for \textsc{DBLP}, DiTabDDPM with 50 features achieves the best explanation graphs with a maximum ground-truth faithfulness of 0.882. For \textsc{IMDB}, we obtain similarly high maximum scores of 0.91 for both DiTabDDPM (2 discr. feat.) and TabDDPM (2 discr. feat.).
The key advantage of using ground-truth faithfulness for evaluation is robustness to spurious, out-of-distribution explanations: high GNN scores can arise from out-of-distribution graphs that would never occur in practice, whereas high ground-truth faithfulness ensures that the explanatory motifs align with the true building blocks of the data and therefore better reflect real-world behavior.

\paragraph{Interpretation} (1)~The explanations generated by DiGNNExplainer for our real-world datasets (Table~\ref{tab:table-gnn-faithfulness}) are faithful, as they achieve both high predictive faithfulness and ground-truth faithfulness.
(2)~Our baseline explainers yield a lower ground-truth faithfulness due to their inability to model heterogeneous structures and lack of node features.
(3)~The low predictive faithfulness of DiGNNExplainer on the synthetic datasets arises because the 3-layer GCN is heavily biased toward predicting the base node class.
As a result, only one out of the four classes (base, house, cycle, grid) is predicted correctly in many cases, which leads to a low average predictive faithfulness across classes.
(4) The low ground-truth faithfulness of VAE (Table~\ref{tab:table-evaluation-feature-selection}) suggests VAE-based graph generation is unable to preserve the underlying heterogeneous structures.
(5) The high predictive faithfulness of VAE for \textsc{IMDB} can be explained by the fact that the GNN is trained with all features of the dataset, where all node types except movie have continuous features, and the continuous features generated using VAE are more realistic.
(6) Realistic discrete features generated by learning a discrete space in DiTabDDPM result in highly faithful explanations, but we also observe that the node features for faithful explanations are not always the most realistic, i.e., do not have the highest cosine similarity values.
(7) On \textsc{DBLP} (Table~\ref{tab:table-evaluation-feature-selection}), using a larger feature size produces both high predictive faithfulness and ground-truth faithfulness scores, indicating that the GNN model can be trusted. Overall, observations (5)--(7) indicate that the node feature generation technique substantially influences the quality of explanation graphs. 

\subsection{Experiment on Feature Selection}

In Table~\ref{tab:table-evaluation-feature-selection}, we select varying numbers of features (2, 4, 8, 50 for \textsc{DBLP} and 2, 5, 10, 3066 for \textsc{IMDB}) using Variance Threshold. As an alternative, Table~\ref{tab:table-additional-frequency} in the Appendix reports results using the top $k$ most frequent features as a selection method. Irrespective of the feature selection method, we obtain good explanation graphs, for example, with 50 features using DiTabDDPM on \textsc{DBLP}.

\subsection{Ablation Study}

We conduct an ablation study to investigate the effect of using multiple DiGress models for different graph sizes versus a fixed-size graph model. We report predictive faithfulness and ground-truth faithfulness scores for three settings:
(1)~a single diffusion model trained on a fixed graph size (12 for \textsc{DBLP} and 7 for \textsc{IMDB}; Table~\ref{tab:table-ablation-single-vs-multiple-sizes}, left),
(2)~a single diffusion model trained on multiple graph sizes (10--15; Table~\ref{tab:table-ablation-single-vs-multiple-sizes}, right), and
(3)~our main setting, which trains a separate DiGress model for each graph size (10--15; Table~\ref{tab:table-evaluation-feature-selection}).
Overall, we observe that training one model per graph size yields the best performance. In particular, for \textsc{DBLP}, training a single model on multiple sizes results in a lower ground-truth faithfulness of 0.730, compared to 0.882 in the main experiments. 

\begin{table}[tb]
  \centering
  \scriptsize
  \caption{Comparison of predictive faithfulness (PF) and ground-truth faithfulness (GF) for ablation study using a fixed graph size versus using multiple graph sizes by training a single Diffusion model.}
  \label{tab:table-ablation-single-vs-multiple-sizes}
  \setlength{\tabcolsep}{5pt}
  \begin{tabular}{@{}lcccc@{}}
    \toprule
    \textbf{Approach} & \multicolumn{2}{c}{\textbf{Fixed}} & \multicolumn{2}{c}{\textbf{Multiple}}\\
    \cmidrule(lr){2-3}\cmidrule(lr){4-5}
    & PF $\uparrow$ & GF $\uparrow$ & PF $\uparrow$ & GF $\uparrow$ \\
    \midrule
    \multicolumn{5}{c}{DBLP} \\ 
    \midrule
    DiTabDDPM (50 feat.)        & \textbf{0.999} \(\pm\) 0.0 & 0.717 \(\pm\) 0.040          & \textbf{0.999} \(\pm\) 0.0 & 0.730 \(\pm\) 0.036 \\
    DiTabDDPM (8 discr. feat.)  & 0.872 \(\pm\) 0.102        & 0.787 \(\pm\) 0.037          & 0.971 \(\pm\) 0.030        & 0.777 \(\pm\) 0.043 \\
    DiTabDDPM (4 discr. feat.)  & 0.770 \(\pm\) 0.141        & \textbf{0.792} \(\pm\) 0.043 & 0.770 \(\pm\) 0.110        & \textbf{0.829} \(\pm\) 0.071 \\
    DiTabDDPM (2 discr. feat.)  & 0.925 \(\pm\) 0.073        & 0.735 \(\pm\) 0.066          & 0.938 \(\pm\) 0.071        & 0.812 \(\pm\) 0.057 \\
    \midrule
    \multicolumn{5}{c}{IMDB} \\ 
    \midrule
    DiTabDDPM (3066 feat., all) & 0.976 \(\pm\) 0.058 & 0.860 \(\pm\) 0.051          & \textbf{0.983} \(\pm\) 0.021 & 0.853 \(\pm\) 0.026 \\
    DiTabDDPM (10 discr. feat.) & 0.967 \(\pm\) 0.056 & 0.863 \(\pm\) 0.069          & 0.972 \(\pm\) 0.050          & 0.853 \(\pm\) 0.030 \\
    DiTabDDPM (5 discr. feat.)  & 0.951 \(\pm\) 0.087 & \textbf{0.886} \(\pm\) 0.022 & 0.956 \(\pm\) 0.065          & \textbf{0.883} \(\pm\) 0.026 \\
    DiTabDDPM (2 discr. feat.)  & 0.902 \(\pm\) 0.099 & 0.836  \(\pm\) 0.037         & 0.934 \(\pm\) 0.100          & 0.876  \(\pm\) 0.033 \\
    \bottomrule
  \end{tabular}
\end{table}

\subsection{Runtime Analysis} 

We compare our runtime against the baselines XGNN, GNNInterpreter, and D4Explainer (Table~\ref{tab:table-runtime-all}). Unlike graph generation techniques that primarily assess scalability with respect to increasing graph size, we evaluate scalability with respect to the size of the original dataset. All measurements report the total time needed for all graph sizes. Compared to D4Explainer, DiGNNExplainer is computationally faster on the large datasets \textsc{DBLP} and \textsc{IMDB}. The runtime of DiGNNExplainer remains approximately constant for both small (Tree-Cycle, 871 nodes) and large (\textsc{DBLP}, 26128 nodes) datasets since our approach samples only small explanation graphs.

Moreover, we break down the runtime of our approach by dataset and by component (Table~\ref{tab:table-runtimes-dignn}) showing that the dominant cost of DiGNNExplainer is graph generation training. For node classification datasets, we additionally run the Forest Fire Sampler (FFS) to extract training subgraphs for each explanation graph size (e.g., 10--15, i.e., six sizes). For each size, we sample $k=200$ subgraphs, resulting in 1,200 sampled graphs in total. We empirically observe that the runtime grows from the small synthetic datasets to \textsc{IMDB} and \textsc{DBLP}, reflecting higher sampling overhead on larger underlying graphs. Next, we train DiGress on these sampled graphs. The Graph Generation Training (GGT) and Graph Generation After Training (GGAT) steps scale linearly with the number of sampled and generated graphs, respectively, as well as with the number of diffusion steps. Additionally, we train a model for node feature generation, where the runtimes for Node Feature Generation Training/After Training (NFGT/NFGAT) are only incurred for datasets with generated node attributes (here: \textsc{DBLP} and \textsc{IMDB}) and depend on the feature dimensionality and model hyperparameters. The Post-Processing (PP) step checks whether the generated graph is consistent with the metagraph. Although graph matching can be computationally hard in the worst case, this step is fast in practice for our small explanation graphs, as confirmed by our empirical runtimes. Finally, the total runtime for Graph Selection (GS), including the required GNN predictions scales linearly with the number of graphs and with the dimensionality of the node features.

\begin{table}[tb]
  \centering
  \caption{Runtime (in secs.) comparisons for all graph sizes with baseline approaches.}
  \label{tab:table-runtime-all}
  \setlength{\tabcolsep}{1.1pt}
  \scriptsize
  \begin{tabular}{@{}l@{\hskip -1pt}rrrrrrr@{}}
    \toprule
    \textbf{Approaches} & \textbf{\textsc{DBLP}}  & \textbf{\textsc{IMDB}} & \textbf{\textsc{MUTAG}} & \textbf{\textsc{BA-Shapes}} & \textbf{\textsc{Tree-Cycle}} & \textbf{\textsc{Tree-Grids}} & \textbf{\textsc{BA-3Motif}} \\ \midrule
    XGNN   & 1102 & 363 & 27 & 351 & 119 & 155 & 79 \\ 
    GNNInterpreter & 79 & 102 & 5 & 143 & 73 &73 &13 \\
    D4Explainer & 39175 & 19406 & 658 & 5703 & 659 & 822 & 801 \\ 
    \textbf{DiGNNExplainer} & 5875 & 4702 & 6910 & 4637 & 5192 & 5501 & 1992 \\ 
    \bottomrule
  \end{tabular}
\end{table}

\begin{table}[tb]
  \centering
  \setlength{\tabcolsep}{7.8pt}
  \scriptsize
  \caption{Breakdown of runtime (in secs.) for each component (Forest Fire Sampler (FFS), Graph Generation Training (GGT), Graph Generation After Training (GGAT), Node Feature Generation Training (NFGT), Node Feature Generation After Training (NFGAT), Postprocessing (PP), Graph selection (GS)) of DiGNNExplainer for all datasets.}
  \label{tab:table-runtimes-dignn}
    \begin{tabular}{@{}lrrrrrrr@{}}
      \toprule
      &\textbf{FFS} &\textbf{GGT} &\textbf{GGAT}&\textbf{NFGT} &\textbf{NFGAT}&\textbf{PP}&\textbf{GS} \\ 
      \midrule
      \textsc{DBLP}       & 7.92 & 5297.4 & 255.5   & 252.10 & 41.01 &  20.48 & 0.72 \\ 
      \textsc{IMDB}       & 5.08 & 4054.8 & 192.3   & 108.73 & 36.46 & 304.57 & 0.61 \\ 
      \textsc{MUTAG}      &   -- & 6699.4 & 210.2   &     -- &    -- &     -- & 0.83 \\
      \textsc{BA-Shapes}  & 3.80 & 4380.5   & 252.6 &     -- &    -- &     -- & 0.69 \\ 
      \textsc{Tree-Cycle} & 3.65 & 4926.1   & 253.8 &     -- &    -- &     -- & 0.91 \\ 
      \textsc{Tree-Grids} & 3.75 & 5238.4   & 258.5   &     -- &    -- &     -- & 0.79 \\ 
      \textsc{BA-3Motif}  &   -- & 1935.6   &  49.8 &     -- &    -- &     -- & 1.49 \\
      \bottomrule
    \end{tabular}
\end{table}

\subsection{Case Study}

In this section, we present case study examples of explanation graphs with node features. For each class, we visualize the explanation graph that maximizes the GNN’s predicted probability for that class and corresponding feature distributions. Concretely, we show the feature vector of the author node with the highest probability and the continuous distribution of all term nodes.

\begin{figure}[tb]
\centering
\begin{minipage}{\columnwidth}
  \begin{subfigure}{0.23\linewidth}
    \centering
    \includegraphics[width=1\linewidth]{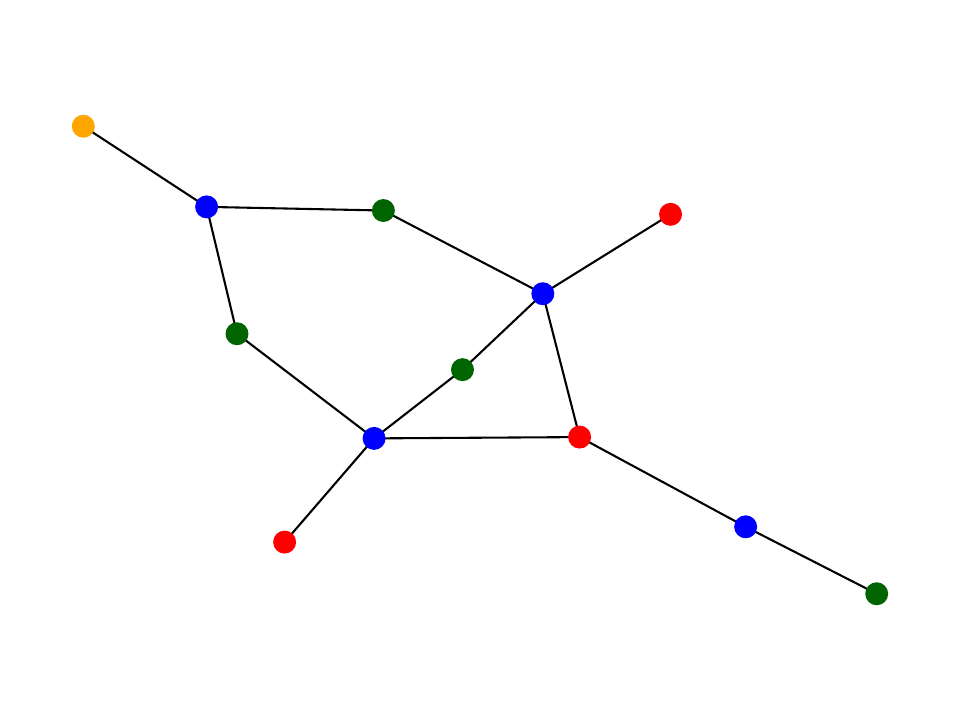}
    \small
    \begin{tabular}{lc}
      Nodes & 12 \\
      Prob. & 1.0 \\
    \end{tabular}
    \caption{class 0}
	\end{subfigure}%
    \hfill%
    \begin{subfigure}{0.23\linewidth}%
      \centering
    \includegraphics[width=1\linewidth]{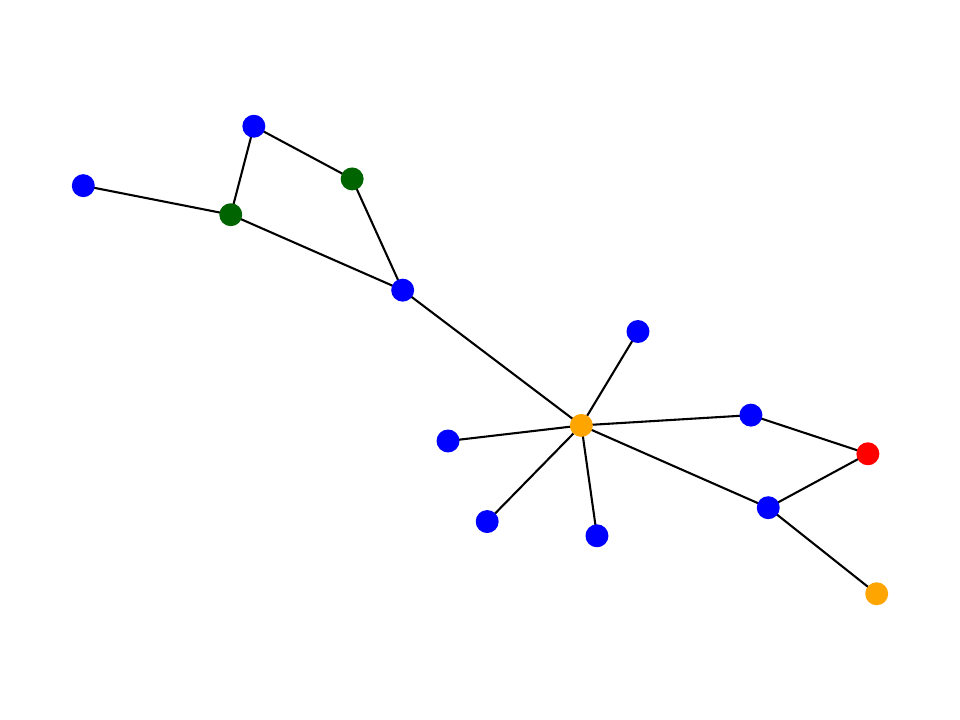}
      \small
      \begin{tabular}{lc}
        Nodes & 15 \\
        Prob. & 1.0 \\
      \end{tabular}
      \caption{class 1}
    \end{subfigure}%
    \hfill%
    \begin{subfigure}{0.23\linewidth}
      \centering
      \includegraphics[width=1\linewidth]{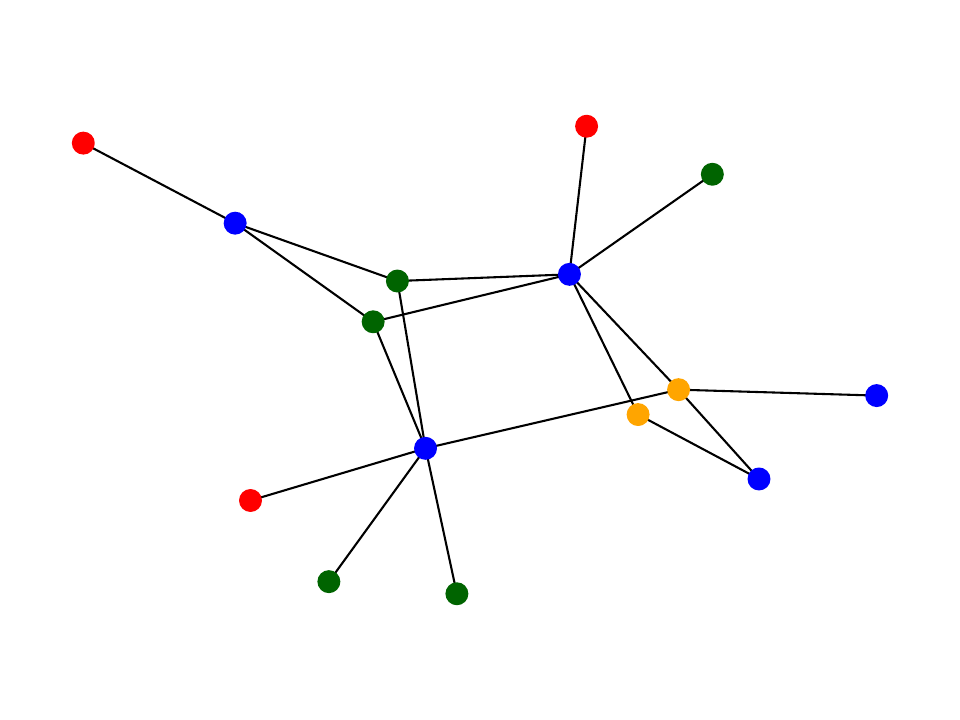}
      \small
      \begin{tabular}{lc}
        Nodes & 14 \\
        Prob. & 1.0 \\
      \end{tabular}
      \caption{class 2}
    \end{subfigure}%
    \hfill%
    \begin{subfigure}{0.23\linewidth}
      \centering
      \includegraphics[width=1\linewidth]{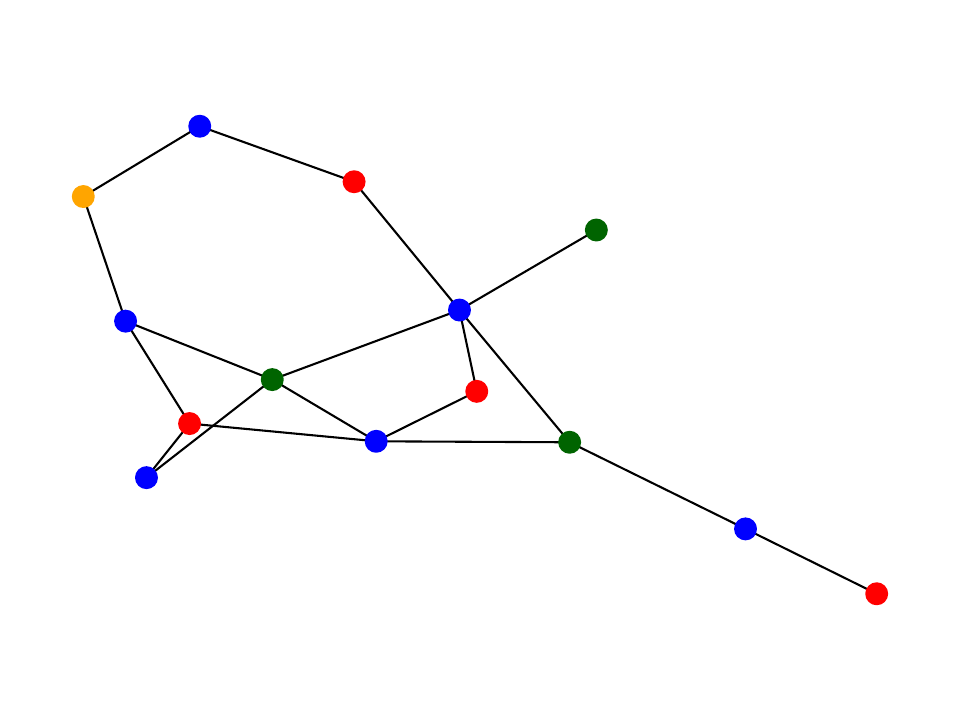}
      \small
      \begin{tabular}{lc}
        Nodes & 14 \\
        Prob. & 1.0 \\
      \end{tabular}
      \caption{class 3}
   \end{subfigure}
   \caption{Explanation graphs for each class of author nodes of the DBLP dataset that maximizes the prediction of the class. Node colors indicate types: paper (blue), author (red), term (dark green), and conference (orange).}
   \label{fig:dblp-max-pred}
\end{minipage}
\par\vspace{0.5cm}
\begin{minipage}{\columnwidth}%
  \begin{subfigure}[t]{0.22\linewidth}%
    \centering%
    \includegraphics[width=1\linewidth]{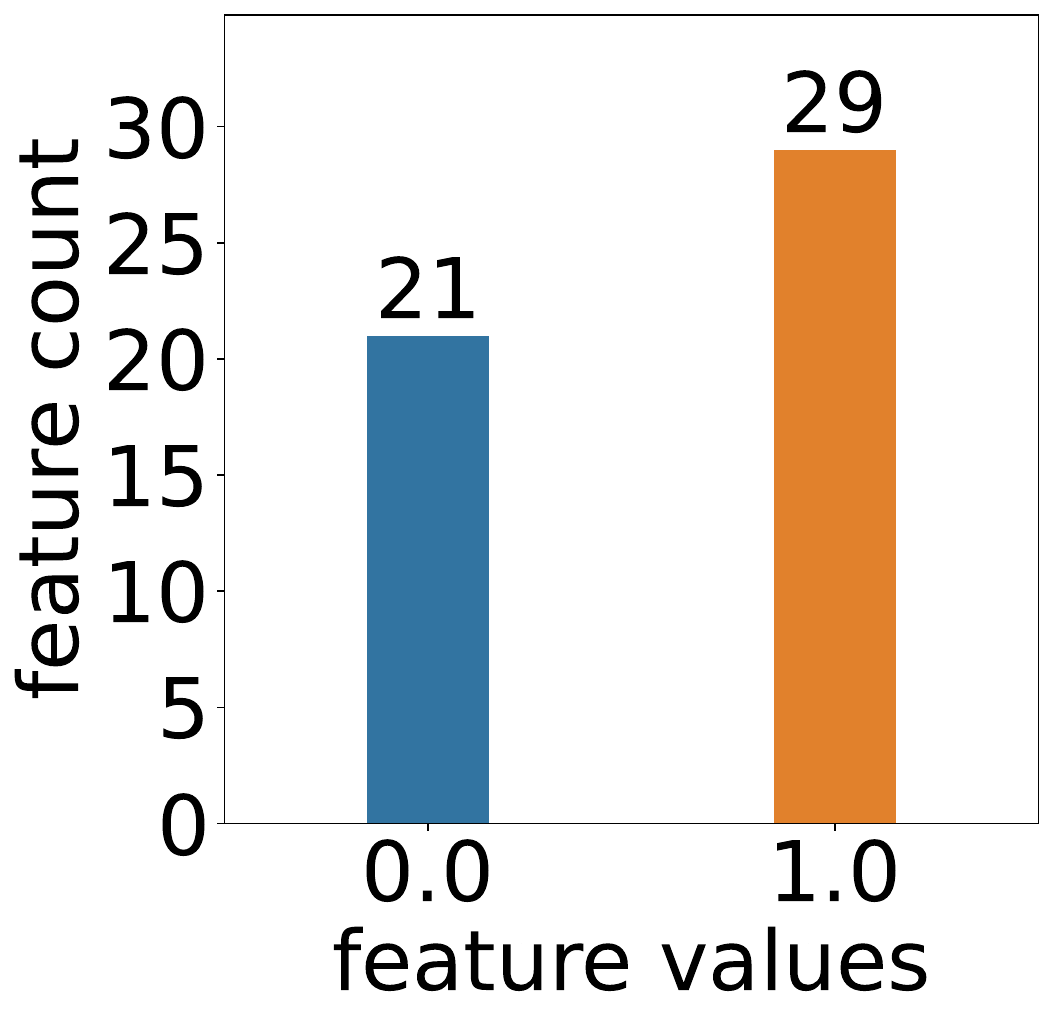}%
    \caption{class 0}%
  \end{subfigure}%
  \hfill%
  \begin{subfigure}[t]{0.22\linewidth}
    \centering
    \includegraphics[width=\linewidth]{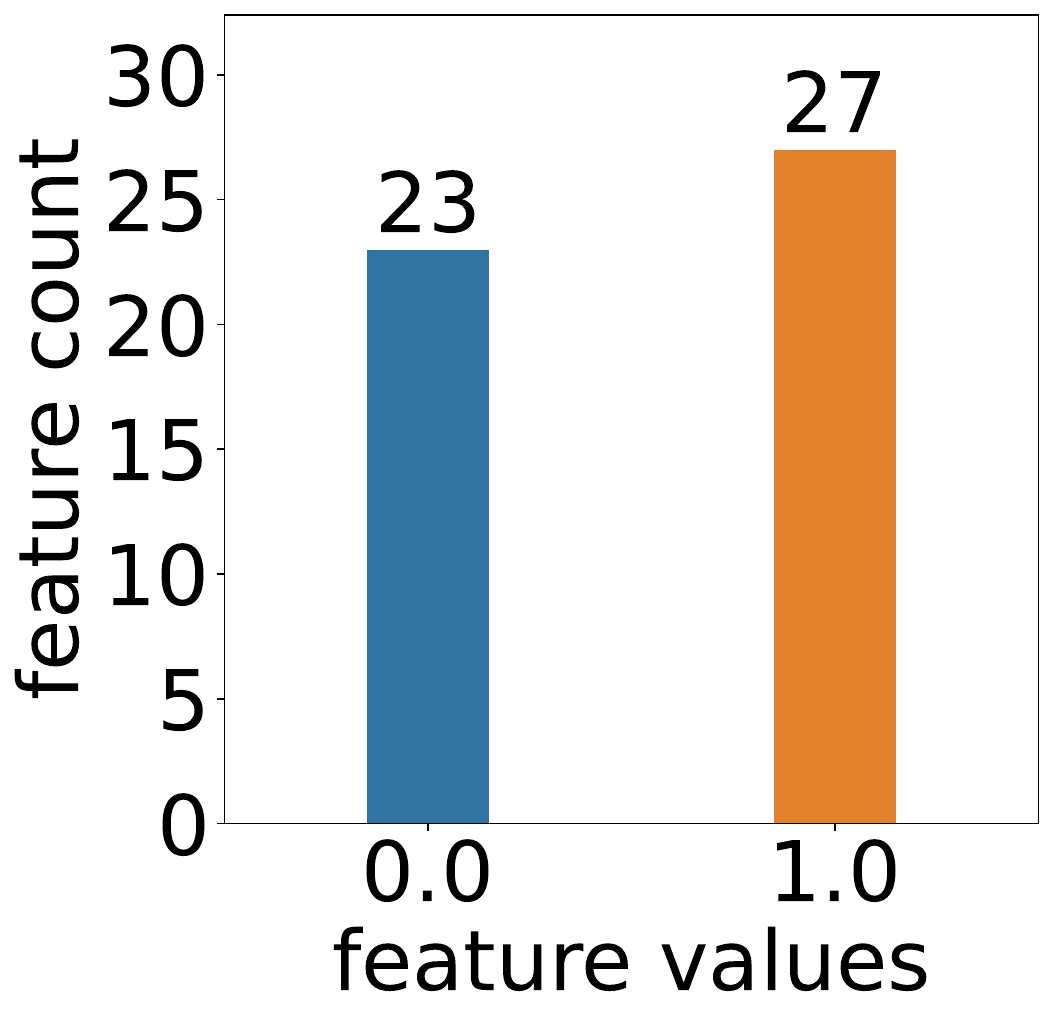}
    \caption{class 1}
  \end{subfigure}%
  \hfill%
  \begin{subfigure}[t]{0.22\columnwidth}
    \centering
    \includegraphics[width=\linewidth]{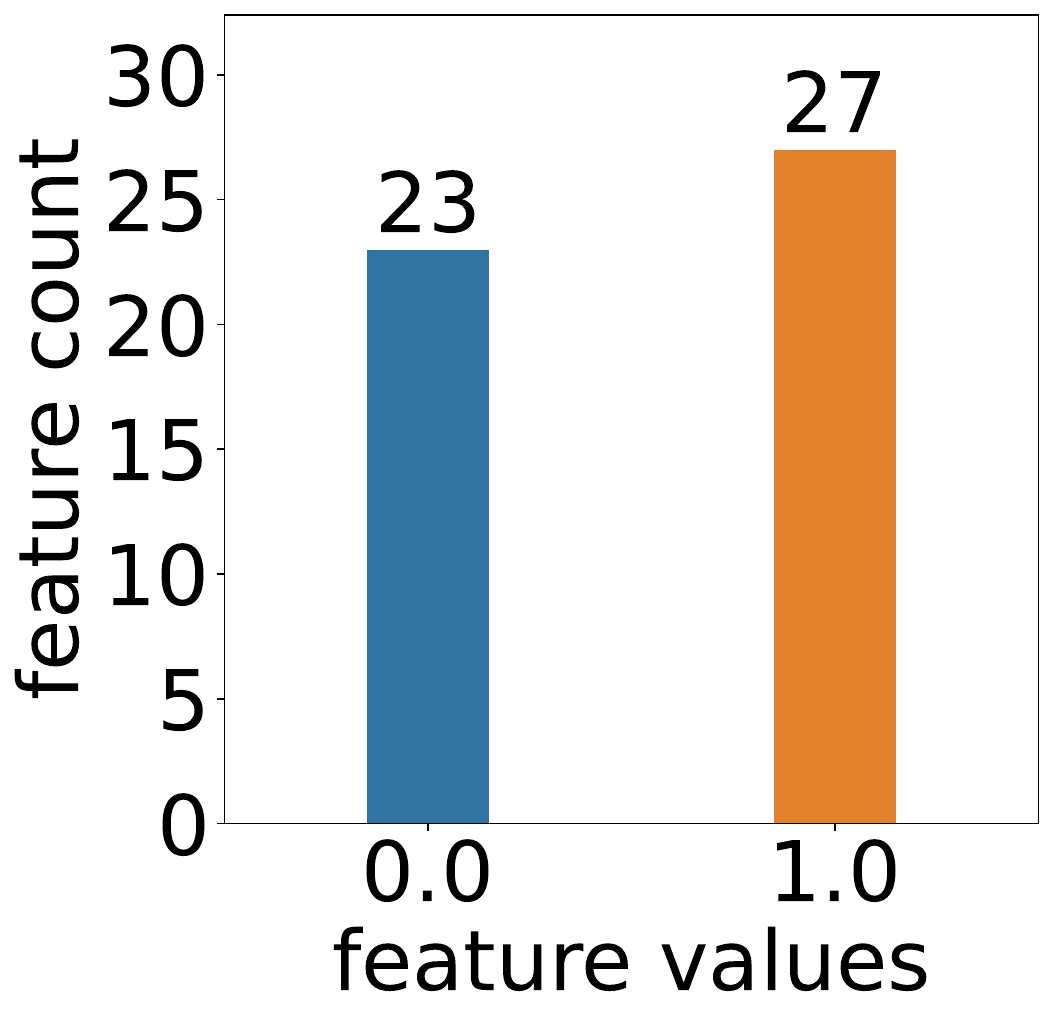}
    \caption{class 2}
  \end{subfigure}%
  \hfill%
    \begin{subfigure}[t]{0.22\columnwidth}
    \centering
    \includegraphics[width=\linewidth]{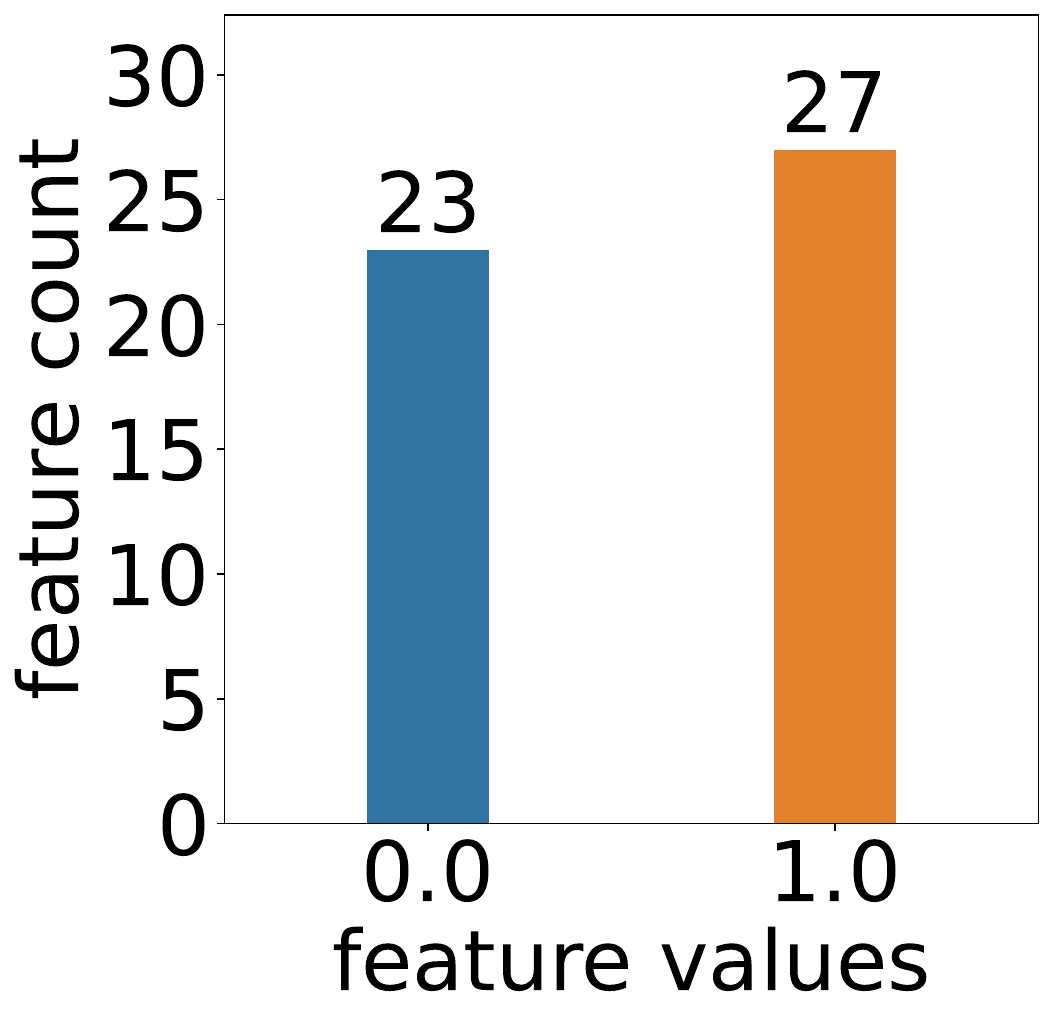}
    \caption{class 3}
  \end{subfigure}
  \caption{Frequency distribution plots visualizing the synthetic bag-of-words node feature vector of the author node of the explanation graph that is classified with maximum probability for a class.}
  \label{fig:dblp-author-node-feature}
\end{minipage}%
\par\vspace{0.5cm}
\begin{minipage}{\columnwidth}%
  \begin{subfigure}[t]{0.22\linewidth}%
    \centering%
    \includegraphics[width=1\linewidth]{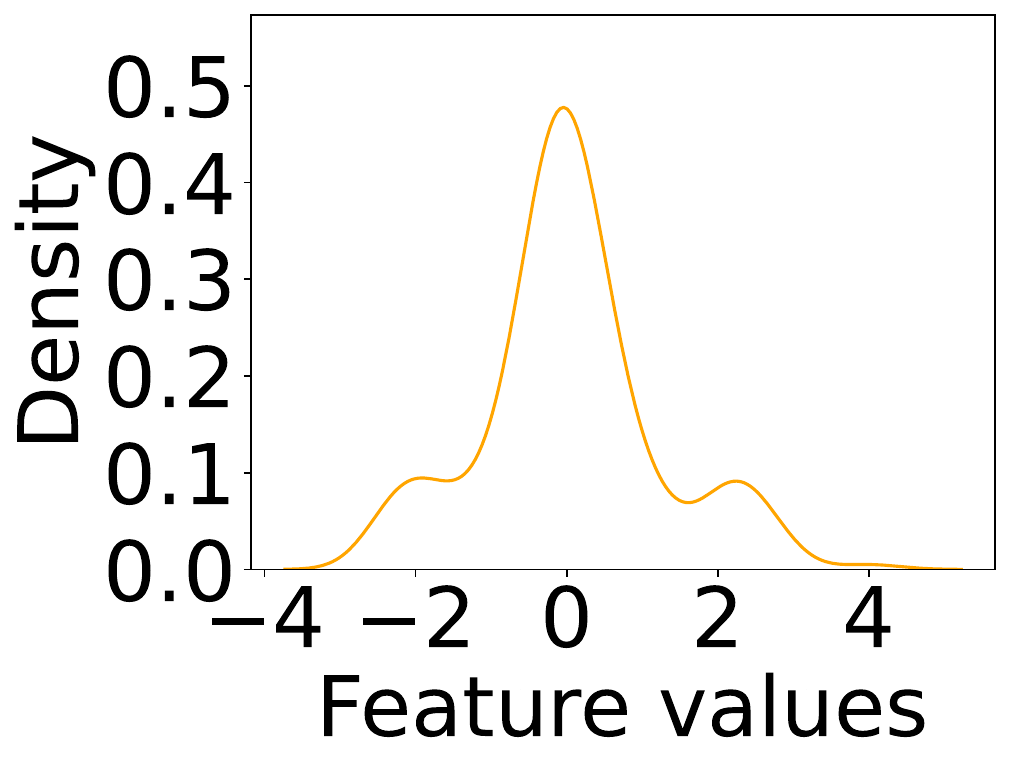}%
    \caption{class 0}%
  \end{subfigure}%
  \hfill%
  \begin{subfigure}[t]{0.22\linewidth}
    \centering
 \includegraphics[width=1\linewidth]{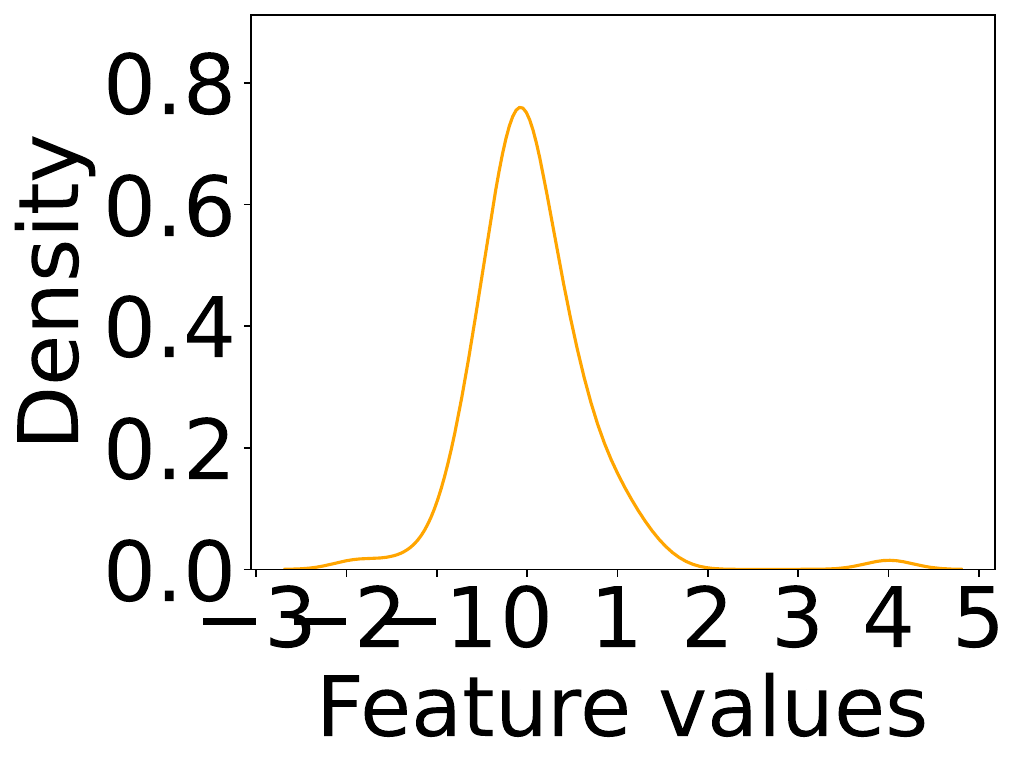}
    \caption{class 1}
  \end{subfigure}%
  \hfill%
  \begin{subfigure}[t]{0.22\linewidth}
    \centering
    \includegraphics[width=\linewidth]{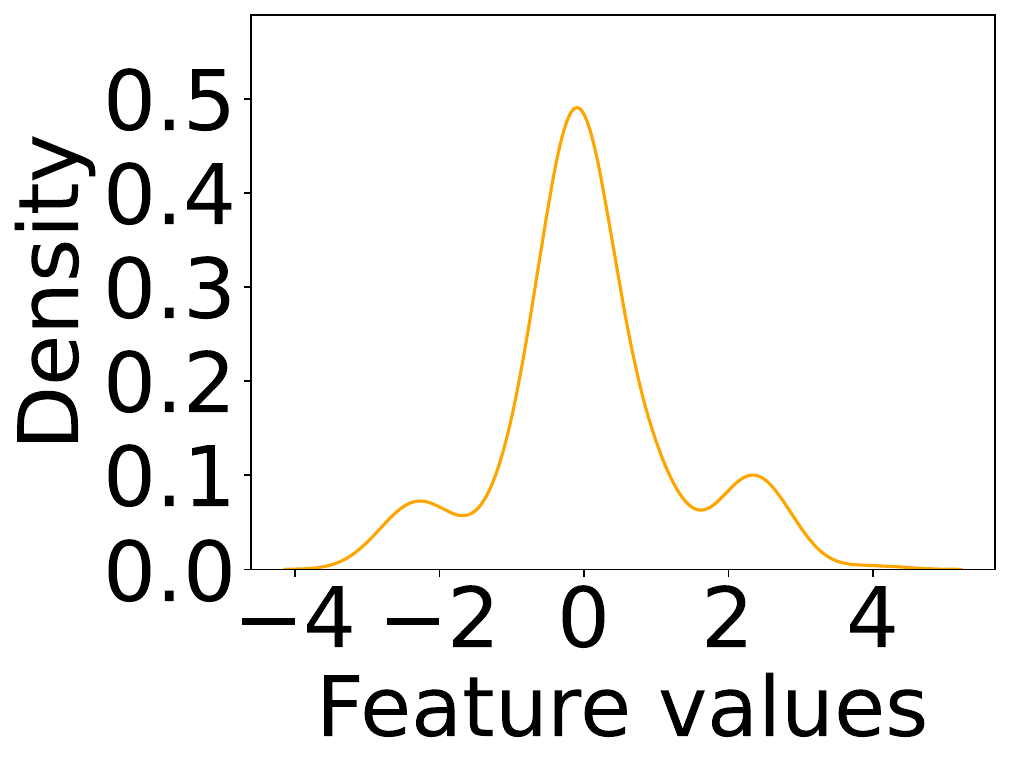}
    \caption{class 2}
  \end{subfigure}%
  \hfill%
  \begin{subfigure}[t]{0.22\linewidth}
    \centering
    \includegraphics[width=\linewidth]{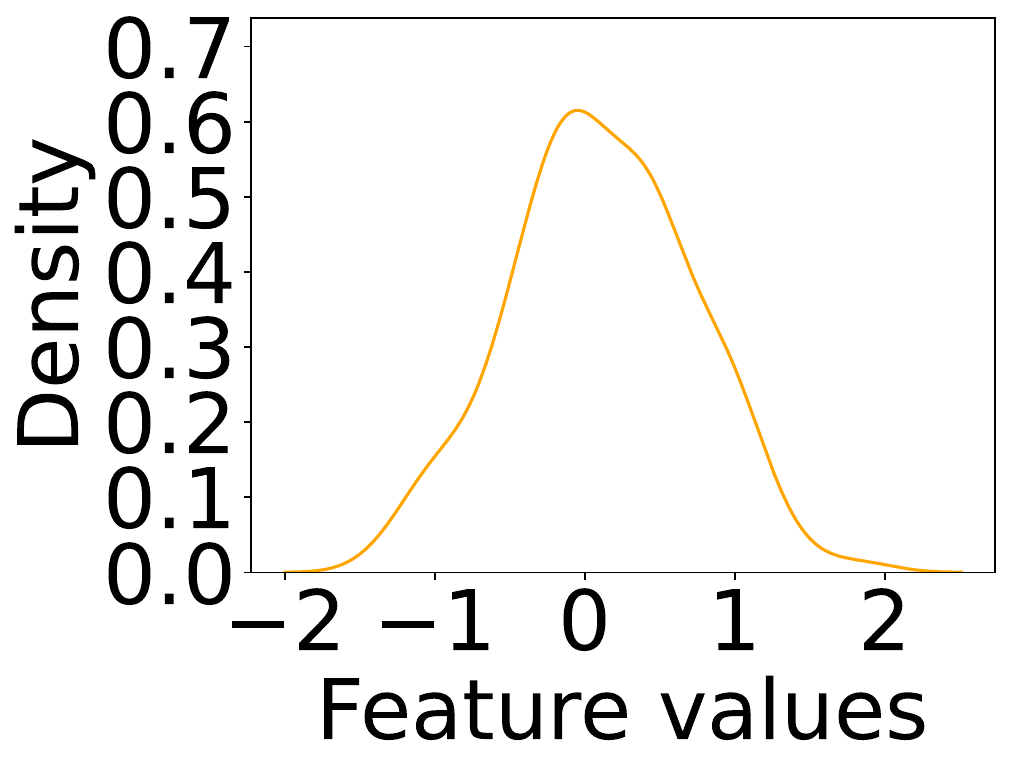}
    \caption{class 3}
  \end{subfigure}
  \caption{Density plots of the continuous term-node feature values in the explanation graphs. For each class, we show the feature-value distribution of the term nodes in the explanation graph that maximizes the predicted probability for that class.}
  \label{fig:dblp-term-node-feature}
\end{minipage}
\end{figure}

Figures~\ref{fig:dblp-max-pred}, \ref{fig:dblp-author-node-feature}, and \ref{fig:dblp-term-node-feature} show the explanation graphs and node features for the DiTabDDPM (50 discr. feat.) approach for \textsc{DBLP}. The size of the explanation graphs is between 12 and 15. All author classes are predicted with a probability of 1.0. The ground-truth faithfulness obtained for author classes is 0.8275. Figure~\ref{fig:dblp-author-node-feature} shows the distribution of the bag-of-words features (0.0/1.0) of the author node predicted with maximum probability for each class. The term nodes have continuous features (Figure~\ref{fig:dblp-term-node-feature}) with a mean between 0.04 and 0.11 and a standard deviation between 0.60 and 1.22. The explanations for all approaches and all datasets can be found in our evaluation results.%
\footnote{\url{https://github.com/ds-jrg/DiGNNExplainer/tree/main/evaluation}}

\section{Discussion}
We discuss the following improvements and extensions to our approach:
(1)~Our approach separates the process of generating explanation graphs into two distinct steps: graph structure generation and node feature generation.
This decoupling allows discrete and continuous features to be learned in their appropriate spaces, which performs better than joint continuous-space modeling. 
However, it does not capture advanced feature-structure dependence. 
We explored modeling such dependencies in DiTabDDPM and TabDDPM, but observed reduced feature realism, as quantified by cosine similarity.
Therefore, we present our main results without considering advanced feature-structure dependence. In our approach, structural information is retained through node types, and feature-structure variation is captured via separate feature models per node type.
(2)~We obtain consistent runtime for both small and large datasets, making our approach applicable to graphs of varying sizes. Runtime could be further reduced in future work by using multiple GPUs simultaneously, as in recent DiGress implementations.
(3)~Training separate models for different graph sizes improves our explanations, but entails a trade-off between faithfulness and computational time. A faster alternative is to train a single diffusion model on multiple graph sizes. However, our ablation study (Table~\ref{tab:table-ablation-single-vs-multiple-sizes}) shows that this yields weaker evaluation results and generates a higher number of disconnected graphs. 
(4)~Our approach is easily extendable to real-world heterogeneous \emph{directed} graphs, since our GNNs (GraphSAGE and HAN) can be trained on directed graphs. Moreover, DiGNNExplainer, which builds upon DiGress, is able to support directed graphs with only minor modifications, i.e., by modifying the way noise is added to the adjacency matrix. 
(5)~We currently obtain one explanation per class, but our implementation also supports multiple explanation graphs per class, i.e., the top $k$ explanations ranked by softmax probabilities.

\section{Conclusion}
DiGNNExplainer is the first method to explain heterogeneous GNNs on a model-level on real-world datasets using explanation graphs with actual node features. Our experiments show that incorporating actual node features significantly improves explanation faithfulness: DiGNNExplainer achieves higher predictive and ground-truth faithfulness than state-of-the-art baselines, while producing explanation graphs whose statistical properties closely match those of real-world graphs. This is enabled by
(1) node feature generation via our novel discrete diffusion model, DiTabDDPM, which enables the synthesis of realistic discrete features; and (2) our discrete diffusion-based approach for modeling heterogeneous graphs, supporting the generation of structurally realistic explanation graphs. Beyond the citation and movie datasets used in our experiments, DiGNNExplainer is broadly applicable to any domain with heterogeneous graph data and structured node features, such as healthcare, finance, or chemistry. Further runtime optimization is left for future work.

\begin{acks}
This work has been supported by the European Union and the Ministry of Economic Affairs, Industry, Climate Action and Energy of the State of North Rhine-Westphalia (MWIKE NRW) under the grant no EFRE-20801043.
\end{acks}

\clearpage

{
\raggedright
\bibliographystyle{ACM-Reference-Format}
\bibliography{bibliography}
}

\appendix

\section{Diffusion Models}

\begin{figure}[h]
  \centering
  \includegraphics[width=\linewidth]{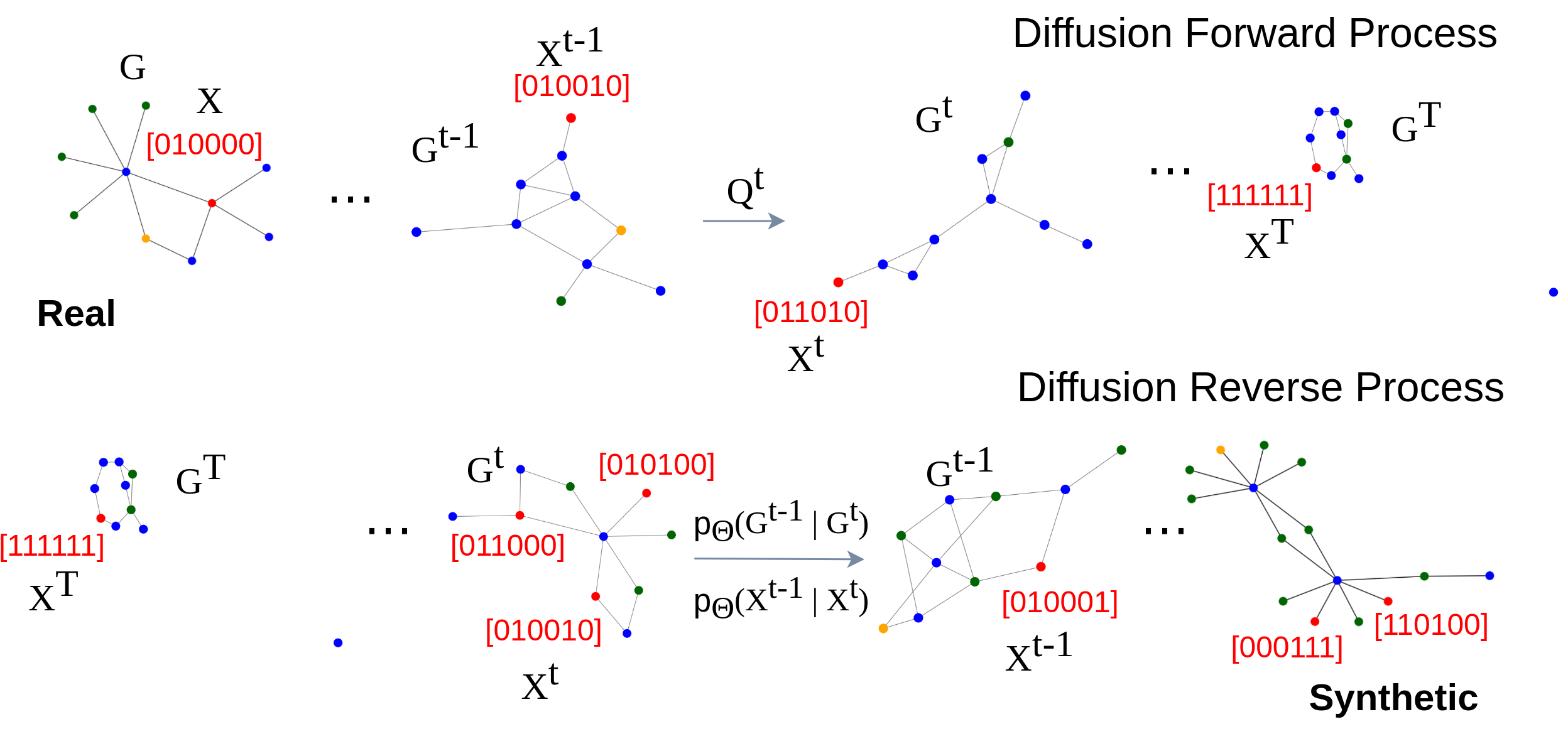}
  \caption{Diffusion steps for generation of synthetic graph and discrete node features.
  }
  \label{fig:diffusion}
\end{figure}

\subsection{DiGress}

DiGress~\cite{Vignac2023DiGress} is a discrete diffusion approach for graph generation, which adds noise through discrete time steps based on a Markov process. The noise model is a Markov process over a discrete state-space. In a discrete state-space, each node and edge transitions to a different state with the addition of noise at every time step. A state refers to a node type or an edge type. Following each transition, the node or edge changes to another type. The probabilities of transition from one state to another are represented as transition matrices for nodes and for edges. In DiGress, the next step \(G_{t}\) (Figure~\ref{fig:diffusion}) is obtained by the product of the transition matrix, \(Q_t\) at step \(t\) and the graph at the previous step, \(G_{t-1}\). The transition probability, \(q\) at \(G_{t}\) is given as:
\begin{equation}
    q(G_t|G_{t-1}) = G_{t-1}Q_t
\end{equation}
at time step \(t-1\). The noise at step \(t\) is calculated as the product of transition matrices for \(t\) steps. DiGress adds noise using a cosine noise scheduler for \(t\) steps. At the end of \(t\) steps, transition matrices for node types and edge types are obtained. The transition probabilities at step \(t\) represent the complete noise distribution for each node and edge. Diffusion is applied to nodes and edges independently by adding marginal noise to each node and edge. The noise model adds noise to an input until the input changes to approximately pure noise. An input node feature vector, \(X\), follows the same noise model to transition to the previous step, \(X_{t-1}\). The pure noise is represented as a uniform distribution, which is the initial prior distribution for the denoising network, \(p_\theta\), during the reverse process. The denoising network in DiGress is made of GraphTransformer layers. Each layer has a multi-headed attention module. The denoising network uses self-attention to recover the input distribution by predicting the noise added at each previous step. The training loss, \(L_{Graph}\) for graph is given as 
\[L_{Graph} = L_{Nodes} + \lambda L_{Edges}\]
where node loss, \( L_{Nodes}\) and edge loss, \(L_{Edges}\) represent the cross-entropy loss between the true node and edge, and predicted probabilities of each node and edge, and \(\lambda \in \mathbb{R}^+\) assigns relative importance to nodes and edges. The graph loss is computed as a combined loss of individual node loss and edge loss. The loss of each node and edge is measured using cross-entropy. DiGress models node features along with the graph structure. The node features are one-hot encoding of node types. Extra features are added to nodes, including structural information that encodes the presence of cycles, and spectral features for positional encodings needed for a transformer model. These extra features are added at each diffusion step, as the message passing networks are unable to capture the presence of cycles. DiGress performs well at generating small heterogeneous graphs with the same node size as the input graph. Discrete node types and edge types are sampled from the complete noise distributions. The samples are input to a denoising network, which uses self-attention to predict the noise and outputs the distributions for node types and edge types for each previous step.

\subsection{TabDDPM}
TabDDPM~\cite{Kotelnikov2023TabDDPM} extends the denoising diffusion probabilistic (DDPM) approach used for images to synthesize tabular data. In TabDDPM, an input feature vector, \(x = [x_{num}, x_{cat_1}\dots x_{cat_C}]\) consists of both numeric features, \(x_{num}\) and categorical features, \(x_{cat_i}\) with \(i \in C\) categories. The categorical features are represented as one-hot encoding of categories. TabDDPM models categorical and binary features using multinomial diffusion and numerical features using continuous noise. The forward process adds Gaussian noise using a Markov process, while the reverse process uses a denoising neural network consisting of MLP layers with SiLU non-linear activation. The denoising network is an MLP architecture of the form~\cite{Gorishniy2021MLP_architecture}:
\begin{equation}
    \mathit{MLP}(x) = Linear(\mathit{MLPBlock}(\dots(\mathit{MLPBlock}(x))))
\end{equation}
where \(\mathit{MLPBlock}(x) = Dropout(ReLU(Linear(x)))\).
The TabDDPM method outperforms GAN and VAE in synthetic tabular data generation. TabDDPM captures training loss using mean-squared error and KL-Divergence. Our modification of this approach for generating node features uses discrete noise to model discrete features and cross-entropy to measure node feature loss during training.

\section{Additional Details for DiGNNExplainer}

\subsection{DiTabDDPM - Denoising network}

Our denoising network
\begin{align*}
\texttt{MLPOut}(\mathbf{x}) &:= \texttt{Lin}(\texttt{SiLU}\dots \\
&\quad\quad \texttt{Lin}(\texttt{SiLU}(\texttt{Lin}(\texttt{Layers}))))
\end{align*}
consists of \texttt{MLPBlock} layers
\begin{align*}
\texttt{Layers}&:=\texttt{MLPBlock}(\dots(\texttt{MLPBlock}(\texttt{MLPIn}))) \\
\texttt{MLPIn}&:=\texttt{Lin}(\texttt{ReLU}(\texttt{SiLU}(\dots\\
              &\quad\quad \texttt{Lin}(\texttt{ReLU}(\texttt{SiLU}(\mathbf{x})))))) \\
\texttt{MLPBlock}&:=\texttt{Lin}(\texttt{Drop}(\texttt{SiLU}(\texttt{Lin}\\
&\quad\quad (\texttt{Drop}(\texttt{MLPIn})))))
\end{align*}
where $\mathbf{x}$ denotes the discrete feature vector for a single node, \texttt{Lin} a linear layer, \texttt{SiLU} a sigmoid linear unit, \texttt{ReLU} a rectified linear unit, and \texttt{Drop} a dropout layer. An \texttt{MLPBlock} consists of normalization and dropout layers with a combination of \texttt{ReLU} and \texttt{SiLU} activations. Normalization is added to the input and output of each \texttt{MLPBlock} layer. The output of an \texttt{MLPBlock} layer is the input for the next \texttt{MLPBlock} layer. In a pilot study, we experimented with different architectures, including different normalization layers and activation functions. We found the given architecture to yield the smallest loss on the validation set.

\subsection{Node Feature Reconstruction Loss}

The node feature reconstruction loss, $L_{recon}$, at validation time, is computed using the sampled features, $\mathbf{x}_0$, drawn from the probability distributions at step $0$, and $\mathbf{x}_t$, drawn from the predicted probability distribution of the denoising network $\theta$, at step $t$. It is given as the expected value of the logarithm of the predicted probability of the input features given the noisy features at step $t$. 
\[
L_{recon} = \mathbb{E}\left[\log {p_\theta}(\mathbf{x}_0|\mathbf{x}_t)\right]
\]

\subsection{Postprocessing}

Two nodes $\tau_1, \tau_2$ in the metagraph are connected if and only if there exist two nodes $v_1$ and $v_2$ in the graph such that $v_1$ has type $\tau_1$ and $v_2$ has type $\tau_2$. The generated graph is consistent if and only if the metagraph of the generated graph is a subgraph of the metagraph of the dataset.

\subsection{Explanation Graph Selection}

Let $\mathcal{G}$ be the set of valid graphs obtained in the above step (Postprocessing). Given a \emph{graph classification task}, for every graph $G \in \mathcal{G}$, we make a prediction $\hat{y}_G := f(G)$ where $\hat{y}_G \in [0,1]^C$ denotes the vector of predicted softmax probabilities and $f$ our trained GNN. Then, for every class~$c_i$ with $i \in \{1, \ldots, C\}$, we take the graph 
\[
G_i := \argmax_{G \in \mathcal{G}} \hat{y}_{G, i}
\]
with the highest predicted score as the explanation graph for class~$c_i$. Similarly, given a \emph{node classification task}, where we need to make a prediction for every node of type $\tau$, for every valid graph $G \in \mathcal{G}$ and every node $v\in G$ that has type $\tau$, we make a prediction $\hat{y}_{(G,v)}:=f(G, v)$ where $\hat{y}_{(G,v)} \in [0,1]^C$ denotes the vector of predicted softmax probabilities and $f$ our GNN. Then, for every class~$c_i$ with $i \in \{1, \ldots, C\}$, we take the graph
\[
G_i := \argmax_{G \in \mathcal{G}} \max_{v \in G, v \text{ has type }\tau} \hat{y}_{(G,v),i}
\]
with the highest predicted score as the explanation graph for class $c_i$. This way, we obtain one explanation graph per class.

\section{Datasets}

\begin{table}[tb]
  \centering
  \setlength{\tabcolsep}{3.3pt}
  \scriptsize
  \caption{Dataset statistics: number of graphs, average number of nodes per graph, number of classified nodes (e.g., author nodes for DBLP), number of classified graphs, number of features of the node to be classified, number of classes to be predicted, number of continuous features, and number of discrete features.  }
  \label{tab:table-datasets}
  \begin{tabular}{@{}lrrrrrrrr@{}}
      \toprule
      \textbf{Datasets} &  \textbf{Graphs} & \textbf{Nodes} & \textbf{Nodes} & \textbf{Graphs} & \textbf{Feat.} & \textbf{Classes} & \textbf{Cont.}  & \textbf{Discr.} \\
      & & \textbf{(avg.)} & \textbf{(classified)} & \textbf{(classified)} & & & \textbf{(feat.)} & \textbf{(feat.)} \\
      \midrule
      \textsc{DBLP}       &     1 & 26,128 & 4,057 &    -- &   334 & 4 &    50 & 4,565 \\
      \textsc{IMDB}       &     1 & 11,616 & 4,278 &    -- & 3,066 & 3 & 6,132 & 3,066 \\
      \textsc{MUTAG}      &   188 &     17 &    -- &   188 &    -- & 2 &    -- &    -- \\
      \textsc{BA-Shapes}  &     1 &    700 &   700 &    -- &    -- & 4 &    -- &    -- \\
      \textsc{Tree-Cycle} &     1 &    871 &   871 &    -- &    -- & 2 &    -- &    -- \\
      \textsc{Tree-Grids} &     1 &  1,231 & 1,231 &    -- &    -- & 2 &    -- &    -- \\
      \textsc{BA-3Motif}  & 3,000 &     21 &    -- & 3,000 &    -- & 3 &    -- &    -- \\
      \bottomrule
    \end{tabular}
\end{table}

We run experiments for DiGNNExplainer on the following datasets in Table~\ref{tab:table-datasets}: \textsc{DBLP}~\cite{Fu2020MAGNN} is a heterogeneous dataset of a citation network with four node types---author, paper, term, and conference. Author node features are bag-of-words representations of their paper keywords. Paper nodes have bag-of-words features, term nodes have continuous features, and conference nodes have no features. 
\textsc{IMDB}~\cite{Fu2020MAGNN} is a heterogeneous dataset for movies with three node types---movie, director, and actor. The movie node features include bag-of-words representations of their plot keywords. Director and actor nodes have continuous features.
\textsc{MUTAG}~\cite{Morris2020TUDataset} is a molecule graph classification dataset where each graph can be classified into two classes---mutagenic and non-mutagenic, and each node has a single node type, atom.
The synthetic \textsc{BA-Shapes}~\cite{Ying2019GNNExplainer} dataset is constructed by randomly attaching house motifs to the nodes of a Barabási–Albert (BA) base graph. All nodes are of a single type.
\textsc{Tree-Cycle}~\cite{Ying2019GNNExplainer} and \textsc{Tree-Grids}~\cite{Ying2019GNNExplainer} are constructed by randomly attaching cycle motifs and n-by-n grid motifs, respectively, to the nodes of a balanced binary tree base graph. 
\textsc{BA-3Motif}~\cite{Wang2021ReFine} is a motif graph classification dataset where a house/cycle/grid motif is attached randomly to the nodes of Barabási–Albert (BA) base graphs. The graphs can be classified into any of three types, as per the type of motif attached.

\section{Validity Analysis}

We observe that a higher number of graphs generated using DiGNNExplainer are valid (i.e., consistent with the metagraph of the dataset) for \textsc{DBLP} compared to \textsc{IMDB} (Table~\ref{tab:table-validity-analysis}). For homogeneous datasets, the number of connected graphs is the same as valid graphs, as homogeneous graphs do not require validation using a metagraph. 

\begin{table}[tb]
  \centering
  \scriptsize
  \setlength{\tabcolsep}{6pt}
  \caption{Analysis of the validity of generated graphs according to the metagraph.}
  \label{tab:table-validity-analysis}
  \begin{tabular}{@{}lrrrr@{}}
    \toprule
    \textbf{Approach} & \textbf{\#Graphs} & \textbf{\#Graphs} & \textbf{\#Graphs}  & \textbf{Avg. nodes} \\ 
    & \textbf{(Generated)} & \textbf{(Connected)} & \textbf{(After validity check)} & \textbf{per graph} \\
    \midrule
    \multicolumn{5}{c}{\textsc{DBLP}} \\
    \midrule
    DiGNNExplainer & 1536 & 1416 & 231 & 12.5\\
    VAE(baseline)  & 1536 & 1283 & 210 & 12.5\\
    \midrule
    \multicolumn{5}{c}{\textsc{IMDB}} \\
    \midrule
    DiGNNExplainer & 1536 & 1367 & 188 & 7.92 \\
    VAE(baseline)  & 1536 &  948 & 513 & 7.92 \\
    \midrule
    \multicolumn{5}{c}{\textsc{MUTAG}} \\
    \midrule
    DiGNNExplainer & 1536 &  731 &  731 & 13 \\
    VAE(baseline)  & 1536 & 1242 & 1242 & 13 \\
    \midrule
    \multicolumn{5}{c}{\textsc{BA-Shapes}} \\
    \midrule
    DiGNNExplainer & 1536 & 1058 & 1058 & 12.5 \\
    VAE(baseline)  & 1536 & 1323 & 1323 & 12.5 \\
    \midrule
    \multicolumn{5}{c}{\textsc{Tree-Cycle}} \\
    \midrule
    DiGNNExplainer & 1536 & 1398 & 1398 & 12.5 \\
    VAE(baseline)  & 1536 & 1237 & 1237 & 12.5 \\
    \midrule
    \multicolumn{5}{c}{\textsc{Tree-Grids}} \\
    \midrule
    DiGNNExplainer & 1536 & 1196 & 1196 & 12.5 \\
    VAE(baseline)  & 1536 & 1312 & 1312 & 12.5 \\
    \midrule
    \multicolumn{5}{c}{\textsc{BA-3Motif}} \\
    \midrule
    DiGNNExplainer & 1536 & 1525 & 1525 & 15 \\
    VAE(baseline)  & 1536 & 1369 & 1369 & 15 \\
    \bottomrule
  \end{tabular}
\end{table}

\newpage

\section{Community Detection}

We use communities or motifs to evaluate the ground-truth faithfulness of the explanation graphs (see Section~\ref{subsec:evaluation-setup}), following previous work~\cite{Ying2019GNNExplainer,Yuan2020XGNN}. Table~\ref{tab:table-qualitative} (right) shows example motifs.

For \emph{real-world} graphs (\textsc{DBLP}, \textsc{IMDB}, \textsc{MUTAG}), we employ Louvain community detection%
\footnote{\url{https://python-louvain.readthedocs.io/en/latest/api.html}}%
~\cite{Blondel2008Fast}
which identifies the largest substructures in input graphs. Unlike existing techniques~\cite{Subramonian2021MOTIF-Driven,Yu2022MotifExplainer} that often rely on blackbox models for motif detection, our method is more similar to GraphXAI~\cite{agarwal2023evaluating}, which detects ground-truth substructures in the real-world dataset MUTAG
via substructure matching. For \emph{synthetic} graphs (\textsc{BA-3Motif}, \textsc{BA-SHAPES}, \textsc{Tree-Cycle}, and \textsc{Tree-Grids}), we generate motifs using ShapeGGen~\cite{agarwal2023evaluating}.

We use \emph{class-specific} motifs when motifs can be associated with a prediction class, and \emph{common} motifs otherwise. For the \emph{graph classification datasets} (\textsc{MUTAG}, \textsc{BA-3MOTIF}), each graph belongs to a single class, so we construct class-specific motif sets. Similarly, for the \emph{real-world node classification datasets} \textsc{DBLP} and \textsc{IMDB}, we obtain class-specific motifs. In \textsc{DBLP}, most (104 out of 120) of the sampled graphs using ForestFireSampler contain author nodes from a single class, in \textsc{IMDB}, 40 out of 143 graphs contain movie nodes from a single class. If a sampled graph has nodes of different classes (e.g., containing author nodes from classes 1 and 3), we do not use it for motif detection, because the obtained motifs might be common across classes. In this way, we detect motifs that help to distinguish different classes.

In contrast, for the \emph{synthetic node classification datasets} \textsc{BA-Shapes}, \textsc{Tree-Cycle}, and \textsc{Tree-Grids}, the task is to predict node labels and every graph contains nodes of all labels. For example, in \textsc{BA-Shapes}, a house motif has all four node labels. As a result, motifs are naturally shared across labels/classes, and we evaluate faithfulness using a single set of \emph{common} motifs. Table~\ref{tab:table-motif-statistics} shows the statistics for class-specific and common motifs for our datasets.

\begin{table}[tb]
  \centering
  \setlength{\tabcolsep}{0.5pt}
  \caption{Explanation graphs for Class 1}
  \label{tab:table-qualitative}
  \resizebox{\columnwidth}{!}{
    \begin{tabular}{ >{\flushleft\arraybackslash}m{0.5in}  >{\centering\arraybackslash}m{0.75in} >{\centering\arraybackslash}m{0.95in} >{\centering\arraybackslash}m{0.85in}>{\centering\arraybackslash}m{1.05in} |>{\centering\arraybackslash}m{0.85in} >{\centering\arraybackslash}m{0.6in}>{\centering\arraybackslash}m{0.6in}}
      \toprule
      & \textbf{XGNN} & \textbf{GNNInterpreter} & \textbf{D4Explainer} & \textbf{DiGNNExplainer} & & \textbf{Motifs} & \\
      \midrule
      \textsc{DBLP} 
        & \includegraphics[width=1\linewidth]{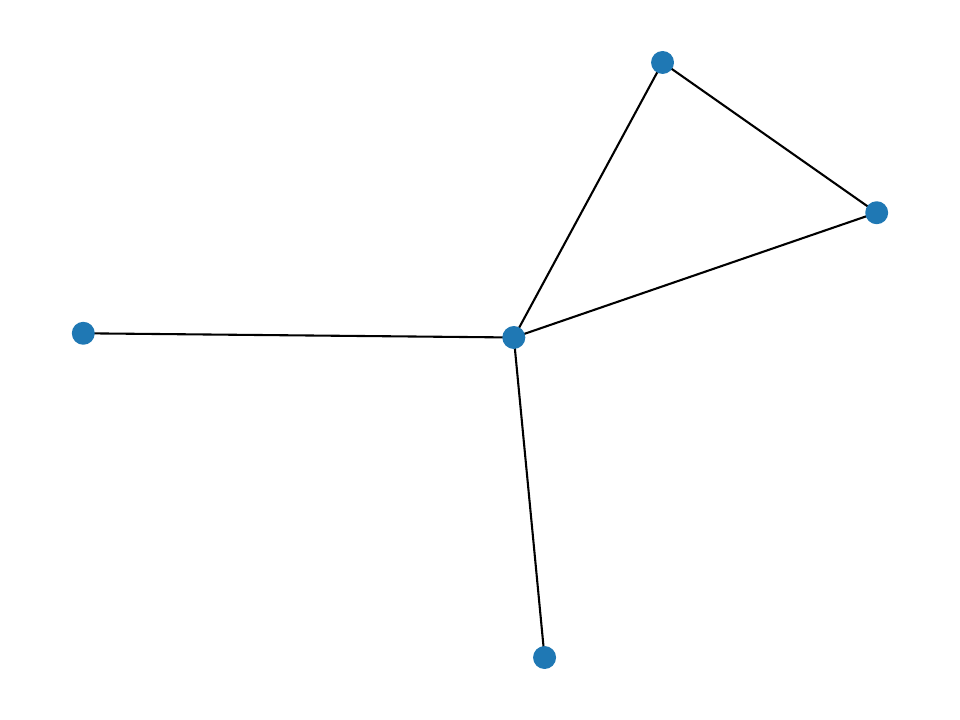}
        & \includegraphics[width=1\linewidth]{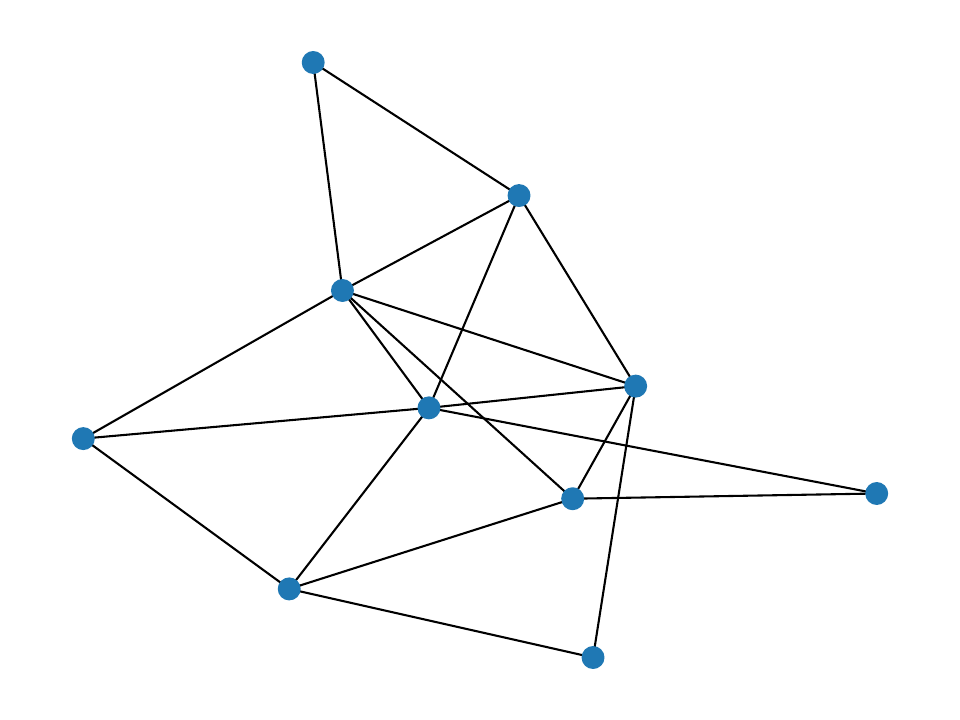}
        & \includegraphics[width=1\linewidth]{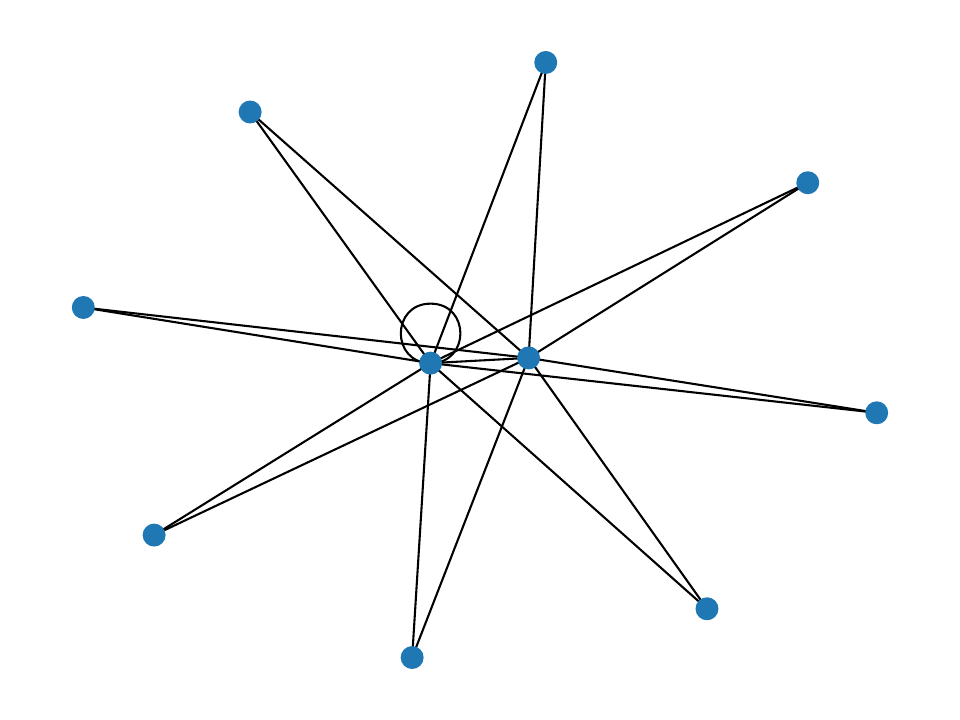}
        & \includegraphics[width=1\linewidth]{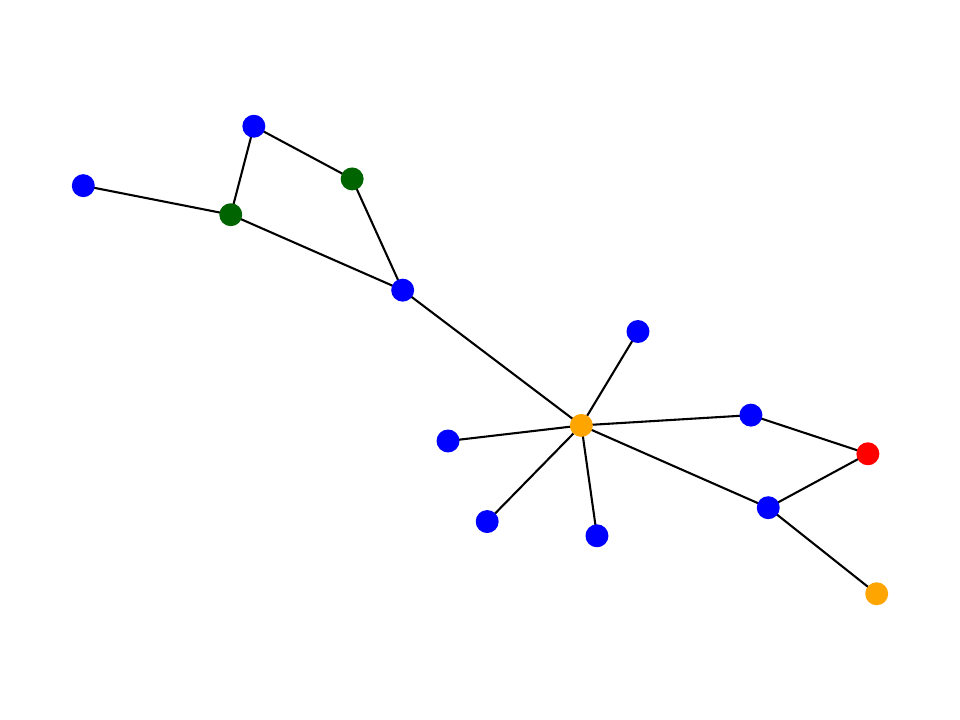}
        & \includegraphics[width=1.5cm]{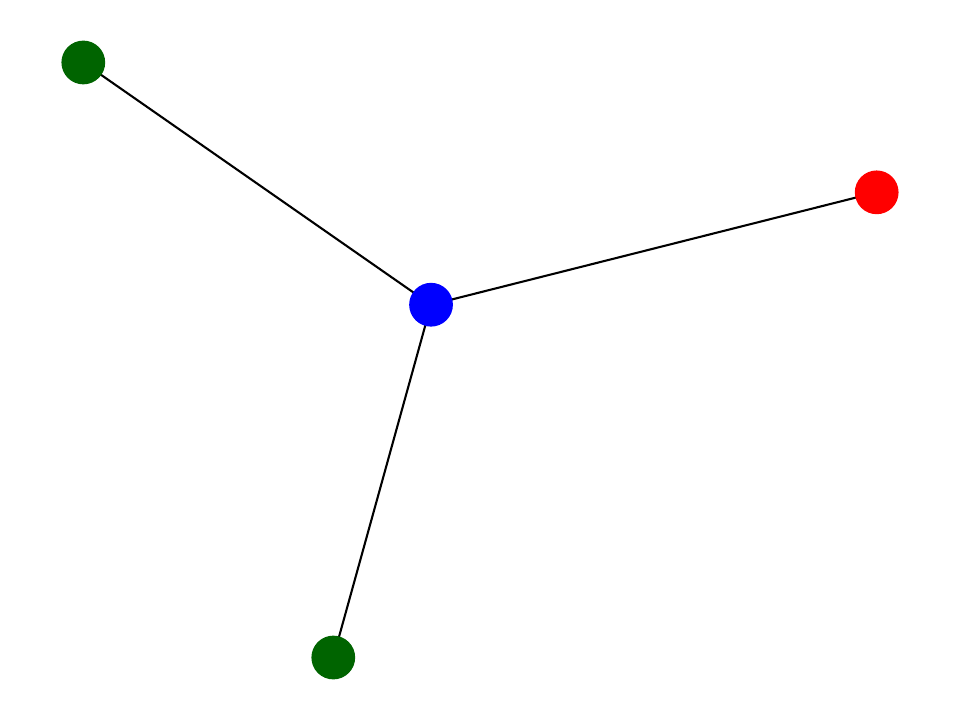}
        & \includegraphics[width=1.5cm]{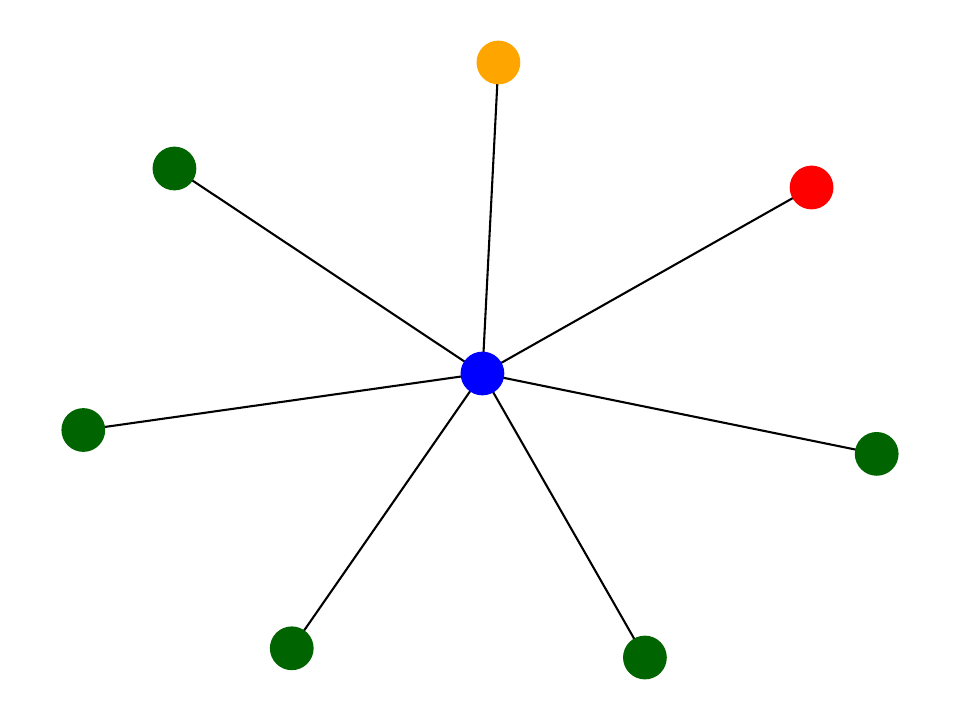}
        & \includegraphics[width=1.5cm]{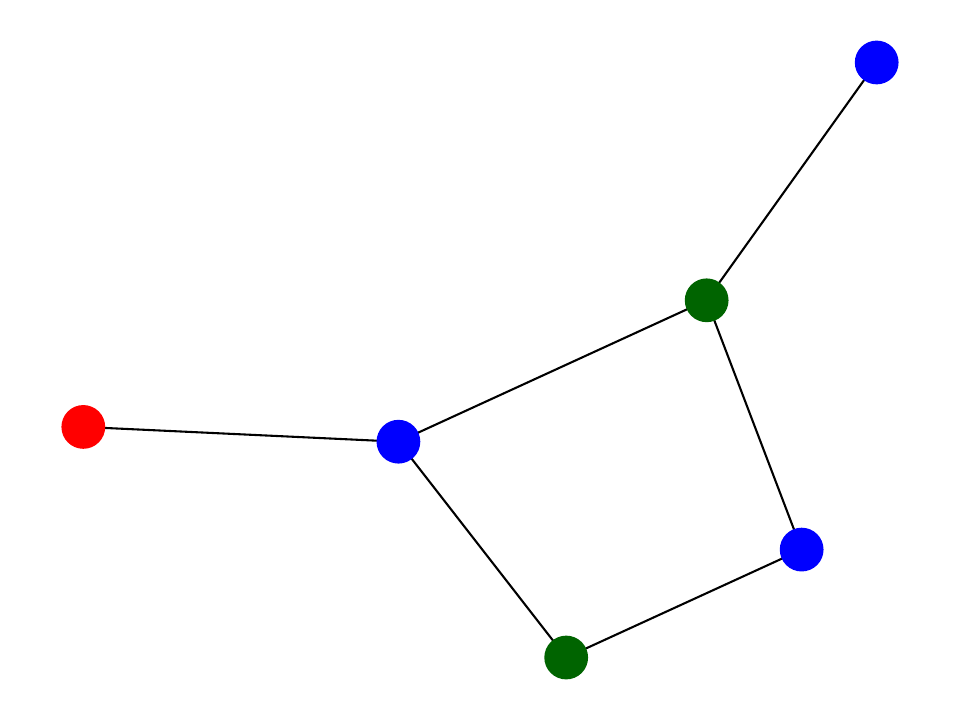} \\ 
 
      \textsc{IMDB}
        & \includegraphics[width=1\linewidth]{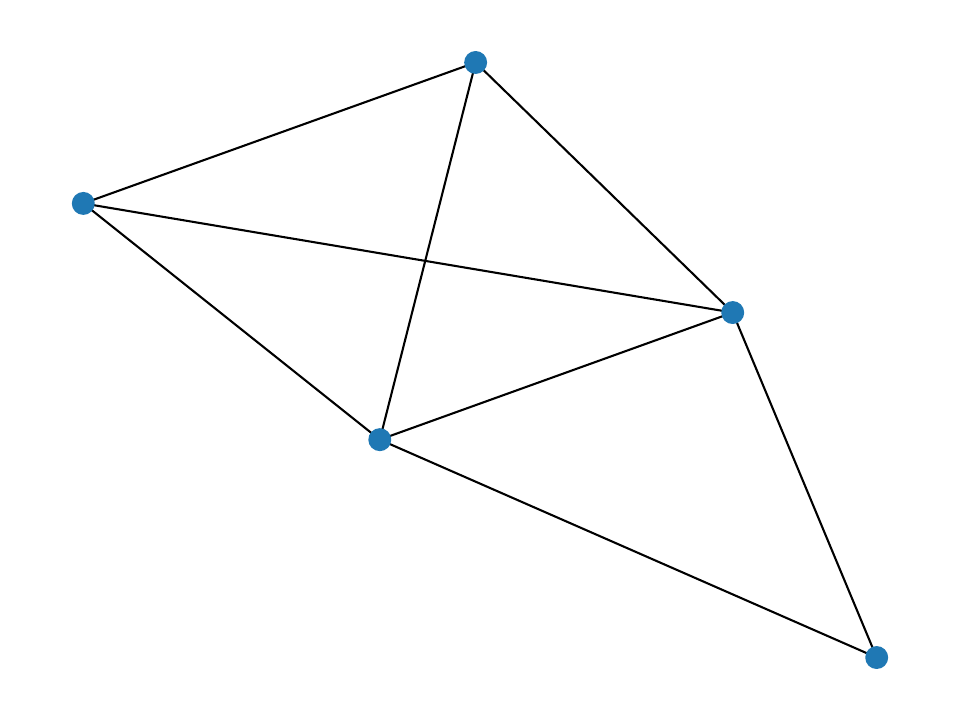}
        & \includegraphics[width=1\linewidth]{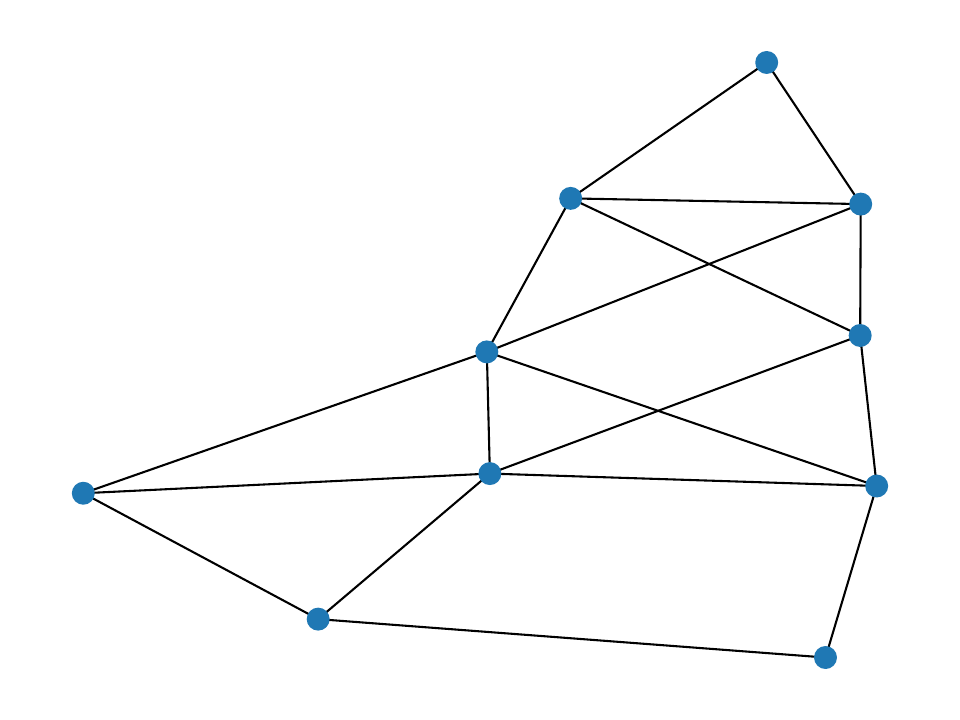}
        & \includegraphics[width=1\linewidth]{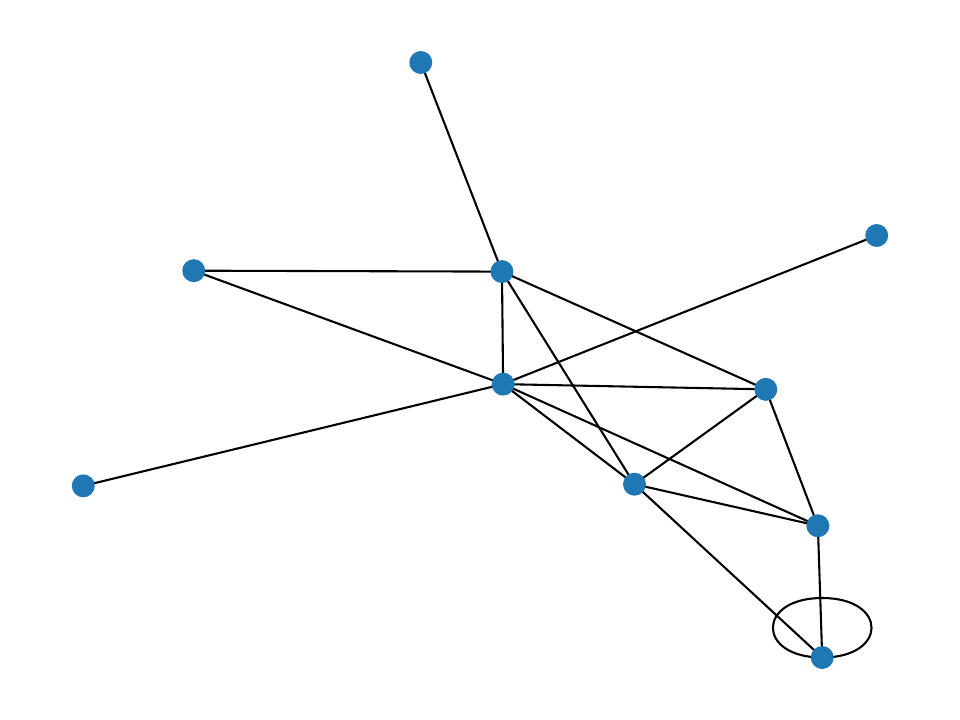}
        & \includegraphics[width=1\linewidth]{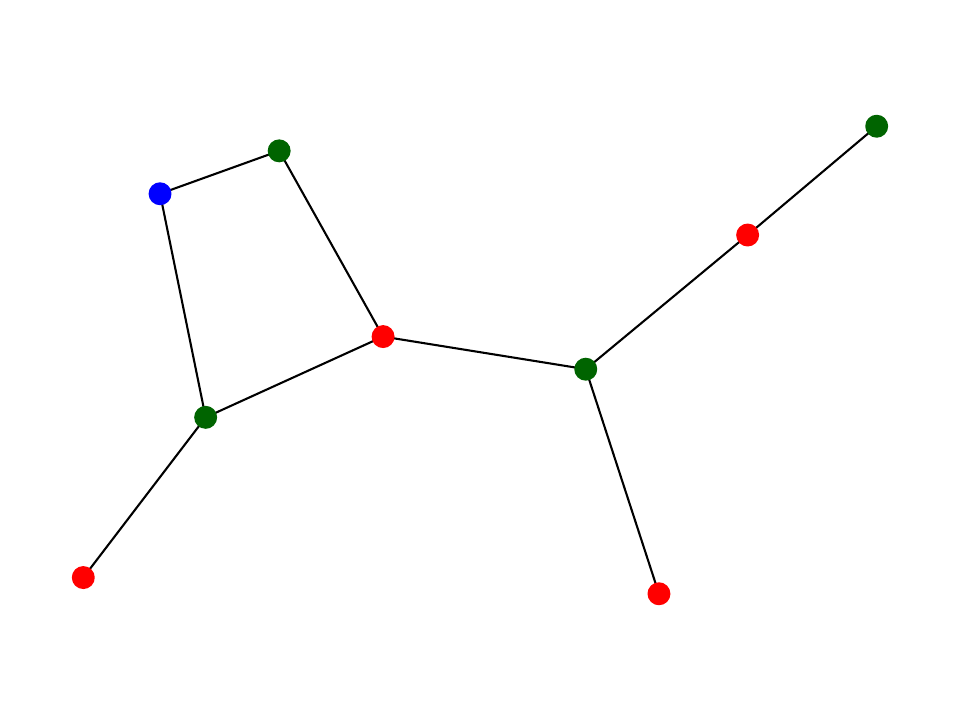}
        & \includegraphics[width=1.5cm]{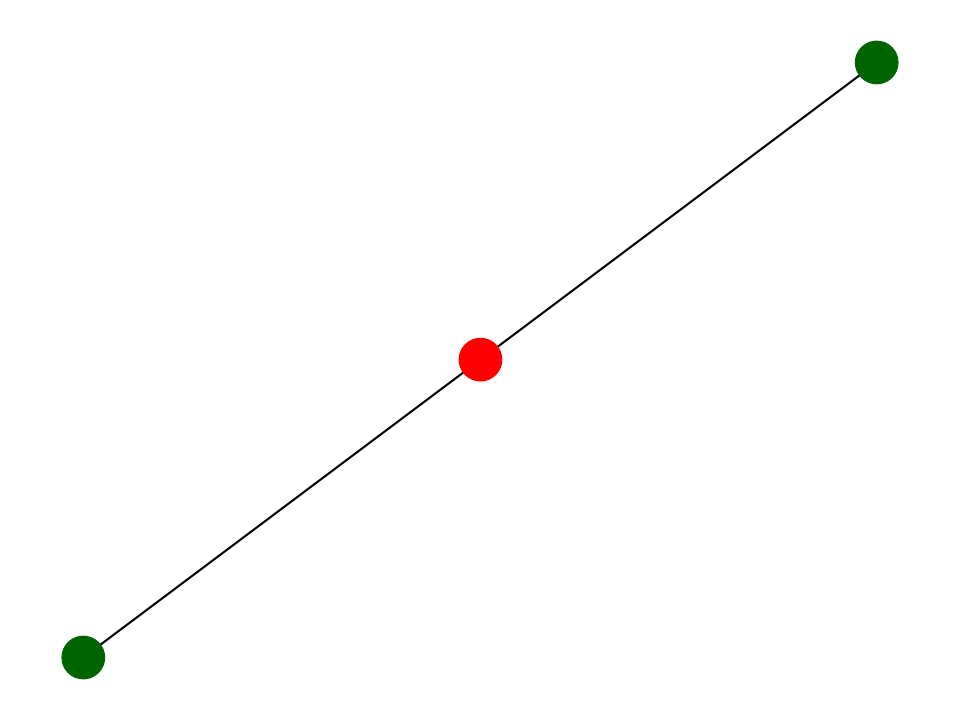}
        & \includegraphics[width=1.5cm]{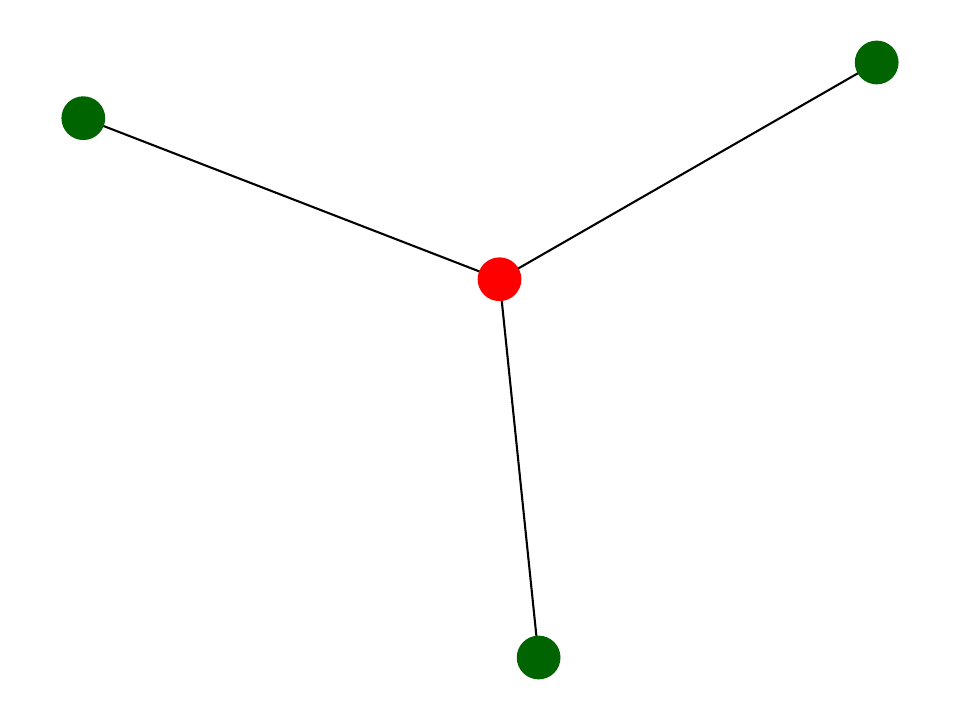}
        & \includegraphics[width=1.5cm]{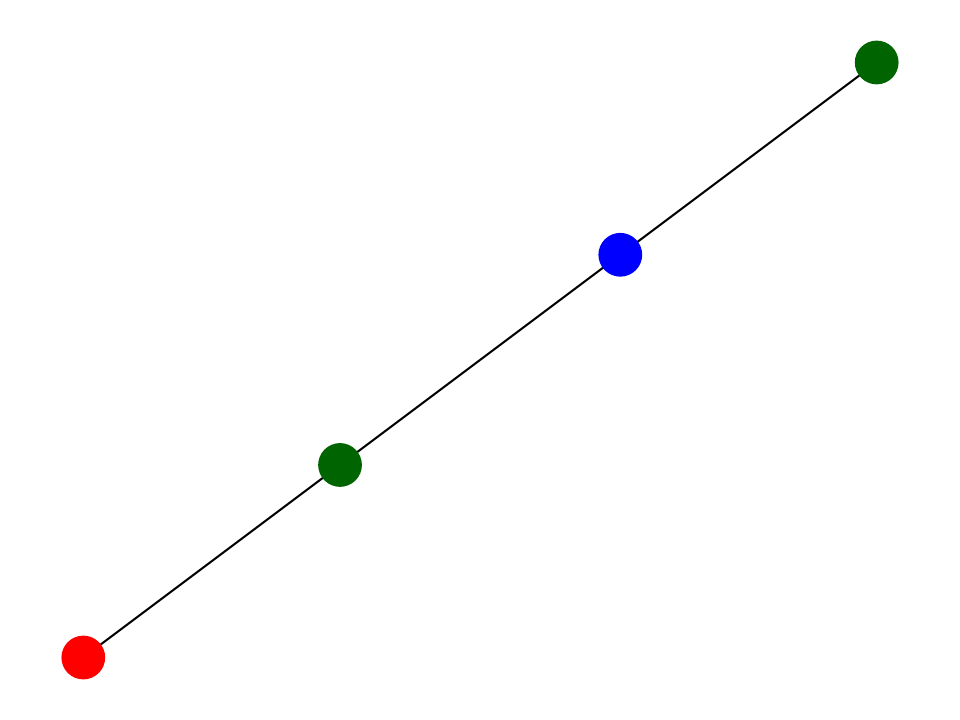}  \\ 

      \textsc{MUTAG}
        & \includegraphics[width=1\linewidth]{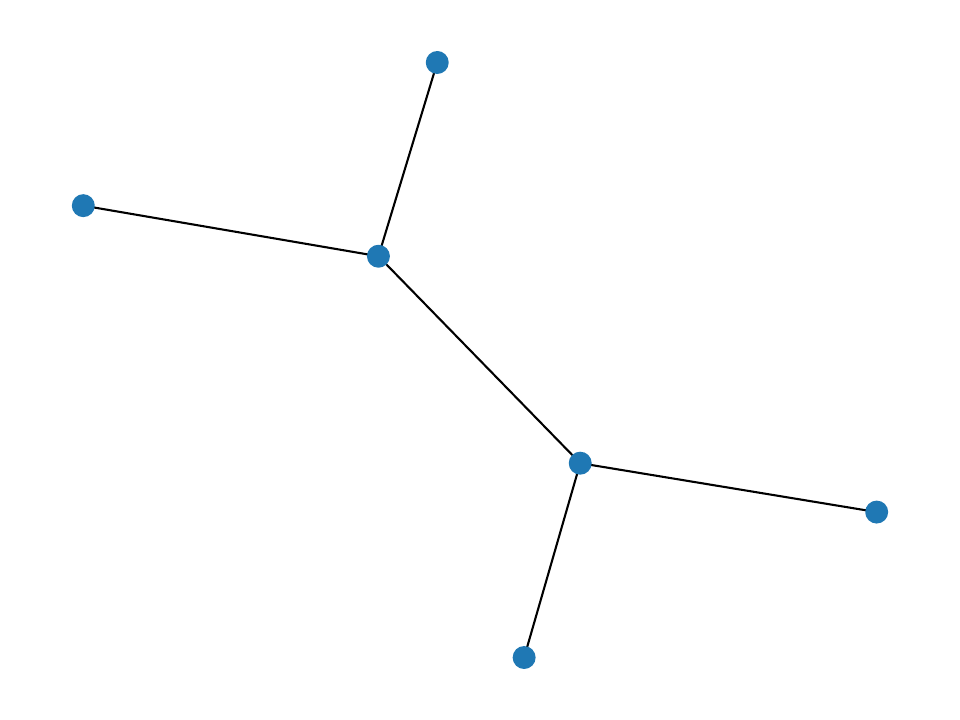}
        & \includegraphics[width=1\linewidth]{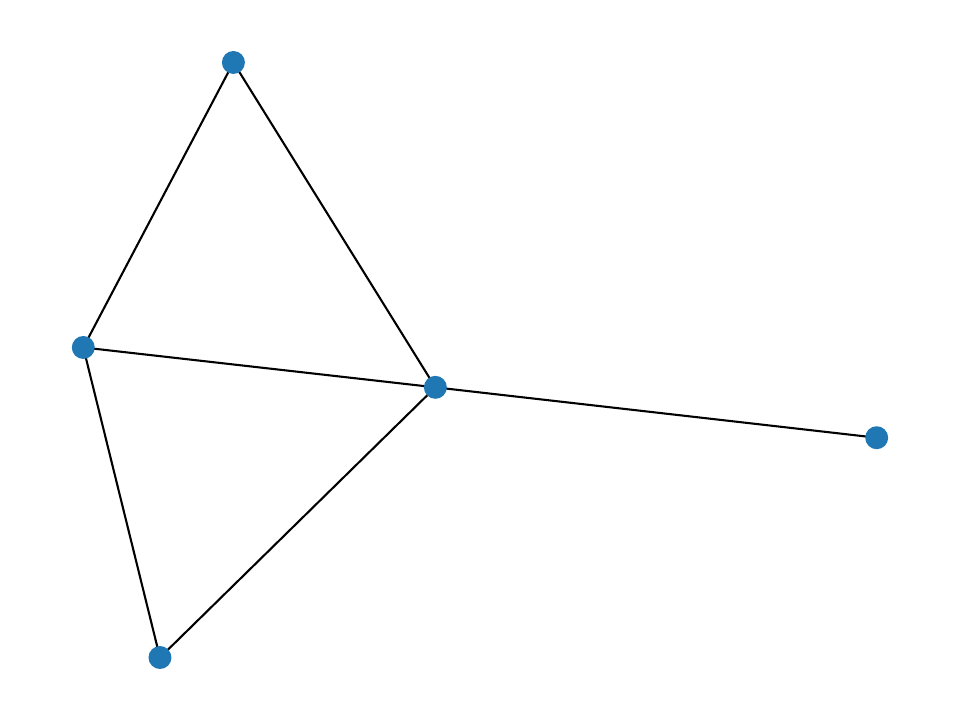}
        & \includegraphics[width=1\linewidth]{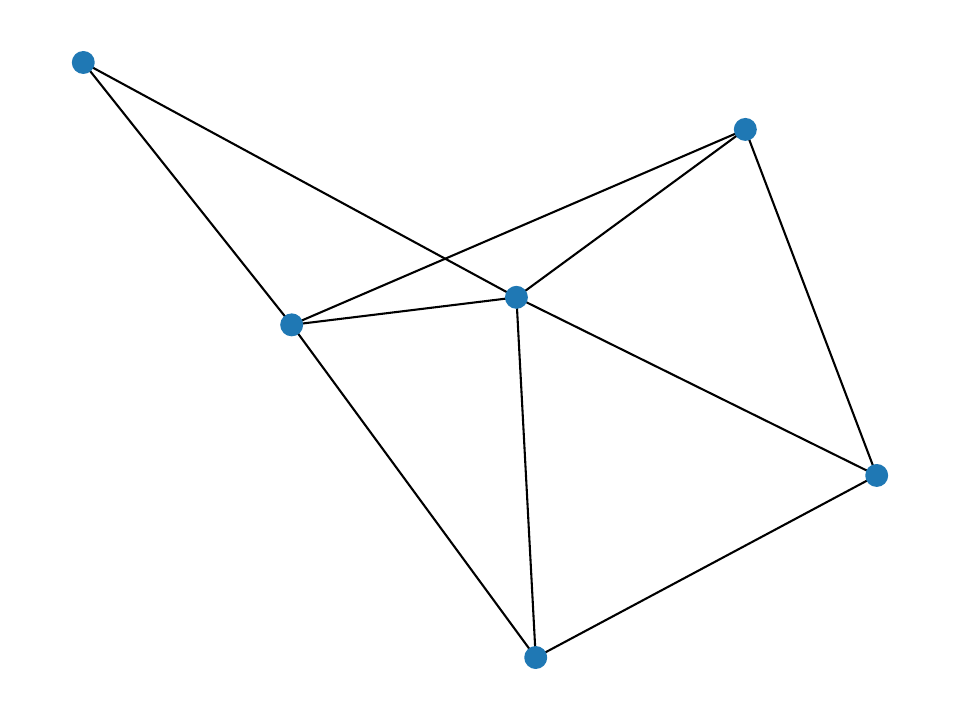}
        & \includegraphics[width=1\linewidth]{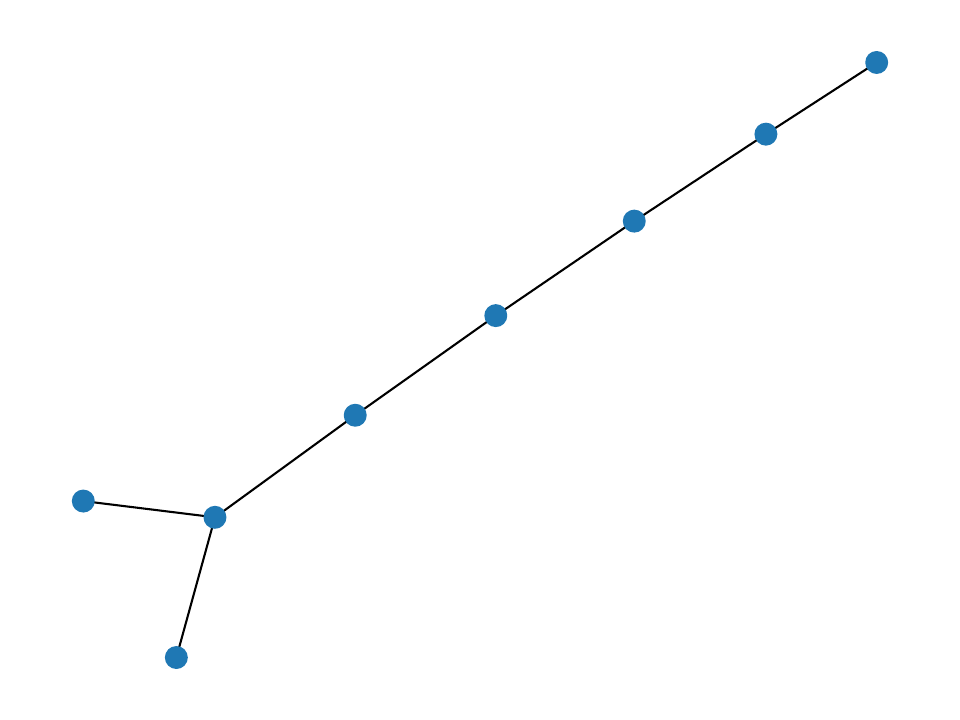}
        & \includegraphics[width=1.5cm]{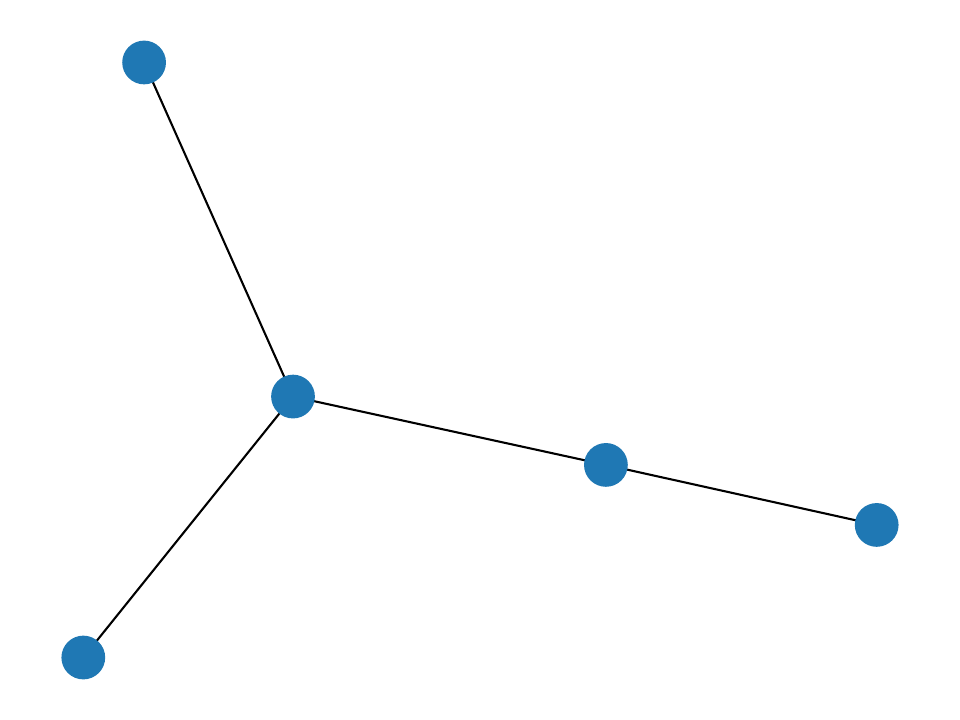}
        & \includegraphics[width=1.5cm]{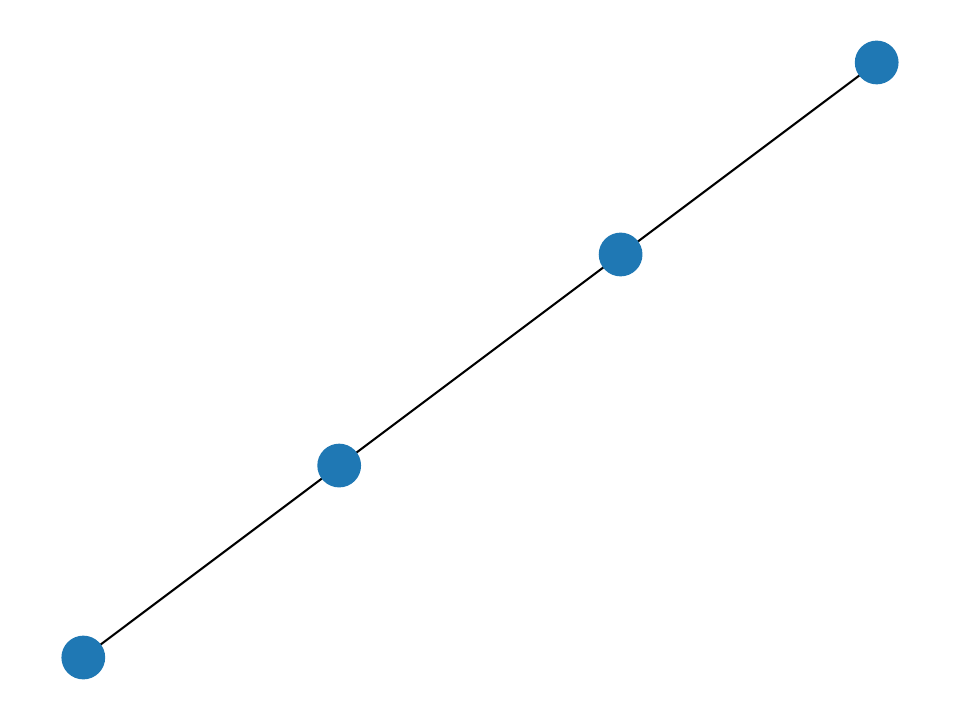}
        & \includegraphics[width=1.5cm]{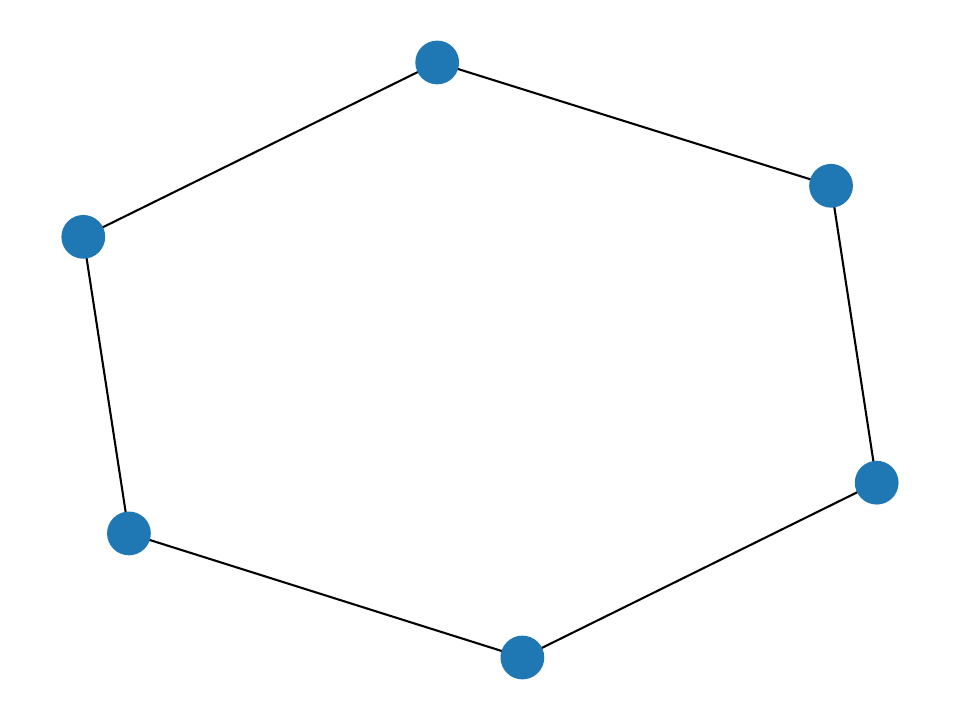} \\
    
      \textsc{BA-Shapes}
        & \includegraphics[width=1\linewidth]{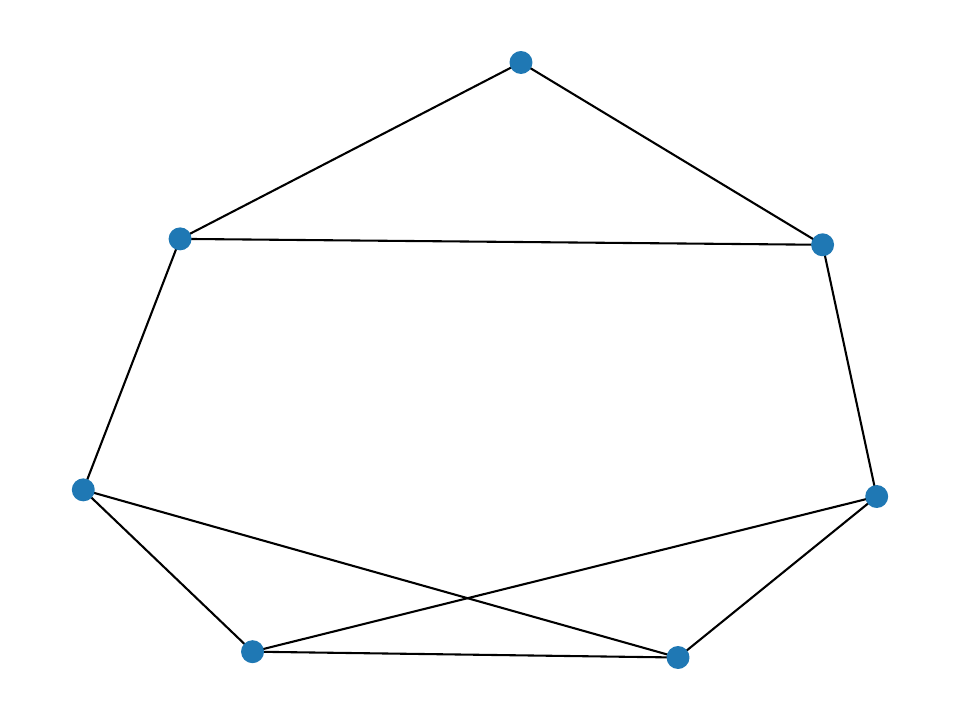}
        & \includegraphics[width=1\linewidth]{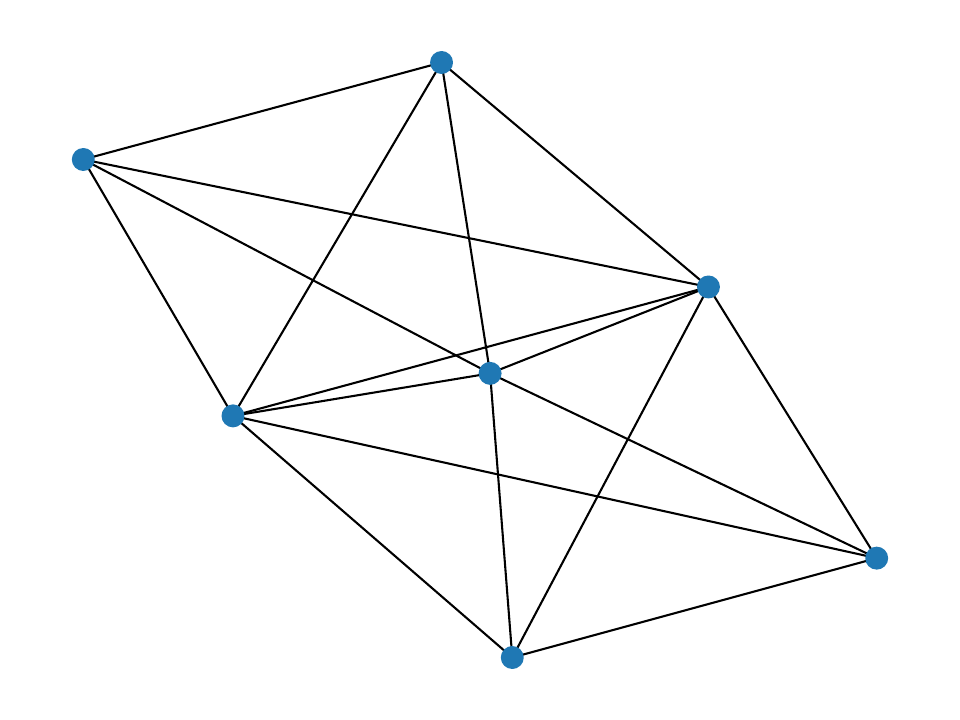}
        & \includegraphics[width=1\linewidth]{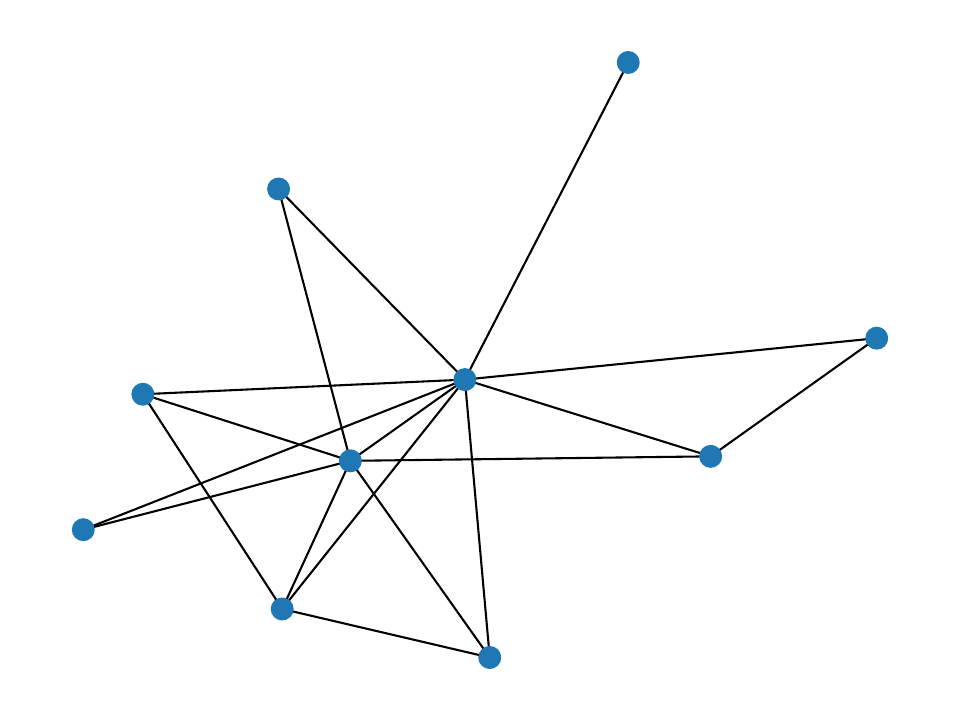}
        & \includegraphics[width=1\linewidth]{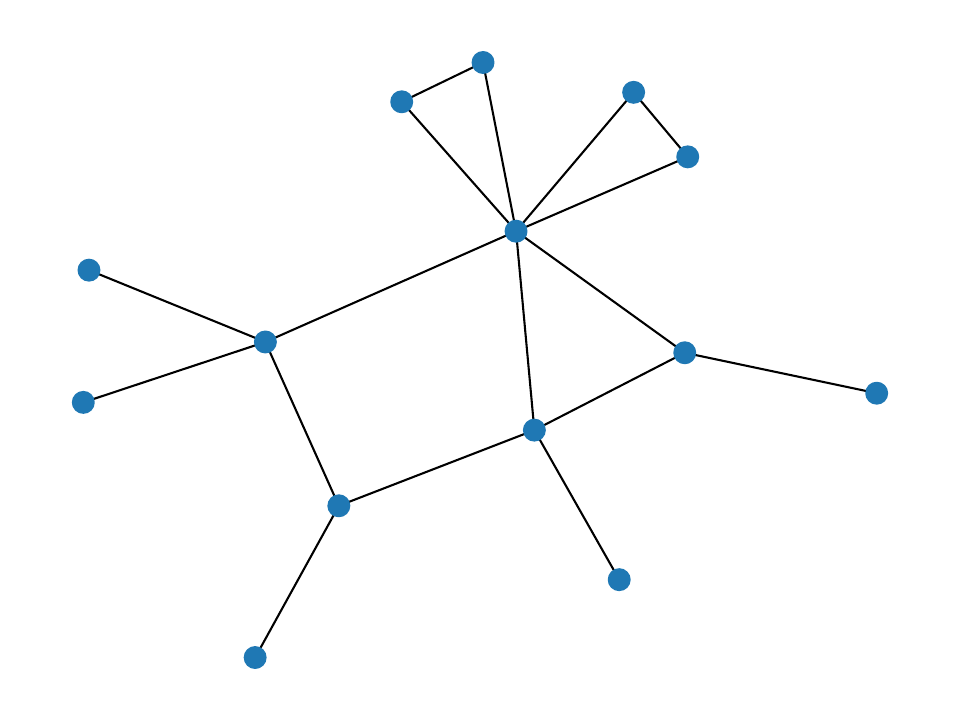}
        & \includegraphics[width=1.5cm]{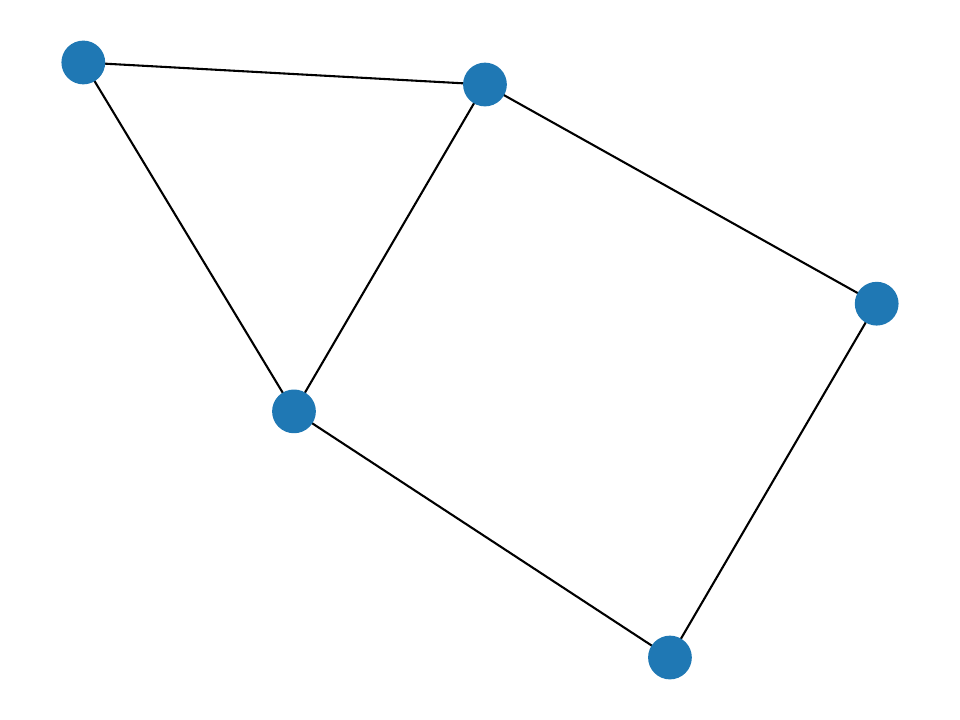}
        & \includegraphics[width=1.5cm]{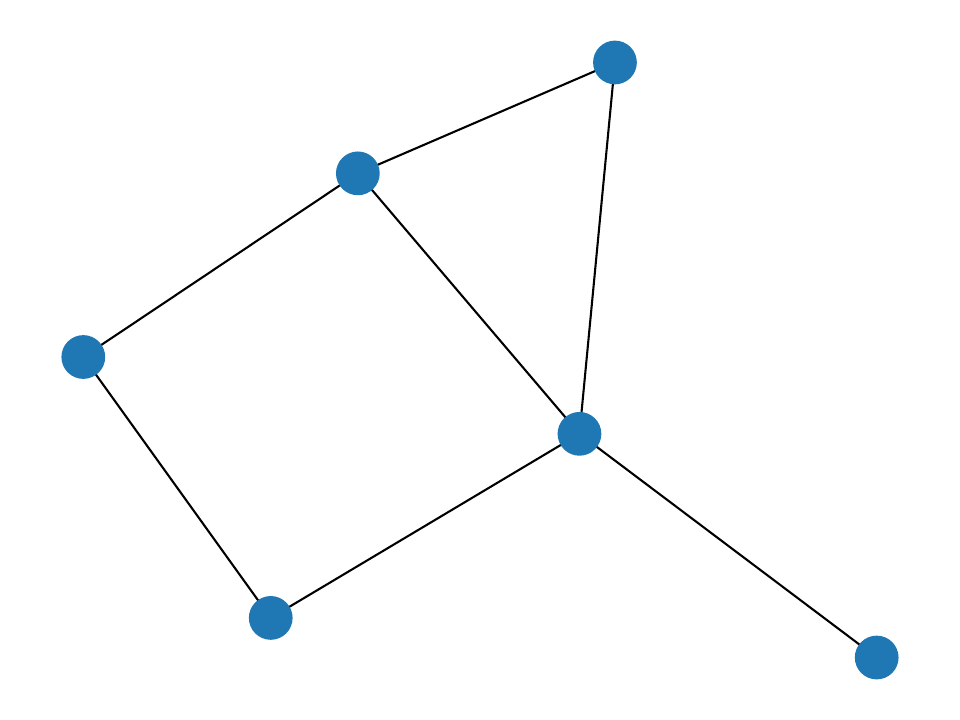}  \\ 
    
      \textsc{Tree-Cycle}
        & \includegraphics[width=1\linewidth]{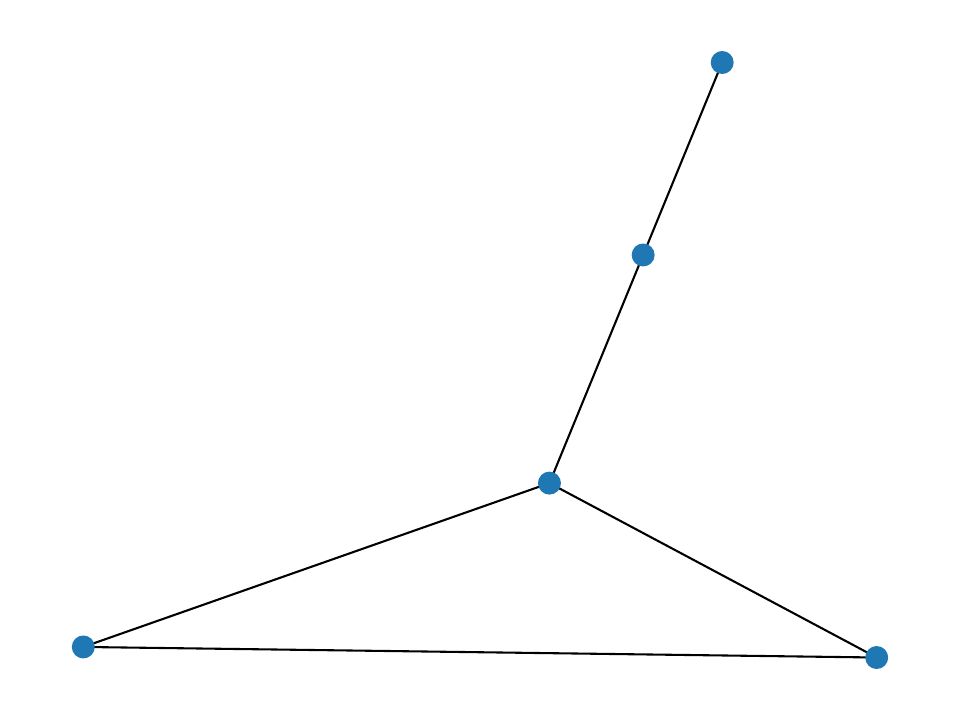}
        & \includegraphics[width=1\linewidth]{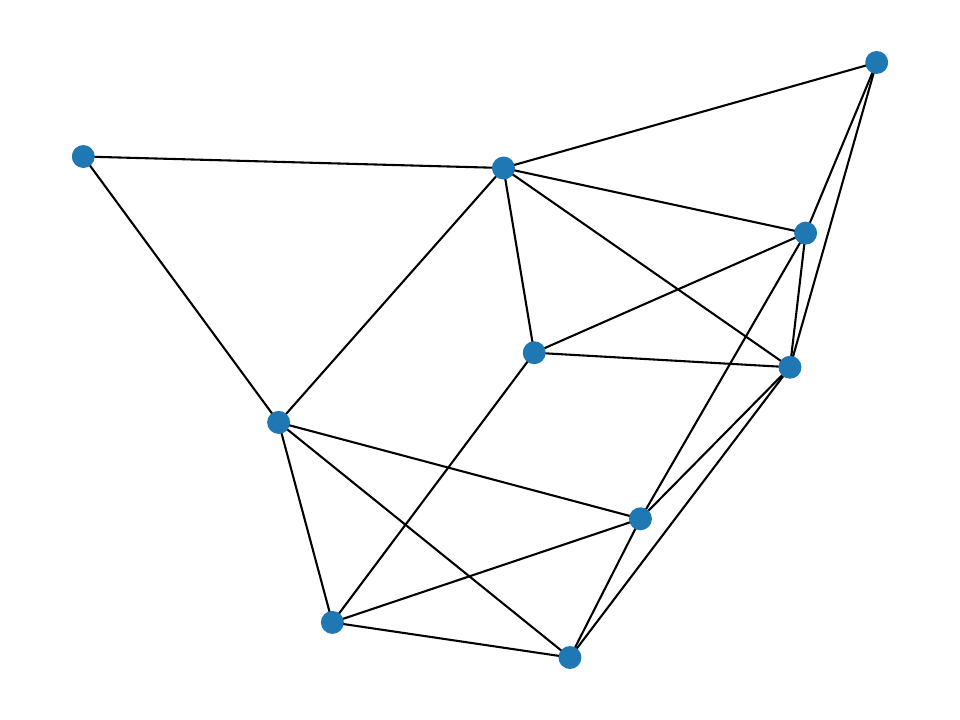}
        & \includegraphics[width=1\linewidth]{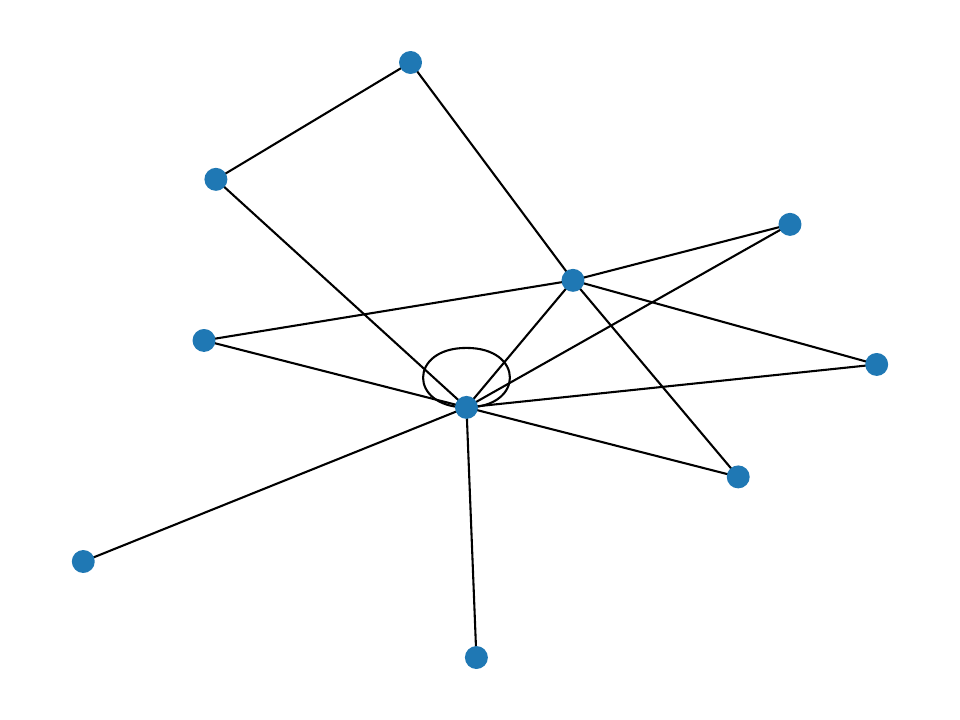}
        & \includegraphics[width=1\linewidth]{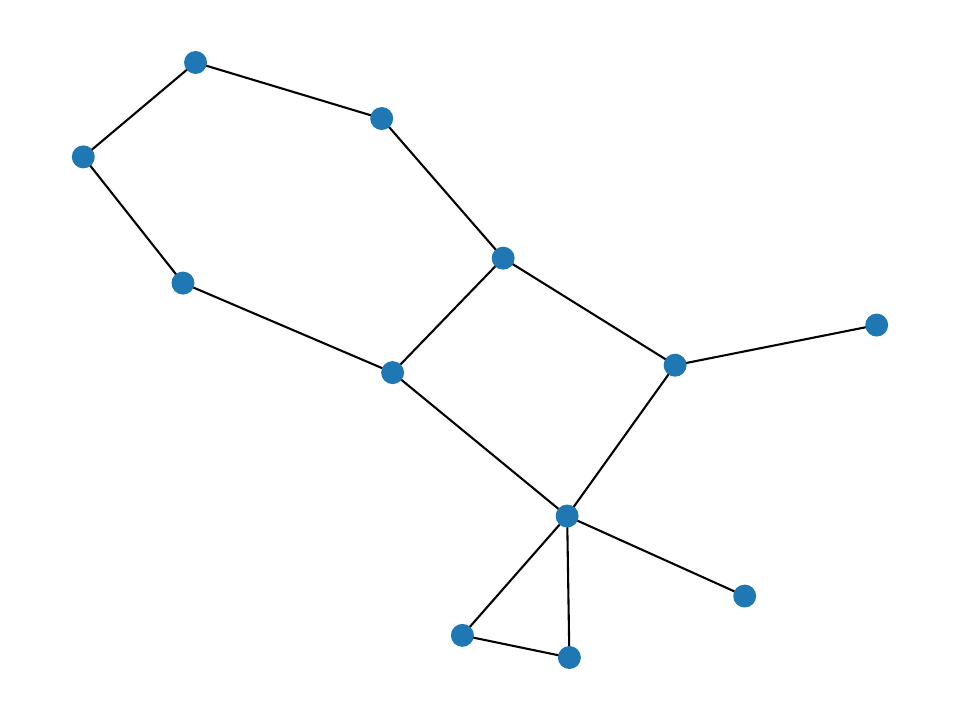}
        & \includegraphics[width=1.5cm]{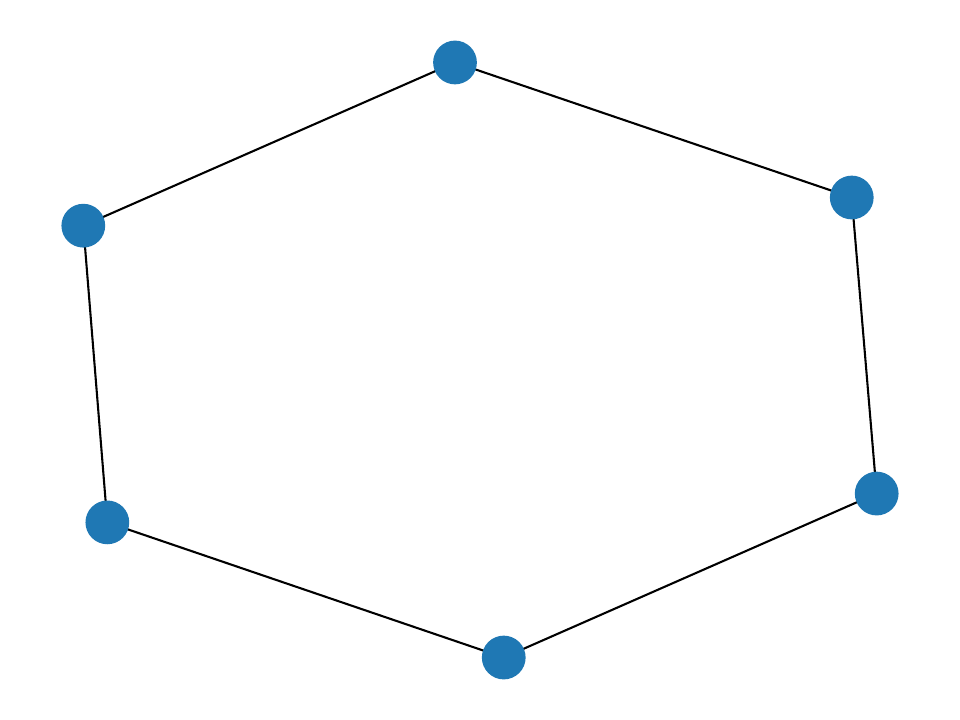}
        & \includegraphics[width=1.5cm]{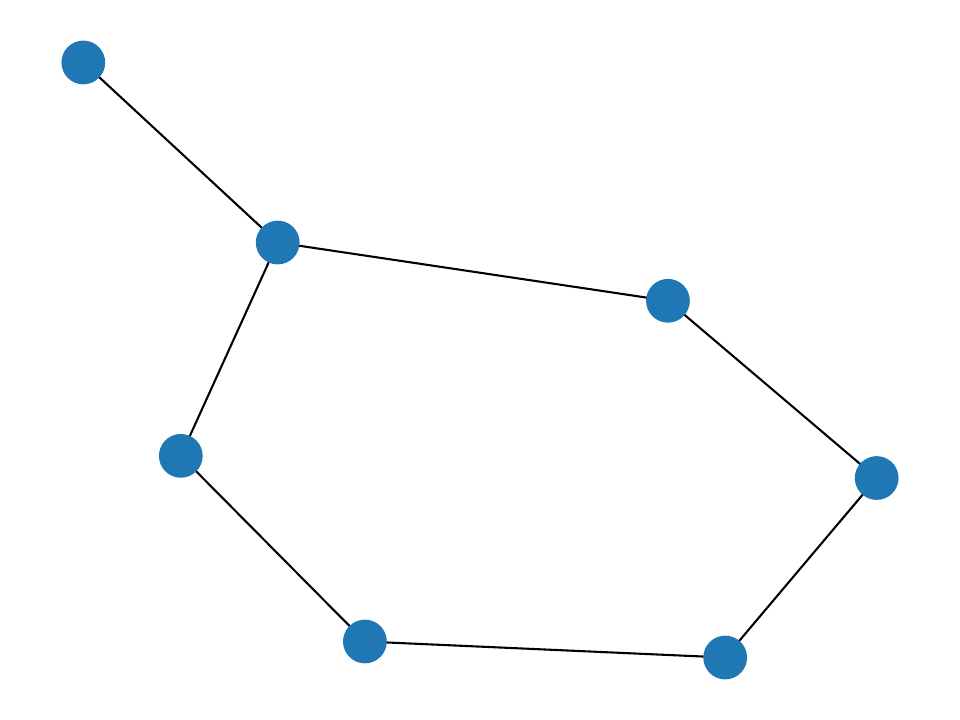} \\ 
    
      \textsc{Tree-Grids}
        & \includegraphics[width=1\linewidth]{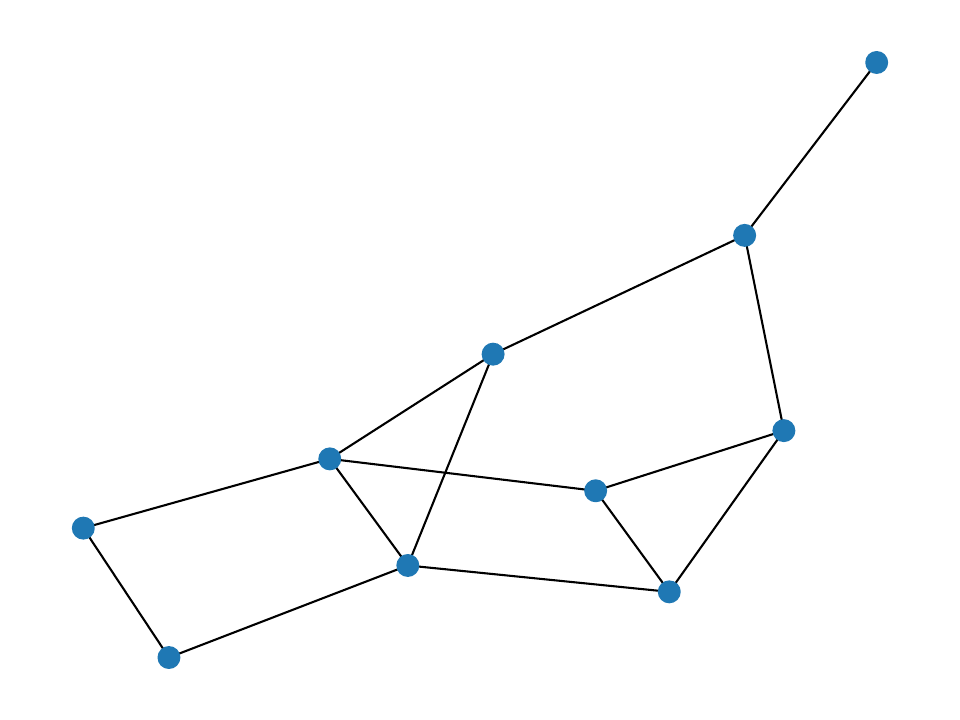}
        & \includegraphics[width=1\linewidth]{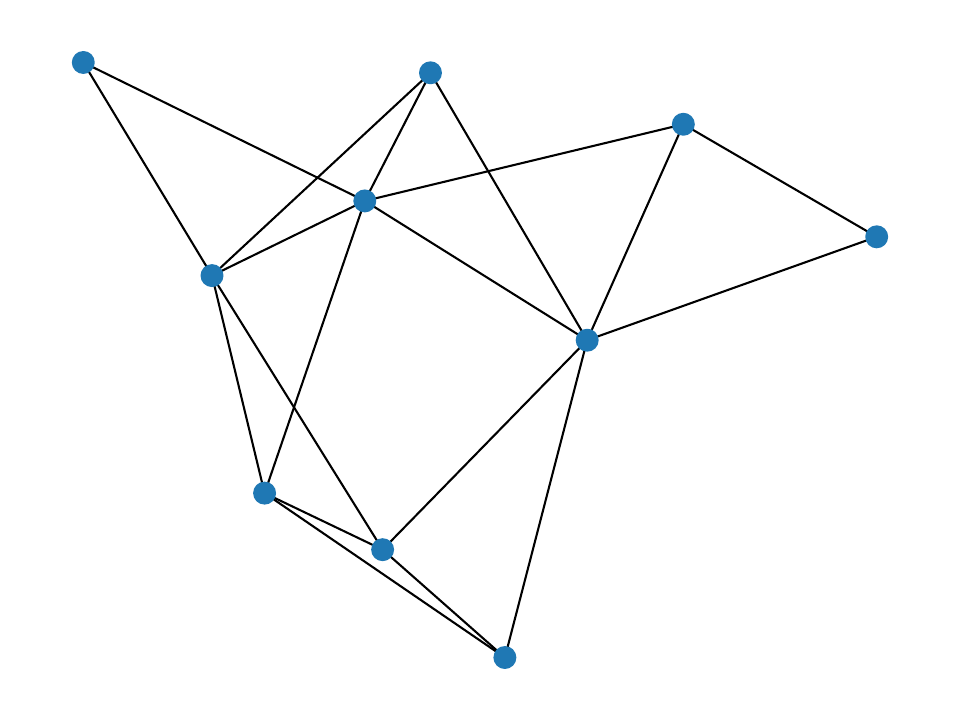}
        & \includegraphics[width=1\linewidth]{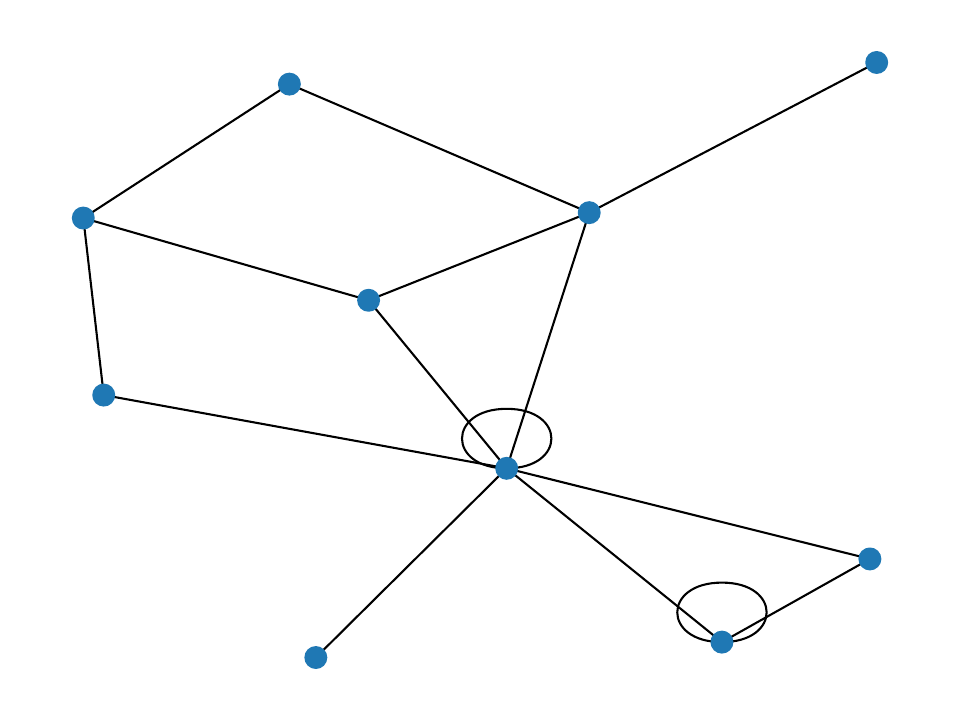}
        & \includegraphics[width=1\linewidth]{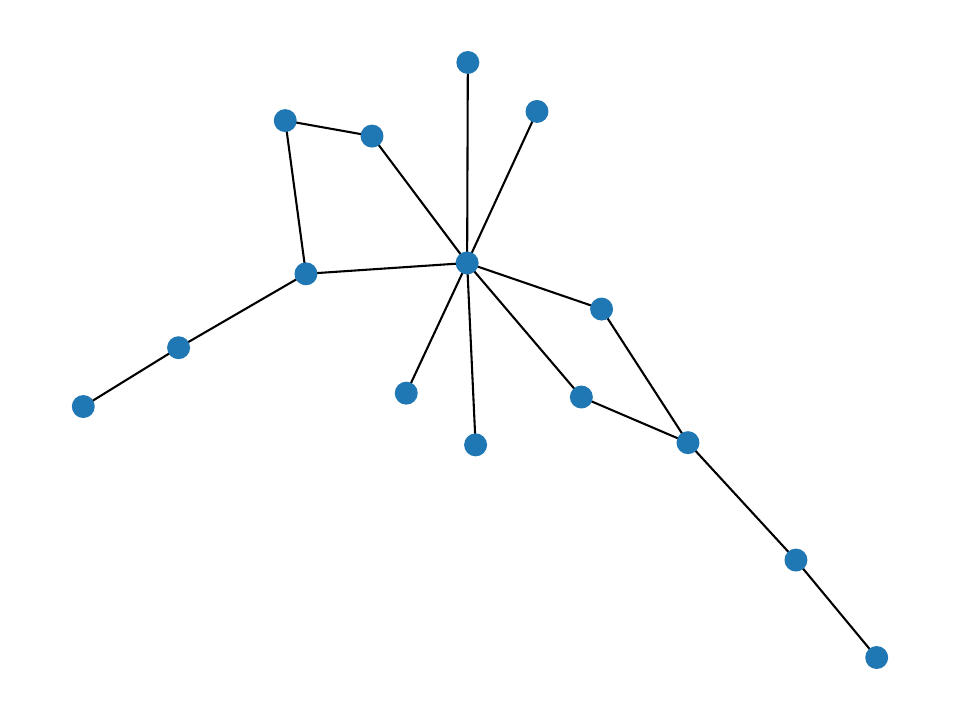}
        & \includegraphics[width=1.5cm]{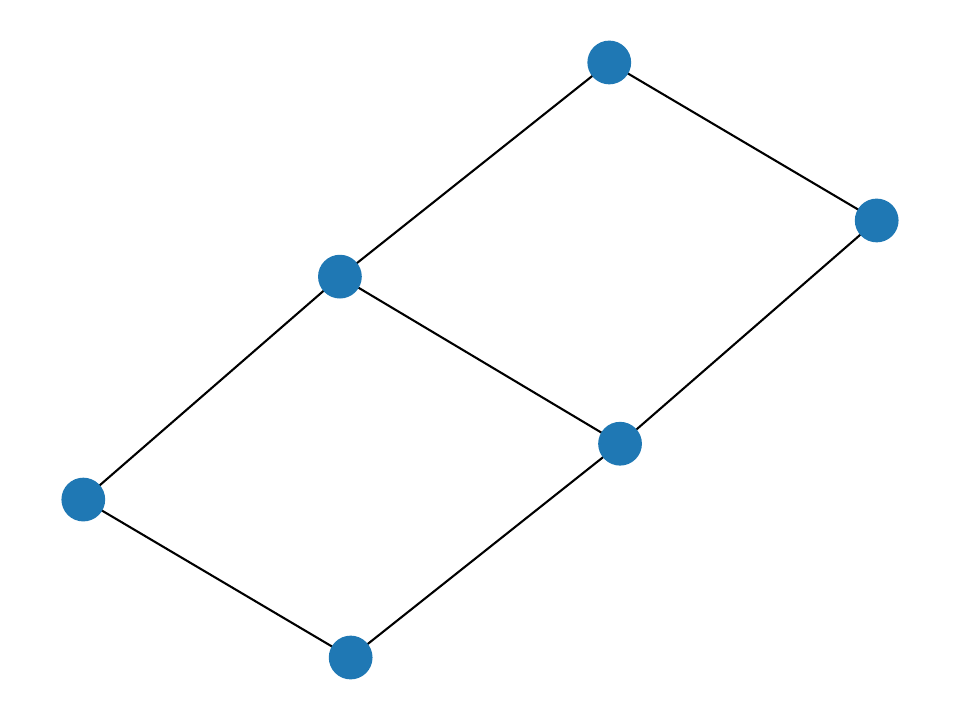}
        & \includegraphics[width=1.5cm]{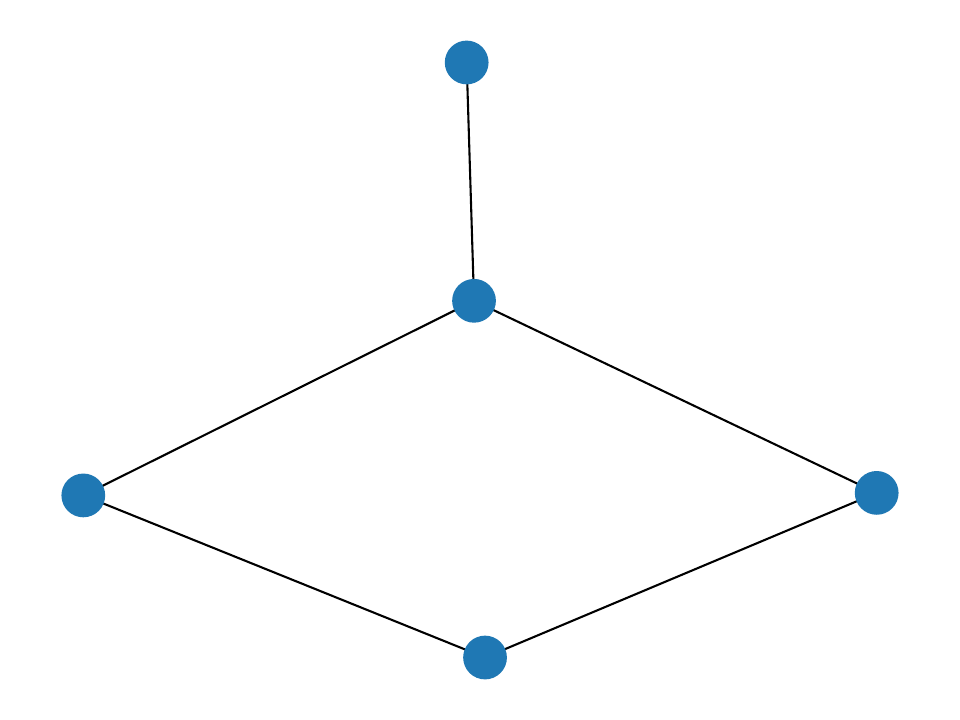}
        & \includegraphics[width=1.5cm]{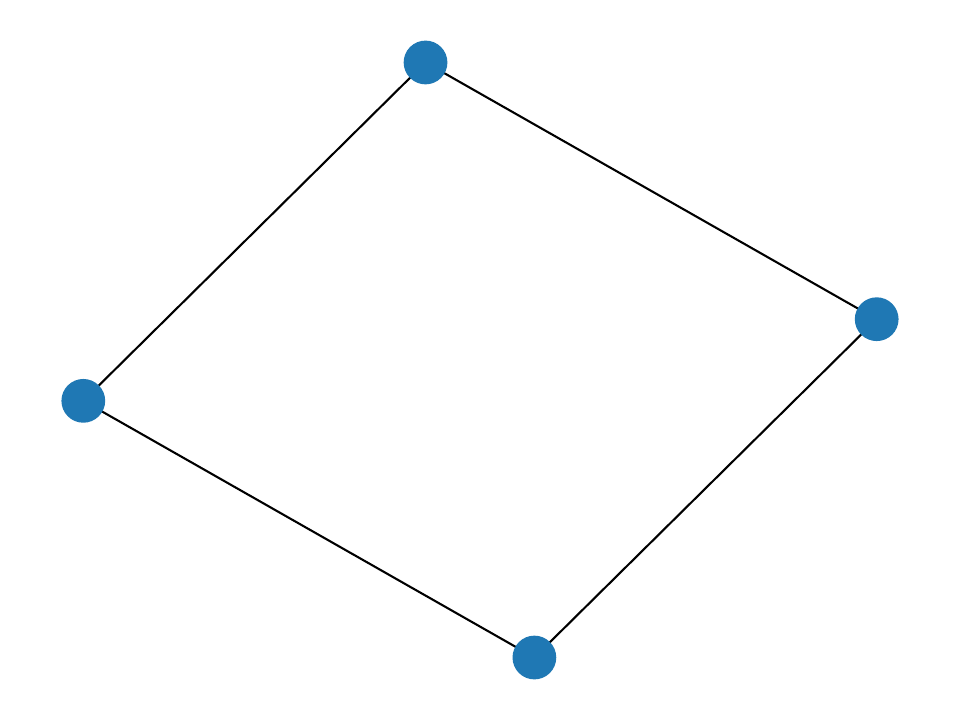} \\ 
    
      \textsc{BA-3Motif}
        & \includegraphics[width=1\linewidth]{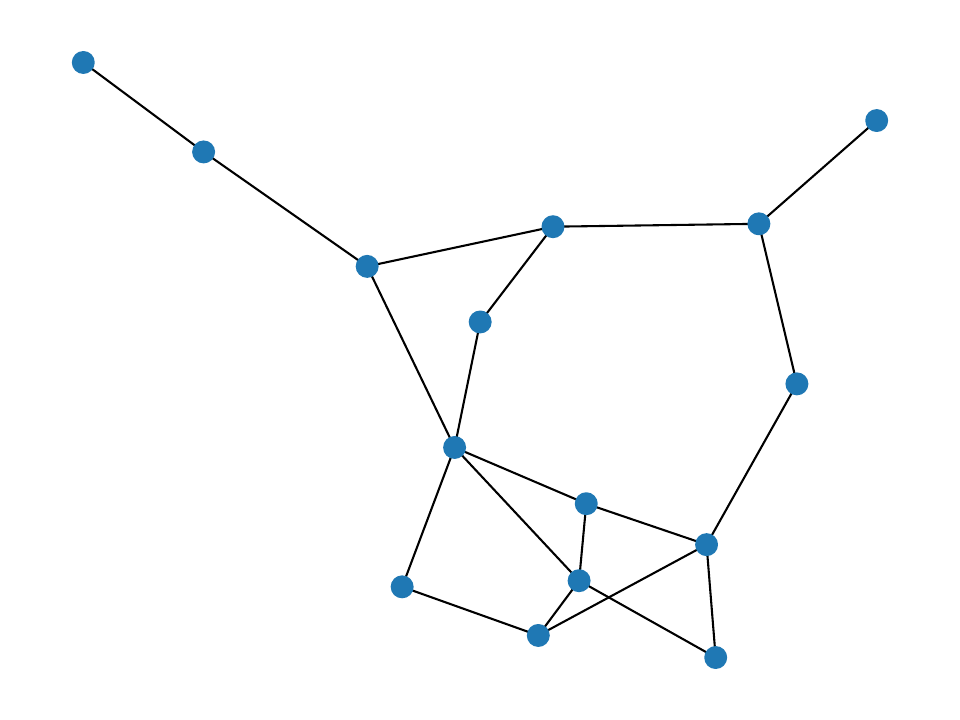}
        & \includegraphics[width=1\linewidth]{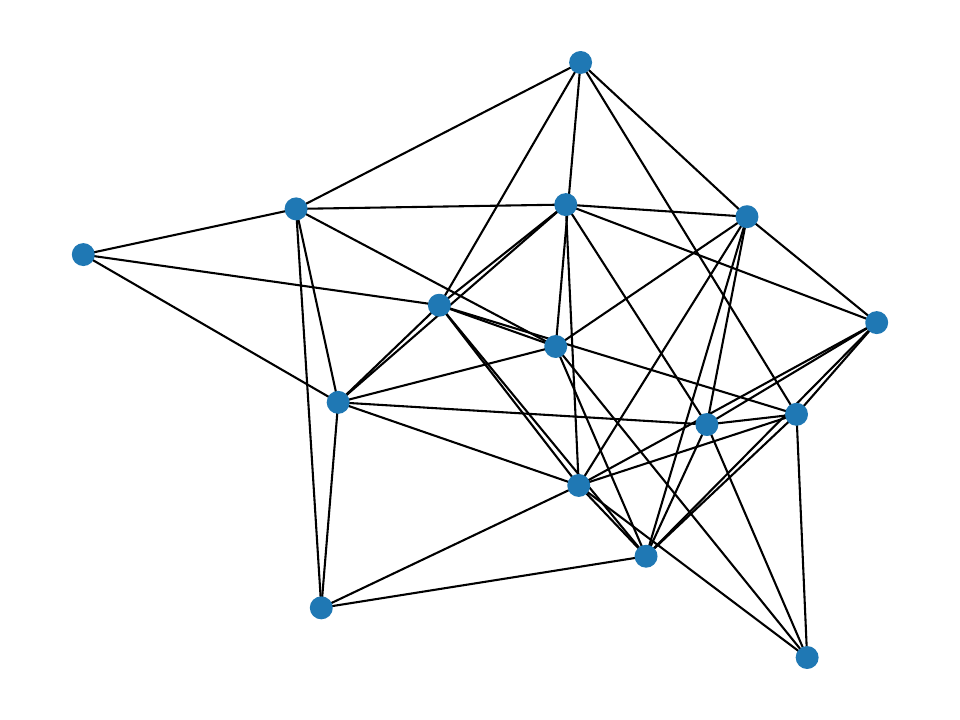}
        & \includegraphics[width=1\linewidth]{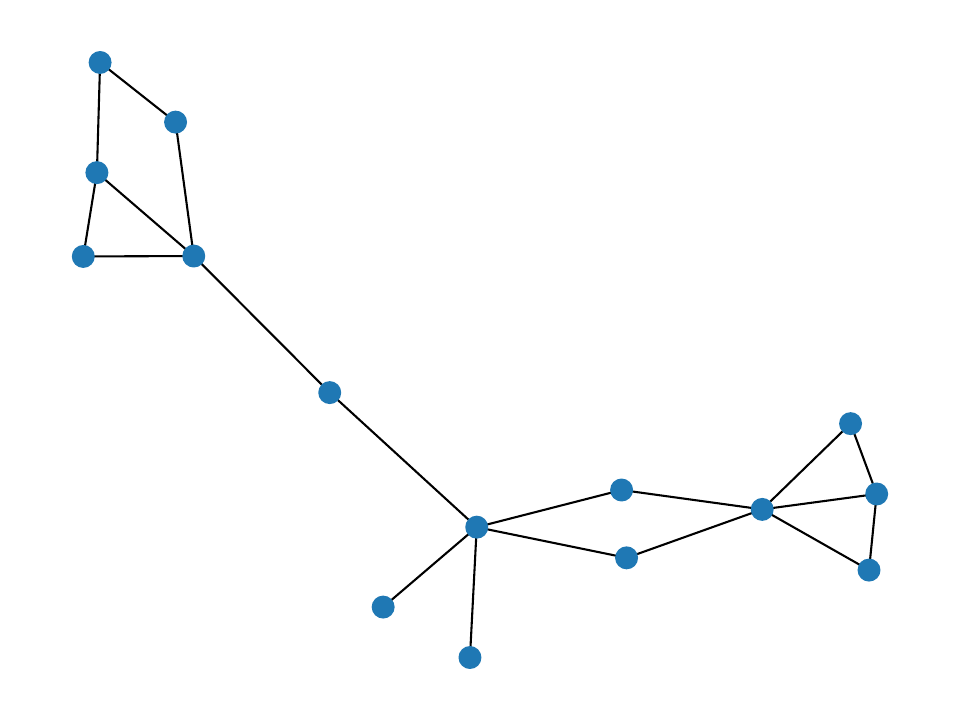}
        & \includegraphics[width=1\linewidth]{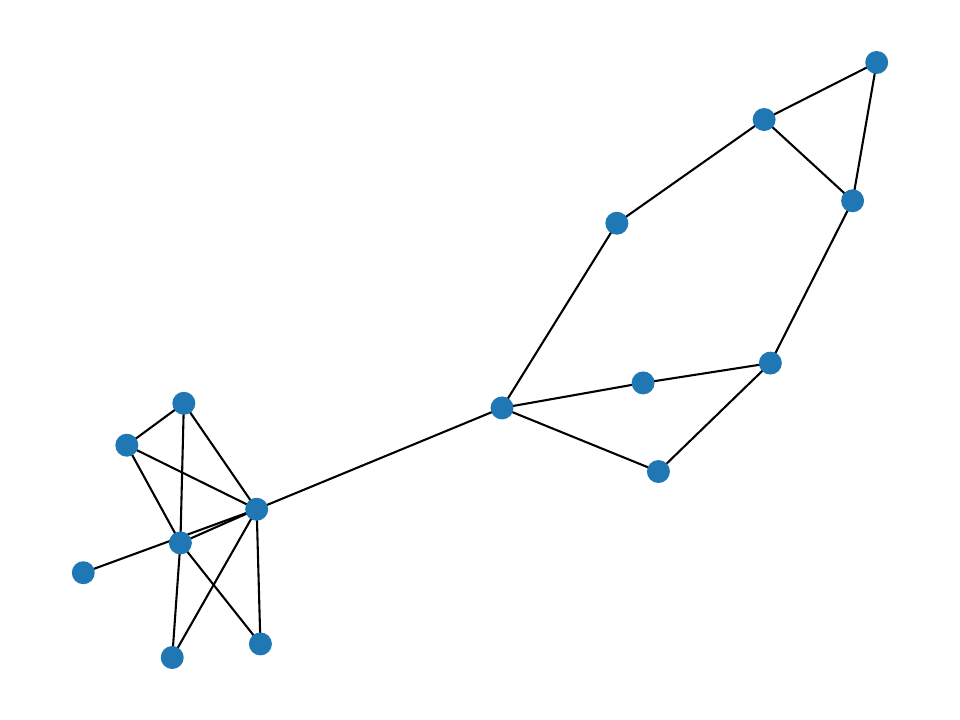}
        & \includegraphics[width=1.5cm]{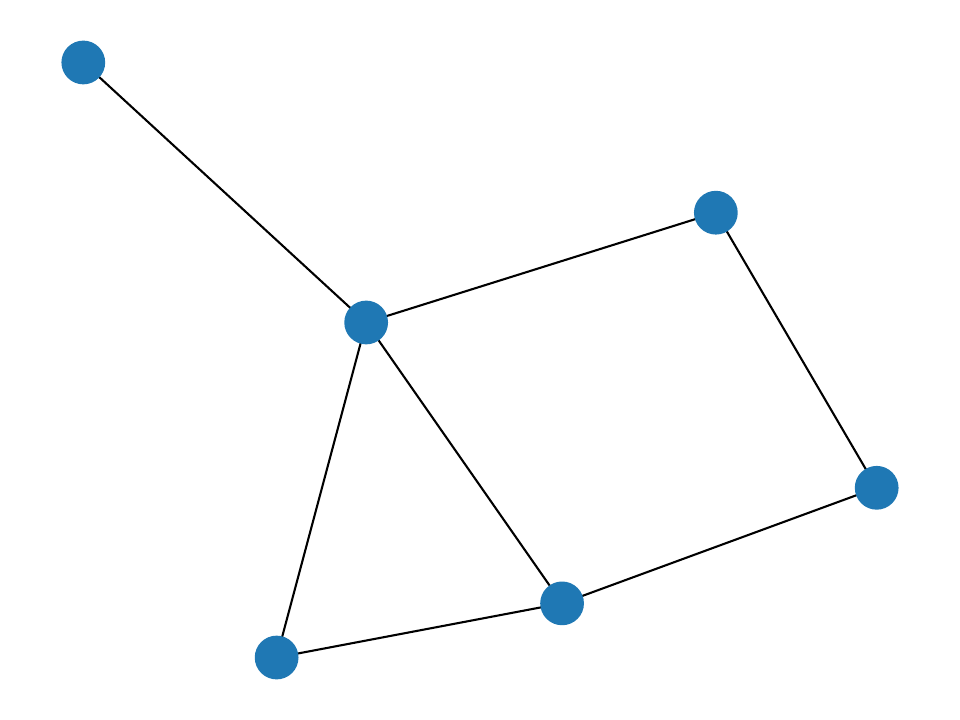}
        & \includegraphics[width=1.5cm]{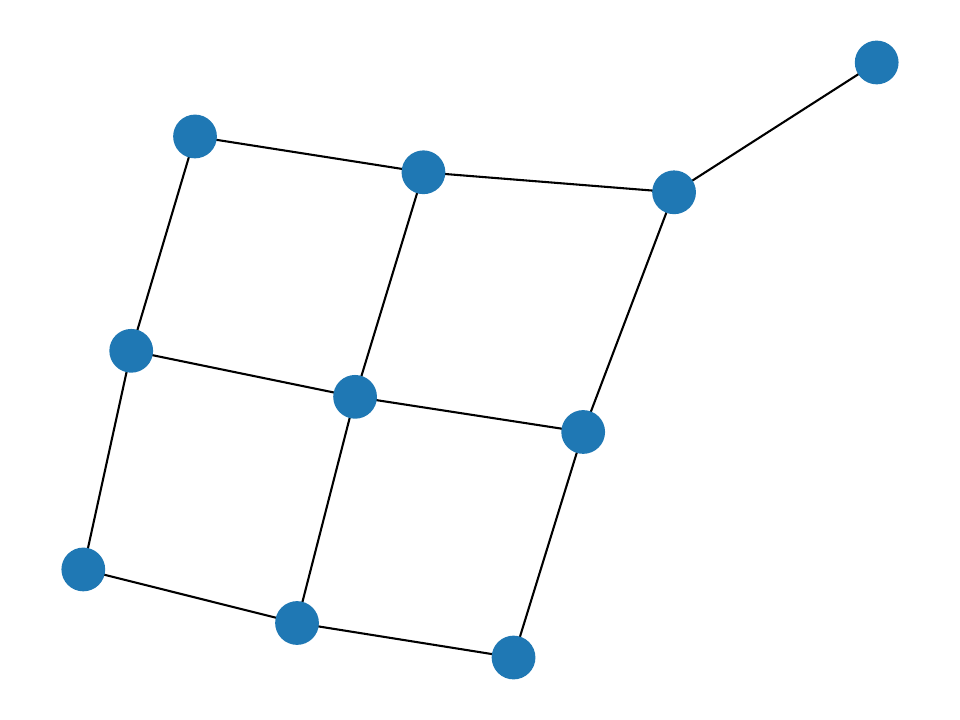}
        & \includegraphics[width=1.5cm]{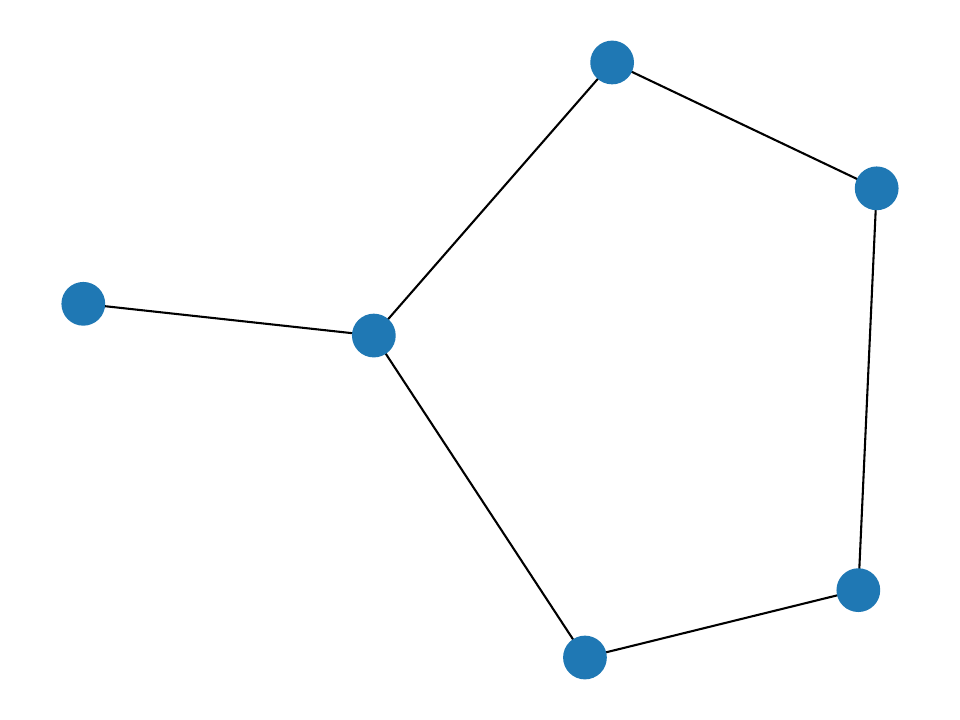} \\
      \bottomrule
    \end{tabular}
  }
\end{table}

\begin{table}[tb]
  \centering
  \setlength{\tabcolsep}{3pt}
  \scriptsize
  \caption{Motif statistics. Top: For datasets with \emph{class-specific} motifs, we report, for each class, the number of detected motifs that \emph{occur} and \emph{do not occur} in the explanation graphs. Bottom: For datasets with common motifs, we report the total number of motifs and how many motifs appear in the best explanation graph of \emph{every} class, in at least \(\frac{1}{2}\) of the classes, and in \emph{none} of the classes.}
  \label{tab:table-motif-statistics}
  \setlength{\tabcolsep}{4.1pt}
  \begin{tabular}{@{}lcccccccc@{}}
    \toprule
    \textbf{Dataset} & \multicolumn{2}{c}{\textbf{Class 0}} & \multicolumn{2}{c}{\textbf{Class 1}} & \multicolumn{2}{c}{\textbf{Class 2}} & \multicolumn{2}{c}{\textbf{Class 3}}\\
    \cmidrule(lr){2-3} \cmidrule(lr){4-5} \cmidrule(lr){6-7} \cmidrule(lr){8-9}
    & occur & not occur & occur & not occur & occur & not occur & occur & not occur \\
    \midrule
    \textsc{DBLP}      &  8 &  2 & 8 &  2 & 10 &  0 &  8 &  2 \\
    \textsc{IMDB}      &  9 &  1 & 9 &  1 & 10 &  0 & -- & -- \\
    \textsc{MUTAG}     & 30 & 14 & 3 & 11 & -- & -- & -- & -- \\
    \textsc{BA-3Motif} &  7 &  3 & 2 &  8 & 10 &  0 & -- & -- \\
    \bottomrule
  \end{tabular}\\%
  \setlength{\tabcolsep}{4.9pt}%
  \begin{tabular}{@{}lcccc@{}}
    \toprule
    \textbf{Dataset}  & \textbf{No. of motifs} &  \textbf{No. of motifs}   & \textbf{No. of motifs}  & \textbf{No. of motifs} \\
    & & \textbf{(every class)}  &\textbf{ (\(\frac{1}{2} \) of the classes)}  & \textbf{(none of the classes)} \\
    \midrule
    \textsc{BA-Shapes}  & 30 & 14 & 26 & 4 \\
    \textsc{Tree-Cycle} & 30 & 15 & 28 & 2 \\
    \textsc{Tree-Grids} & 30 & 28 & 30 & 0 \\ 
    \bottomrule
  \end{tabular}
\end{table}

\begin{table}[tb]
  \centering
  \scriptsize
  \caption{Feature selection by frequency}
  \label{tab:table-additional-frequency}
  \setlength{\tabcolsep}{10.5pt}
    \begin{tabular}{@{}lccc@{}}
      \toprule
      \textbf{Approach}  & \textbf{Node feat.(cos)$\uparrow$} & \textbf{PF$\uparrow$} & \textbf{GF$\uparrow$}\\
      \midrule
      \multicolumn{4}{c}{DBLP} \\ 
      \midrule
      DiTabDDPM (50 feat.)       & 0.233 & 0.902 \(\pm\) 0.102 & \textbf{0.867} \(\pm\) 0.046  \\
      DiTabDDPM (8 discr. feat.) & 0.355 & 0.797 \(\pm\) 0.062 & 0.852 \(\pm\) 0.054           \\
      DiTabDDPM (4 discr. feat.) & \textbf{0.387} & 0.770 \(\pm\) 0.121 & 0.820 \(\pm\) 0.055           \\
      DiTabDDPM (2 discr. feat.) & 0.384 & 0.972 \(\pm\) 0.041 & 0.845 \(\pm\) 0.052           \\
      \midrule
      TabDDPM (50 feat.)       & 0.194 & \textbf{0.998} \(\pm\) 0.001 & 0.842 \(\pm\) 0.054  \\ 
      TabDDPM (8 discr. feat.)  & 0.221 & 0.976 \(\pm\) 0.034          & 0.852 \(\pm\) 0.052 \\
      TabDDPM (4 discr. feat.) & 0.310 & 0.916 \(\pm\) 0.072          & 0.857 \(\pm\) 0.057  \\
      TabDDPM (2 discr. feat.) & 0.235 & 0.969 \(\pm\) 0.038          & 0.822 \(\pm\) 0.042  \\
      \midrule
      VAE (baseline) & 0.0 & 0.664 \(\pm\) 0.048 & 0.717 \(\pm\) 0.044 \\
      \midrule
      \multicolumn{4}{c}{IMDB} \\ 
      \midrule
      DiTabDDPM (3066 feat., all) & 0.029 & 0.955 \(\pm\) 0.084 & 0.890 \(\pm\) 0.029           \\
      DiTabDDPM (10 discr. feat.) & \textbf{0.083} & 0.977 \(\pm\) 0.066 & \textbf{0.933} \(\pm\) 0.036  \\
      DiTabDDPM (5 discr. feat.)  & 0.069 & 0.970 \(\pm\) 0.065 & 0.900 \(\pm\) 0.036           \\
      DiTabDDPM (2 discr. feat.)  & 0.040 & 0.984 \(\pm\) 0.039 & 0.913  \(\pm\) 0.047          \\
      \midrule
      TabDDPM (3066 feat., all)   & 0.048 & 0.912 \(\pm\) 0.104 & 0.890 \(\pm\) 0.036  \\
      TabDDPM (10 discr. feat.)   & 0.016 & 0.998 \(\pm\) 0.003 & 0.893 \(\pm\) 0.044  \\
      TabDDPM (5 discr. feat.)    & 0.011 & 0.973 \(\pm\) 0.068 & 0.896 \(\pm\) 0.040  \\
      TabDDPM (2 discr. feat.)    & 0.005 & 0.968 \(\pm\) 0.065 & 0.909 \(\pm\) 0.039  \\
      \midrule
      VAE (baseline) & 0.0 & \textbf{0.999} \(\pm\) 0.005 & 0.360 \(\pm\) 0.165  \\
      \bottomrule
    \end{tabular}
\end{table}

\section{Case Study---Explanations for the same class}

Table~\ref{tab:table-qualitative} shows a qualitative analysis of the explanation graphs obtained for the same class (class~1) for all datasets and all approaches, along with example motifs for each dataset.

\section{Feature Selection by Frequency}

Table~\ref{tab:table-additional-frequency} reports predictive faithfulness (PF), ground-truth faithfulness (GF), and cosine similarity (Node Feat. (CS)) for frequency-based feature selection.

\section{Hyperparameters}

For GNN training, we use the following settings. GraphSAGE is trained with a learning rate of 0.005 and a weight decay of 0.001. HAN is trained with a dropout of 0.6, a learning rate of 0.005, and a weight decay of 0.001. For GCN training, on \textsc{MUTAG}, we use a dropout of 0.3 and a learning rate of 0.0005, whereas for synthetic datasets we use 0.5 and 0.005, respectively.

For graph generation with diffusion models, we use a batch size of 256, a learning rate of 0.0002, a weight decay of 1e-12, and 5,000 diffusion steps. For DiTabDDPM, we use a batch size of 32, a learning rate of 0.05, a weight decay of 0.001, and 25,000 diffusion steps. For TabDDPM, we use the same hyperparameters as in the original implementation.

\section{Implementation}

All code is implemented in Python~3.10.6. We utilize Pandas for data manipulation, Matplotlib and Seaborn for visualization, NetworkX for graph plotting, PyTorch Geometric for GNN training, scikit-learn for sampling node features and comparing feature distributions, the Forest Fire Sampler for sampling subgraphs, the Deep Graph Library (DGL) for creating metagraphs, and Python-louvain for community detection.

The experiments were conducted on a machine with Ubuntu 22.04.2 LTS, an AMD Ryzen 9 CPU, and an NVIDIA GeForce RTX 3070 GPU with 8GB of GPU memory.

\end{document}